\documentclass[twocolumn]{fairmeta}
\usepackage[most]{tcolorbox}

\usepackage{amsmath}
\usepackage{amssymb}

\title{Learning Latent Action World Models In The Wild}

\author[1]{Quentin Garrido}
\author[1]{Tushar Nagarajan}
\author[1,2]{Basile Terver}
\author[1]{Nicolas Ballas}
\author[1,3]{Yann LeCun}
\author[1]{Michael Rabbat}

\affiliation[1]{FAIR at Meta}
\affiliation[2]{Inria}
\affiliation[3]{NYU}

\abstract{
Agents capable of reasoning and planning in the real world require the ability of predicting the consequences of their actions. While world models possess this capability, they most often require action labels, that can be complex to obtain at scale. This motivates the learning of latent action models, that can learn an action space from videos alone. Our work addresses the problem of learning latent actions world models on in-the-wild videos, expanding the scope of existing works that focus on simple robotics simulations, video games, or manipulation data. While this allows us to capture richer actions, it also introduces challenges stemming from the video diversity, such as environmental noise, or the lack of a common embodiment across videos.
To address some of the challenges, we discuss properties that actions should follow as well as relevant architectural choices and evaluations. We find that continuous, but constrained, latent actions are able to capture the complexity of actions from in-the-wild videos, something that the common vector quantization does not. We for example find that changes in the environment coming from agents, such as humans entering the room, can be transferred across videos. This highlights the capability of learning actions that are specific to in-the-wild videos.
In the absence of a common embodiment across videos, we are mainly able to learn latent actions that become localized in space, relative to the camera. Nonetheless, we are able to train a controller that maps known actions to latent ones, allowing us to use latent actions as a universal interface and solve planning tasks with our world model with similar performance as action-conditioned baselines. Our analyses and experiments provide a step towards scaling latent action models to the real world.
}

\correspondence{Quentin Garrido at \email{garridoq@meta.com}}

\begin{document}

\maketitle

\section{Introduction}

To build intelligent systems that can reason and plan in the real world, we must build systems that can predict the future, and in particular consequences of their actions\citep{friston2010free,clark2013whatever,bubic2010prediction,lecun2022AMI,sutton1991dyna,ha2018worldmodels,hafner2019dreamer,nguyen1990truck}.
As soon as agents are present in the scene, predicting the future becomes a stochastic endeavor that can be parametrized by possible actions.
Modeling these possible futures is thus necessary to learn good models of the world, ones that can for example be used to solve planning problems. A significant body of literature on world models is available to us assuming that we possess action labels~\citep{ha2018worldmodels,hafner2019dreamer,hafner2023dreamerv3,hu2023gaia1,bar2024navigation,agarwal2025cosmos,assran2025vjepa2}.
This access to actions is a critical bottleneck: the vast majority of video data available online is unlabeled~\citep{zellers2022merlot,miech2019howto100m} and includes diverse embodiments.

\begin{figure}[!t]
    \centering
    \includegraphics[width=0.8\columnwidth]{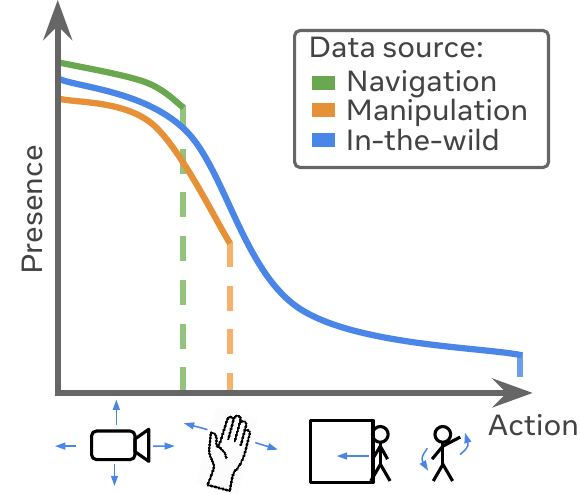}
    \caption{\textbf{Action diversity.} Classically used navigation or manipulation data contains the most general actions, such as camera or hand movements. In-the-wild videos extend this to a much broader distribution of actions, with objects entering the scene or people dancing. }
    \label{fig:act_distrib}
\end{figure}

This gap motivates the idea of learning a latent action model (LAM)~\citep{edwards2019imitating,rybkin2018learning,menapace2022playable,schmidt_actw-without-actions_2024,ye_lapa_2025,yang2025como,chen_igor_2024,cui_dynamo_2024} that can discover the action space from videos alone, without action annotations or a known embodiment. The standard approach is to learn two components jointly. First, an inverse dynamics model (IDM) that, given observations of the past and future, predicts a latent action that explains the difference between the two. Second, a forward model which predicts the future using the past and obtained latent action. After such model is trained, the IDM can be used as part of a VLA pipeline~\citep{bu_univla_2025,ye_lapa_2025} or to train a world model, using the frozen IDM~\citep{gao2025adaworld}.

The type of unlabeled videos that are used is critical to the learned action space, and often an understudied component.
Most LAM studies rely on narrow, task-aligned domains--video games~\citep{bruce2024genie}, tabletop manipulation~\citep{nikulin2025latent}, or curated real manipulation~\citep{bu_univla_2025,gao2025adaworld}--which can yield action spaces specialized to a single embodiment with limited transfer or generalization.
While some works have use more ``natural'' videos such as Ego4D~\citep{grauman2022ego4d}, it usually amounts to a minority of the training data, e.g. 5\% for ~\cite{bu_univla_2025} and ~\cite{gao2025adaworld}, far from leveraging the richness of in-the-wild videos.

\begin{figure*}[tbhp]
    \centering
    \includegraphics[width=\textwidth]{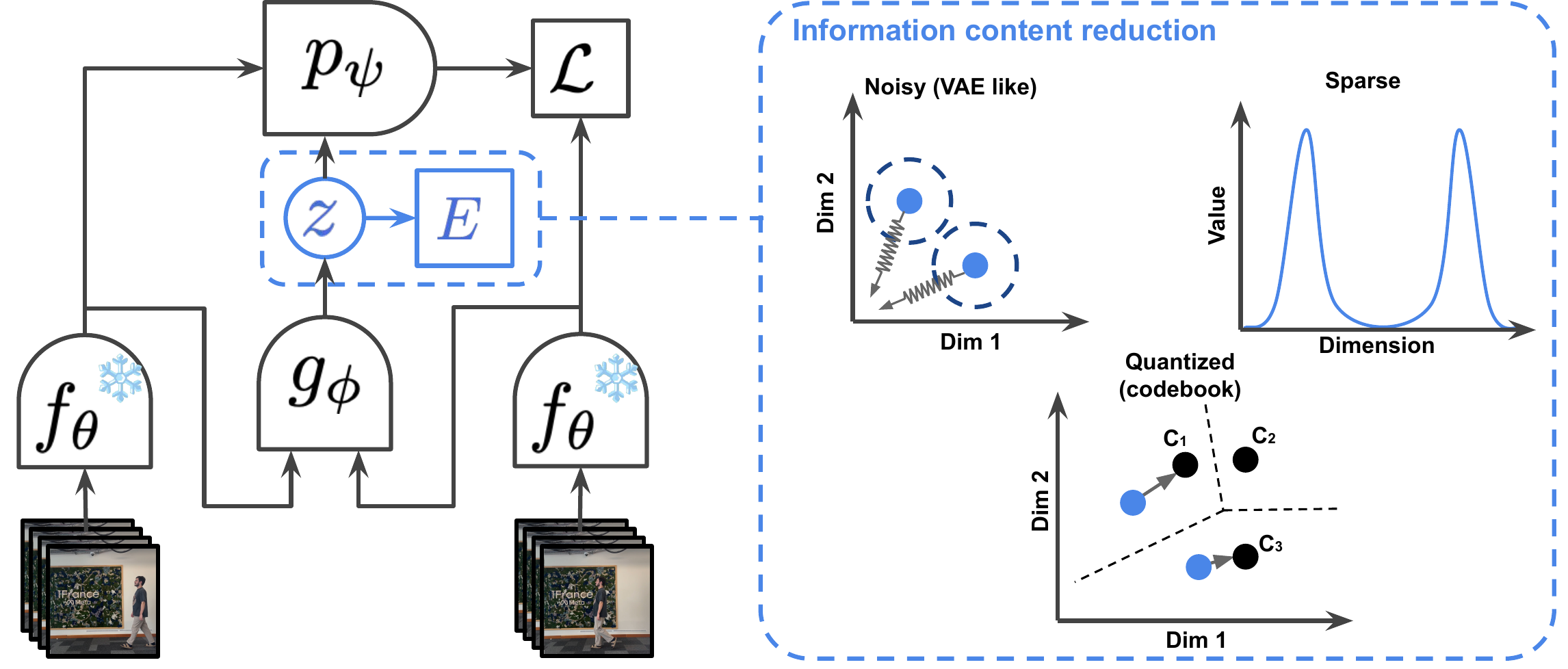}
    \caption{\textbf{Latent action world model.} A classical world model is endowed with actions represented as latent variables. These latent actions are obtained thanks to an inverse dynamics model trained jointly with the world model. To limit their information content (and propensity to cheat), they are regularized using techniques such as noise addition, sparsification, or quantization.}
    \label{fig:fig1}
    \vspace{-1em}
\end{figure*}

To learn a truly general and transferable latent action world model, we argue that we must go beyond these targeted data sources. Sources of natural in-the-wild videos such as HowTo100M~\citep{miech2019howto100m} or YoutubeTemporal-1B~\citep{zellers2022merlot} provide a much richer and general learning environment than usually studied, as illustrated in Figure~\ref{fig:act_distrib}. However this introduces a new set of research challenges that we address in this work to demonstrate the viability of LAM on large scale in-the-wild natural videos\footnote{While our work does not focus on video generation, LAM trained on in-the-wild videos could be used to remove the necessity of text-video pairs~\citep{sun2024video}.}.

First and foremost, the meaning of an "action" on in-the-wild video is not as clearly defined as it is in environments with known action spaces. Metaphorically speaking, the first dimension--or principal component-- of actions could be movements, something shared across video sources. From then we can have a split between ego- and exo-centric actions, which separates actions of the camera wearer and other agents in the environment. In in-the-wild videos, we have a stronger presence of external agents performing diverse actions, on top of what the camera wearer does. Going deeper in the action distribution, in-the-wild videos will contain unique actions such as cars entering the frame, people dancing, fingers forming chords on a fretboard, etc. This leads to an inherent richness of actions that we aim at modeling. In-the-wild videos provide a superset of actions compared video games or manipulation videos, which means that one should still be able to solve more classical navigation or manipulation tasks.
While data sources used in previous works would mainly contain the metaphorical first principal components of actions, trying to model more diverse actions has a risk of capturing more environmental noise~\citep{nikulin2025latent} such as leaves oscillating on trees.
Finally, agents in in-the-wild videos do not have a consistent embodiment that the model can latch onto, which poses challenges for transfer and downstream applicability of the learned latent actions.

The focus of our work thus lies in the study of latent action world models trained on large scale in-the-wild video datasets, studying the inherent challenges, potential pitfalls of latent actions in such setting, as well as demonstrating their viability.\\
\noindent Our contributions are as follows:
\begin{itemize}
    \item We conduct a study on how to regulate the information content of latent actions, focusing on in-the-wild natural videos. We find that while sparse or noisy latent actions can effectively model complex actions, discrete ones struggle to adapt.
    \item We show that the absence of a common embodiment across in-the-wild videos is not an issue when learning latent actions. Latent actions will encode more spatially-localized transformations.
    \item We demonstrate the generality of the learned action space by transferring complex actions between videos. We find that we can effectively transfer motion between objects, or actions such as  someone entering the frame.
    \item We demonstrate how our learned latent action space can be used as a universal action space. By training a small controller to map known actions to latent ones, our world model trained only on natural videos can be controlled to solve robotic manipulation and navigation tasks, achieving planning performance close to models trained on domain-specific, action-labeled data.
\end{itemize}

Overall, our work demonstrates the feasibility of learning a latent action conditioned world model purely using natural in-the-wild videos.

\section{Related works}

\textbf{World Models.}
World Models~\citep{nguyen1990truck,sutton1991dyna,ha2018worldmodels} have become a very active area of research. While a significant body of work had been applied to game data~\citep{alonso2024diffusion,hafner2019dreamer,hafner2023dreamerv3}, applications to more complex environments, such as simulated robotics environment~\citep{seo2023masked,zhou2024dino} or the real world~\citep{hu2023gaia1,agarwal2025cosmos,assran2025vjepa2} have flourished recently. With a plethora of possible embodiments and action space, works such as NWM~\citep{bar2024navigation} focus on locomotion, PEVA~\citep{bai2025peva} on whole body control, or UniSim~\citep{yang2023unisim} which can handle a variety of embodiments though textual control, have appeared.
The promise of such models is not solely to generate visually appealing videos~\citep{brooks2024sora,teng2025magi,agarwal2025cosmos} but mainly lies in their use to solve visual planning tasks. Being able to predict the consequences of actions can enable us to solve problems for navigation~\citep{shah2021rapid}, robotic manipulation in simulation~\citep{nasiriany2024robocasa,liu2023libero,yu2020meta} or in the real world~\citep{khazatsky2024droid}, or even whole body control~\citep{ma2024nymeria}.
Such models can even be used to solve more classical vision tasks such as segmentation and depth forecasting~\citep{baldassarre2025back,karypidis2024dino,luc2017predicting}.
A common issue to obtain models that generalize across embodiments is how to define a common action space ? A solution can for example be to use the maximal dimensionality across considered embodiments, with an embodiment token~\citep{hansen2023td}, but this is not easily scalable. This is where latent action models~\citep{edwards2019imitating,rybkin2018learning,schmidt_actw-without-actions_2024,bruce2024genie} come into play, as one of their promises is to learn an abstract, general latent action space.

\textbf{Latent Action Models.} Latent action models aim at learning actions from unlabeled videos. Latent actions can be inferred using a latent policy~\citep{edwards2019imitating}, or by using an explicit inverse dynamics model (IDM) that predicts the latent action from the past and future frames~\citep{rybkin2018learning,menapace2021playable,menapace2022playable,schmidt_actw-without-actions_2024}. This is then combined with a forward model that predicts the future frame from the past and the latent action.  
The used of an IDM introduces a causal leakage in information and a key challenge is to ensure that the latent actions do not capture too much information, e.g. the entire next frame.
A commonly used approach is to discretize the latent actions. This is the approach of choice in methods such as ILPO~\cite{edwards2019imitating}, LAPO~\citep{schmidt_actw-without-actions_2024}, Genie~\citep{bruce2024genie}, LAPA~\citep{ye_lapa_2025}, or UniVLA~\citep{bu_univla_2025}. This can for example be motivated by prior knowledge of the desired action space~\citep{bruce2024genie}. Other methods such as CLASP~\citep{rybkin2018learning}, CoMo~\citep{yang2025como}, or AdaWorld~\citep{gao2025adaworld} instead opt for a continuous space, which is inherently more flexible. In this case, a regularization term can be added to reduce the information content of the latent actions. Other works instead rely on carefully designed forward model architectures~\cite{menapace2022playable,sun2024video} to structure the latent action space.
Furthermore, while numerous methods use off-the-shelf vision encoders to encode frames, latent actions are still often learned by predicting the future frame in pixel space~\citep{chen2025moto,yang2025como,ye_lapa_2025}. This makes latent actions more susceptible to distractors~\citep{nikulin2025latent}, where the latent actions learn to encode background noise rather than the actions we desire. While a solution is to use supervision~\citep{nikulin2025latent,liang2025clam}, working in an abstract latent spaces and carefully designing latent actions can help avoid some of these issues, as we study throughout our work.
In general, while learning latent actions has clear applicability to world models, methods tend to be developed with VLAs in mind~\citep{bu_univla_2025,ye_lapa_2025}. Even if the approaches are architecturally similar to world models, where the forward model/action decoder can be seen as a world model, it is often discarded. Even when a world model is trained, a two-stage approach is commonly used, where the world model is trained after the inverse dynamics model~\citep{yang2025como}. Concurrently to our work~\cite{wang2025coevolvinglatentactionworld} proposes to treat the forward model as a world model, by using a pretrained video generation model.

\section{Problem setting}

Considering a video $V$ where the state of the world at each timestep $t$ is $s_t$, we are interested in modeling the evolution of the world, i.e. find a function $f$ such that $s_{t+1} = f(s_{0:t})$.
However, the presence of agents as well as general stochasticity make the prediction non deterministic and thus this formulation is insufficient. We can model the uncertainty of the prediction with a latent variable $z_t$ containing the relevant information, such that $s_{t+1} = f(s_{0:t},z_t)$. Another way to model uncertainty is to not consider $s_{t+1}$ directly, but instead output a distribution over possible futures $p(s_{t+1} | s_{0:t})$, as is commonly done in text~\citep{radford2018gpt} or with quantized representations~\citep{hu2023gaia1,agarwal2025cosmos}.\\
\noindent Nonetheless, formalizing future prediction as $s_{t+1} = f(s_{0:t},z_t)$ is appealing as we can interpret part of $z_t$ as actions happening in the scene. This is for example the case when learning a world model for robotics, where in simple environments no stochasticity exists beyond the actions $a_t$ of the agent. We thus have $s_{t+1} = f(s_{0:t},a_t)$. If an environment is stochastic, we have both noise from the environment and actions which prompts a more complex formalism than previously where we want $s_{t+1} = f(s_{0:t},a_t,z_t)$. This is reminiscent of diffusion based world models~\citep{alonso2024diffusion,bar2024navigation} for example.

Latent action models~\citep{edwards2019imitating,rybkin2018learning,schmidt_actw-without-actions_2024} aim at modeling the actions happening in a scene, without capturing exogenous noise that may come from the environment.
To do so, most methods introduce a leak of causality by looking at the future to infer $z_t$. This is commonly done with an inverse dynamics model (IDM)\footnote{We can see $z_t$ as the result of an optimization process minimizing the prediction error over it. Implementing it this way is impractical, but we can see the IDM as performing amortized inference~\citep{amos2023tutorial}. This lends itself well to gradient based optimization at inference time.} that takes as input the past and future frames and outputs the latent action $z_t = g_\phi(s_{t},s_{t+1})$.
\begin{figure*}[tbhp]
    \centering
    \begin{tabular}{@{}m{0.001\textwidth} m{0.99\textwidth}@{}}
        & \makebox[\linewidth][s]{\hspace{.03\linewidth}\textbf{Context}\hspace{.4\linewidth}\textbf{Prediction}} \\
        \rotatebox{90}{\textbf{Groundtruth}} & \includegraphics[width=\linewidth]{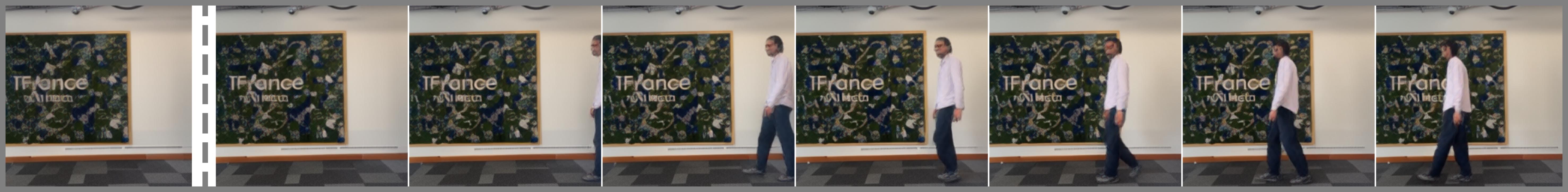} \\
        \rotatebox{90}{\textbf{Sparse}} & \includegraphics[width=\linewidth]{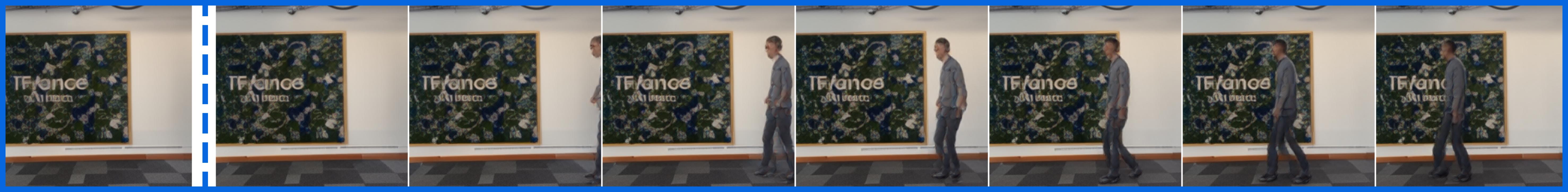} \\
        \rotatebox{90}{\textbf{Noisy}} & \includegraphics[width=\linewidth]{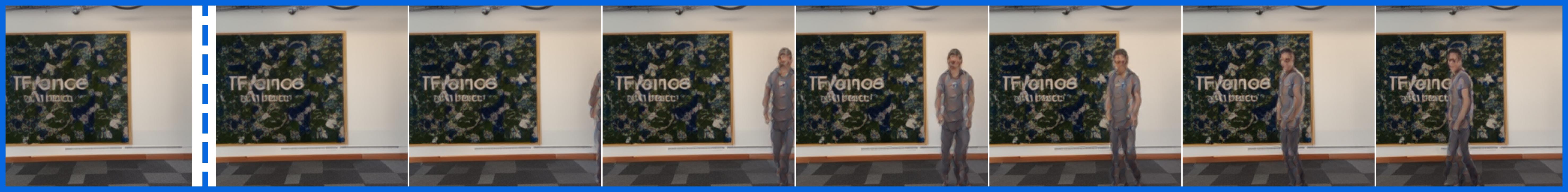} \\
        \rotatebox{90}{\textbf{Discrete}} & \includegraphics[width=\linewidth]{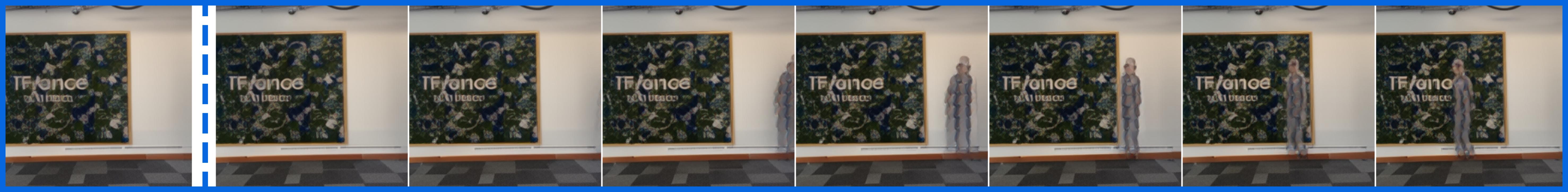} \\
    \end{tabular}

    \caption{\textbf{Sample predictions using the IDM.} We illustrate the highest quality unrollings obtained with different regularization, using the inverse dynamics model. While sparse or noisy latent actions are able to capture a man entering the scene, discrete ones are not able to properly capture such action, even if some motions remains captured.}
    \label{fig:train-unroll}
\end{figure*}
From this, we can then train a world model (also called forward model) $p_\psi$ to estimate $s_{t+1}$ using the following loss function:
\begin{equation*}
    \mathcal{L}_{t} = \| s_{t+1} - p_\psi(s_{0:t},z_t)  \|_1 \;, \text{with} \; z_t = g_\phi(s_{t},s_{t+1}).
\end{equation*}
This works well in clean environments~\citep{hoque2025egodex,yu2020meta} since the stochasticity comes mainly from actions performed by the well-defined agent. However, on videos that are in-the-wild~\citep{zellers2022merlot,miech2019howto100m} there is a significant risk of capturing exogenous noise, such as leaves oscillating on trees. Limiting the information content of latent actions thus becomes paramount, balancing between capturing complex actions and capturing noise, or even worse, encoding the whole next state in the latent action.\\
\noindent In general, this information regularization aims at finding the \textit{minimal} latent actions that can explain the prediction of the future. Throughout this work we focus on three distinct mechanisms, each with pros and cons.

\textbf{Sparsity.} The first one, and perhaps most complex to implement, is sparsity based constraints~\citep{drozdov2024video}. Here, we would like for the latent actions to have as low of an L1 norm as possible. Due to trivial solutions that would reduce the L2 norm of the vectors, concentrate the norm along a few dimensions, or focus too much around the mode of the latent distribution, a few additional regularizations are added. The regularization is then
\begin{equation*}
    \mathcal{L}(Z) = VCM(Z) + \frac{1}{N}\sum_i E(Z_i) ,
\end{equation*}
with
\begin{equation*}
    E(z) = \lambda_{l2}\max\left(\sqrt{D} - \|z\|_2^2,0\right) + \lambda_{l1}\|z\|_1
\end{equation*}
and
\begin{align*}
    VCM(Z) = &\lambda_{V} \frac{1}{D} \sum_d \max\left(1-\sqrt{\text{Var}(Z_{\cdot,d})},0\right) \\
             & + \lambda_{C}\frac{1}{D(D-1)} \sum_{i\neq j} \text{Cov}(Z)_{i,j}^2 \\
             & + \lambda_{M}\frac{1}{N D} \sum_{i,j} Z_{i,j} .
\end{align*}
This Variance-Covariance-Mean (VCM) regularization, inspired by VICReg~\citep{bardes2021vicreg}, ensures an adequate spread of information and forces the sparsity constraints to be properly used by the model. In practice we set the coefficients to $\lambda_{l2}=1$, $\lambda_{V}=0.1$, $\lambda_{C}=0.001$, $\lambda_{M}=0.1$, and vary $\lambda_{1}$ to regulate information content.

\textbf{Noise addition.} Another approach to limit information content in the learned latent actions is to add noise to them, while making sure their norm does not increase and makes the noise negligible. This can be implemented in a similar way as a VAE~\citep{kingma_auto-encoding_2014,gao2025adaworld}. The prior matching term here acts as our regularizer, where the target standard deviation adds noise while the target mean reduces the norm of the latent actions.
\begin{equation*}
    \mathcal{L}(z_t) = - \beta \;D_{KL}\left(q(z_t| s_t,s_{t+1}) || \mathcal{N}(0,1)\right)
\end{equation*}
\textbf{Discretization.} A final approach is to discretize the latent actions. For this, the most common approach is vector quantization~\citep{van2017vqvae} or a variant of it. This serves as a baseline comparison to illustrate a commonly used regularization in previous works\citep{ye_lapa_2025,bu_univla_2025}. In practice, we use the same quantization scheme as UniVLA~\citep{bu_univla_2025}, using classical vector quantization~\citep{van2017vqvae} as well as codebook reset for unused codes.

All of this can be performed in the latent space of trained encoder where $s_t$ and $s_{t+1}$ now are the representations obtained from video frames, which leads us to the complete architecture illustrated in Figure~\ref{fig:fig1}.


\section{Experimental details}

We now turn ourselves to a more practical implementation. A video $V$ of length $T$ is encoded through a frame causal encoder $f_\theta$ --V-JEPA 2-L~\citep{assran2025vjepa2} in our experiments-- producing representations $s_{0:T-1}$. This encoder is kept frozen during training.
We then train the world model $p_\psi(s_{0:t},z_t)$ and inverse dynamics model $g_\phi$ jointly to predict $s_{t+1}$ using the aforementioned prediction loss and latent action regularization.\\
To increase efficiency, we train the model using teacher forcing~\citep{williams1989learning,vaswani2017attention}. By default, $p_\psi$ is implemented as a ViT-L~\citep{dosovitskiy2021vit} using RoPE~\citep{su2021rope,assran2025vjepa2} for positional embeddings. To condition $p_\psi$ on $z$ we use AdaLN-zero~\citep{peebles2023scalable} that we adapt to condition the sequence frame-wise. Our latent actions $z_t$ are 128 dimensional continuous vectors by default.
Unless specified otherwise, all models are trained on YoutubeTemporal-1B~\citep{zellers2022merlot} with 16 frames clips at 4 fps, for $30 000$ iterations at a batch size of $1024$. We use the Muon optimizer~\citep{jordan2024muon} with a learning rate of 0.02 and AdamW~\citep{loshchilov2018adamw} learning rate of $6.25\times10^{-4}$ following a linear warmup over 10\% of the training followed by cosine annealing. We use 0.04 as weight decay.

For visualization purposes, we also train a frame causal video decoder using a ViT-L trained with a combination of $L_1$ and perceptual loss~\citep{johnson2016perceptual,zhang2018unreasonable}. While generation is not core to our work, this is a useful tool to compute perceptual metrics and inspect the model's prediction.
Confer Supplementary Section~\ref{sec:detailed_protocol} for detailed protocols.

\section{Performance of information regularizations}

\begin{figure}[!t]
    \centering
    \includegraphics[width=\columnwidth]{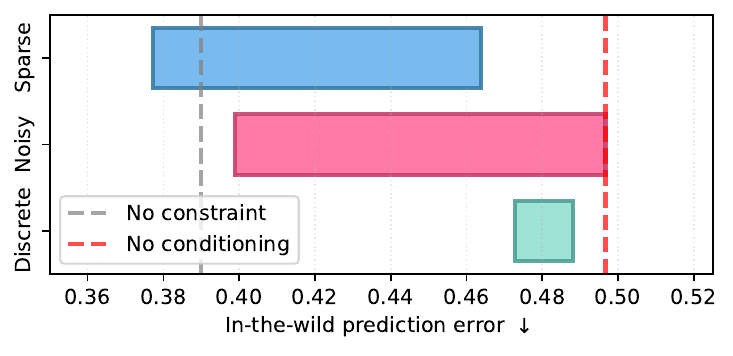}
    \caption{\textbf{IDM performance.} We report the one step prediction error on in-the-wild videos. Adjusting the capacity of sparsity and noise based latent actions allows for varying performance, while quantized ones struggle to adapt to the complexity.}
    \label{fig:train-perf}
\end{figure}

\begin{figure*}[!tbhp]
    \centering
    \includegraphics[width=\textwidth]{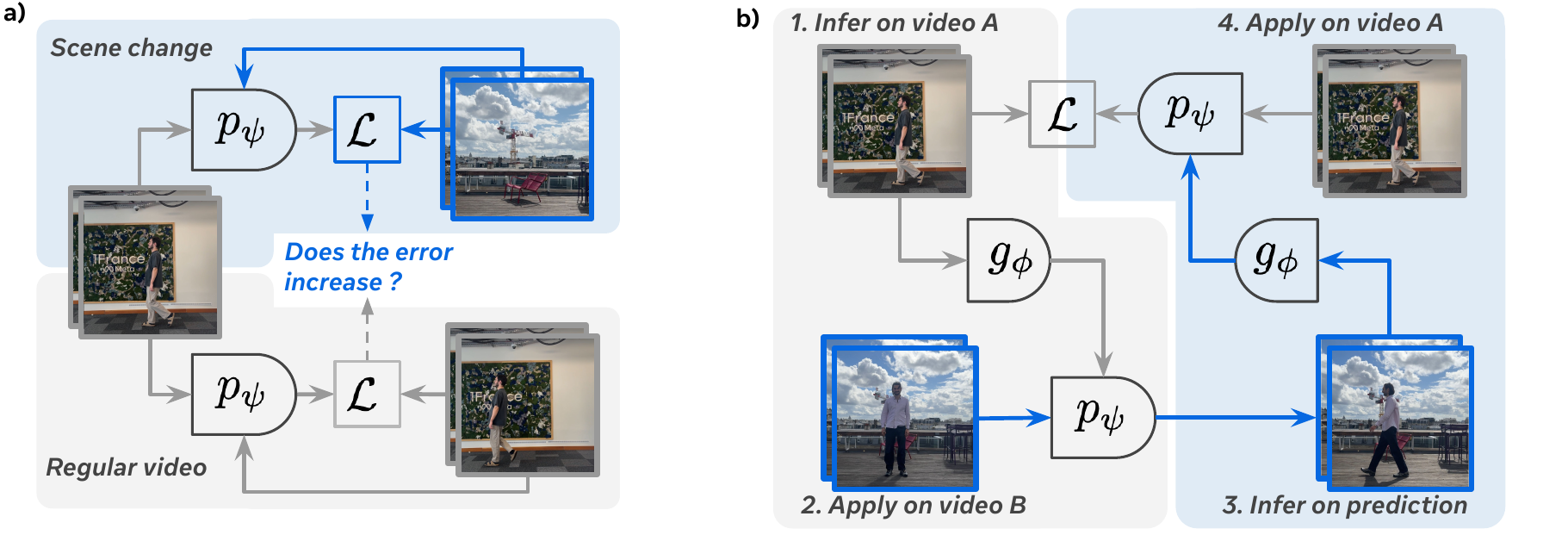}
    \caption{\textbf{Raw latent evaluation.} By artificially stitching videos, we can create abrupt scene changes. Measuring how the prediction error increases when such changes happen compared to the original video tells us how well the model can capture the whole next frame \textbf{(a)}. To measure the transferability of latent actions, we measure if they inference is cycle-consistent. We infer latent actions on video A, then apply them of another random video. From this prediction, we re-infer the latent actions and apply them on video A. If the latent action transfers well, we should obtain a small error with video A \textbf{(b)}. The combination of both metrics ensures that shortcuts are not the source of the transfer.}
    \label{fig:evals_figure}
\end{figure*}

As mentioned previously, we want to capture rich and complex actions that span a wide range of embodiments, as observed in the in-the-wild videos we consider. The first questions we thus want to answer is how different information regularization techniques adapt to this complexity?

While we measure performance in various manners through the remainder of the manuscript, focusing on different aspect and properties, we first examine the prediction quality in an ideal setting. Here we will measure the prediction error of models when unrolling a trajectory, using the inverse dynamics model (and thus the future frame) to infer the actions. This will be an upper bound of performance across all other experiments.

We will say that a regularization is "better" if it leads to a variety of achievable performance and does not saturate easily. Being able to explore a multitude of behaviors also enables us to measure the impact of latent capacity on downstream performance.  As we show in a later section, achieving the lowest prediction error using the inverse dynamics model is not always desirable, as downstream tasks require a balance between complexity and identifiability of latent actions.
As we can see in Figure~\ref{fig:train-perf}, sparse and noisy latent actions are able to achieve a range of performance between unconstrained latent actions (using the whole continuous space) and a deterministic world model. Even at maximal sparsity, we still have $d=128$ latent actions with sparsity constrains, where when the weight $\beta$ of $D_{KL}$ becomes high, noisy latent actions effectively become noise, equivalent to no conditioning. However, the vector quantization based approach struggles to scale its capacity and remains very close to the deterministic baseline.

In the rest of this work, we will talk about this "in-the-wild prediction error" as capacity of the latent actions. Since everything else in the training is identical, the drop in prediction error is attributed to the capacity of the latent actions. Lower prediction error indicates higher capacity latent actions, while a higher one indicate lower capacity latent actions.

On a more qualitative note, in Figure~\ref{fig:train-unroll} we look at a precise, relatively complex, action that exists in natural videos: someone entering and moving in a scene. We find that sparse and noisy latent actions are able to capture this action accurately, while the quantization approach shows more of a blob entering the scene. Interestingly, the exact shirt color is not captured in the latent action, highlighting that it captures a more abstract information than the exact pixels changing. Confer Supplementary Section~\ref{sec:more_idm} for additional visualizations.

\begin{tcolorbox}[colback=metabg, colframe=metafg, title=Takeaway]

A vector quantization based approach struggles to capture complex actions. Noisy or sparse latent actions are able to capture more complex actions when given the capacity.
\end{tcolorbox}

\section{What kind of actions do we learn ?}

While we showed an ideal setting where latent actions are inferred by the IDM, the model could simply cheat and encode the next frame in the latent action. Or we could learn latent actions that cannot be applied on another video, contrary to our goal of them being \textit{minimal explanations}. We thus study these two problems with simple and intuitive metrics. See Figure~\ref{fig:evals_figure} for illustration of the protocols.

\textbf{Future leakage.} To measure how much information about the future state is leaked in the latent actions, we can artificially generate scene changes by swapping ends of videos and measure how much the prediction error increases. If the model perfectly encodes the next frame in the latent we should not be able to see a prediction error spike, and thus this lack of spike is a necessary (but not sufficient\footnote{In this scenario, the only solution is to encode the next frame. This does not mean that in regular conditions the models would always fall back to this behavior.}) condition for a cheating model. Other metrics such as the alignment between between the latent actions from $s_{t-1}$ to $s_t$ and $s_{t+1}$ to $s_t$ have been proposed to measure the degree of leakage~\citep{yang2025como}, but the exact value remains hard to interpret as long as we don't have perfect alignment, and thus copy of the frame.

As we can see in Table~\ref{tab:leakage}, no matter the capacity of the latent actions, we find that the prediction error more than doubles compared to its baseline level. This suggests that no studied model is capable of cheating by encoding the next frame. We hypothesize that the complexity of the used dataset makes it harder for the model to learn this solution.

\begin{figure}[!t]
    \centering
    \begin{tabular}{@{}m{0.001\columnwidth} m{0.6\columnwidth}@{}}
        & \makebox[\columnwidth][s]{\hspace{.03\columnwidth}\textbf{Context}\hspace{.2\columnwidth}\textbf{Prediction}} \\
        \rotatebox{90}{\textbf{Groundtruth}} & \includegraphics[width=0.6\columnwidth]{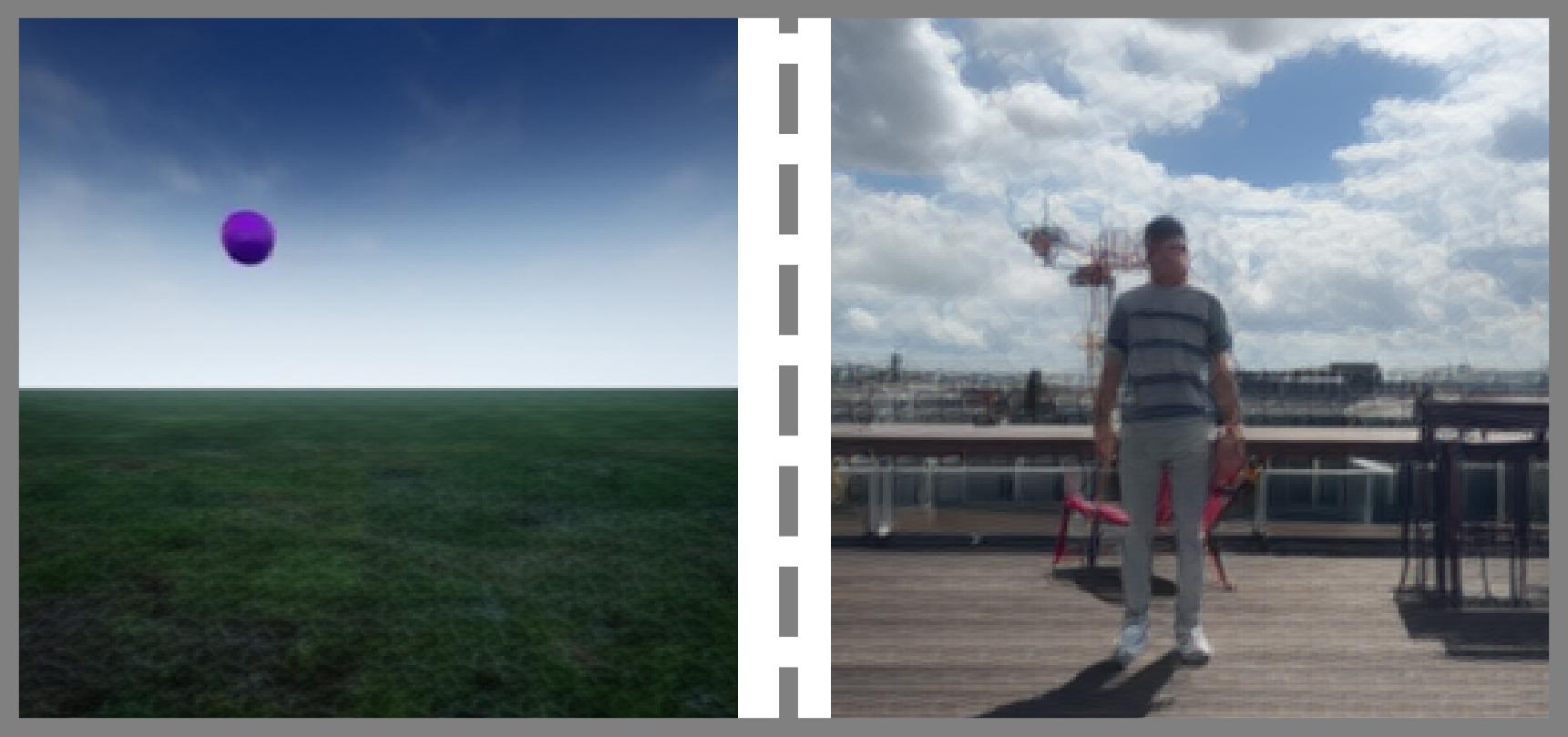} \\
        \rotatebox{90}{\textbf{High Capacity}} & \includegraphics[width=0.6\columnwidth]{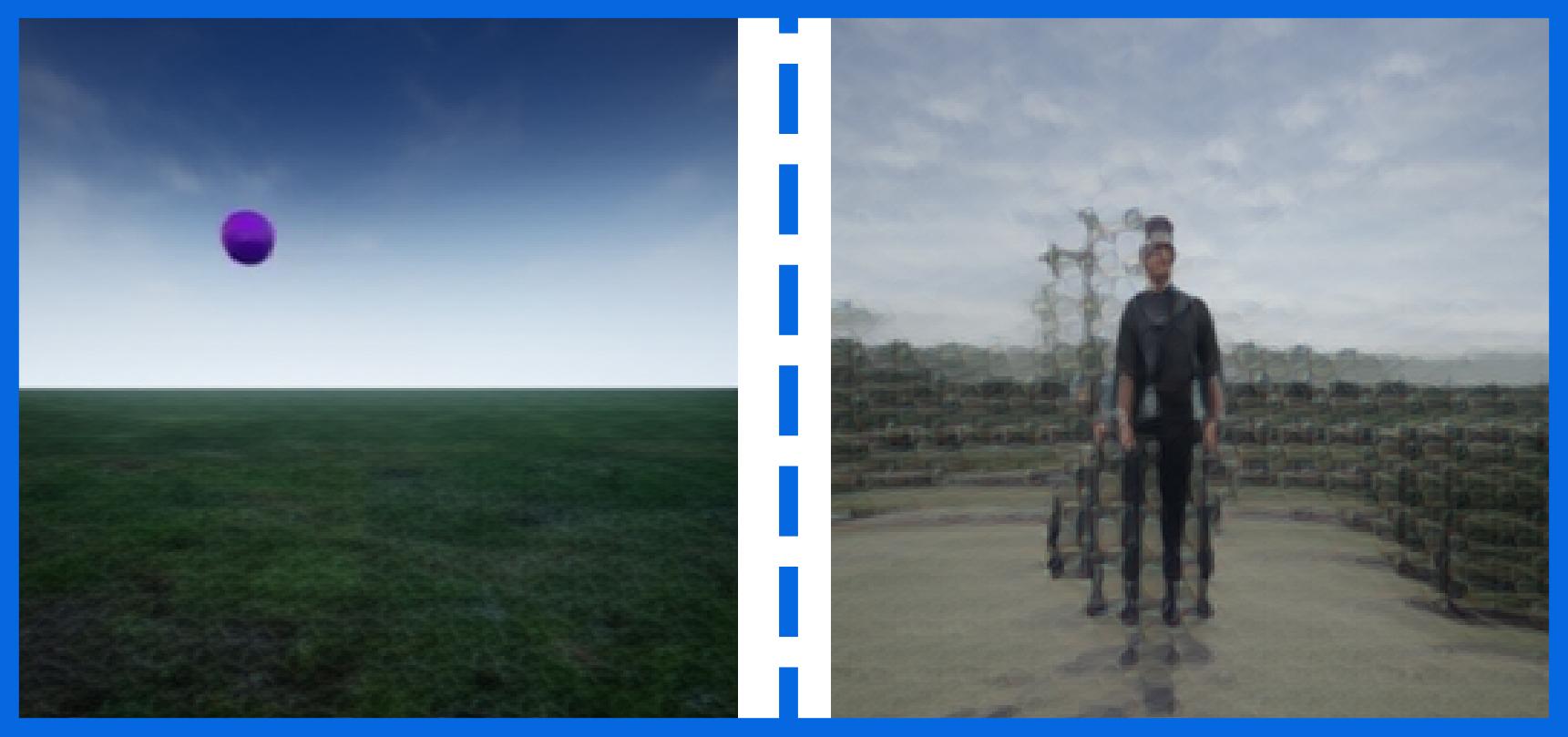} \\
        \rotatebox{90}{\textbf{Low Capacity}} & \includegraphics[width=0.6\columnwidth]{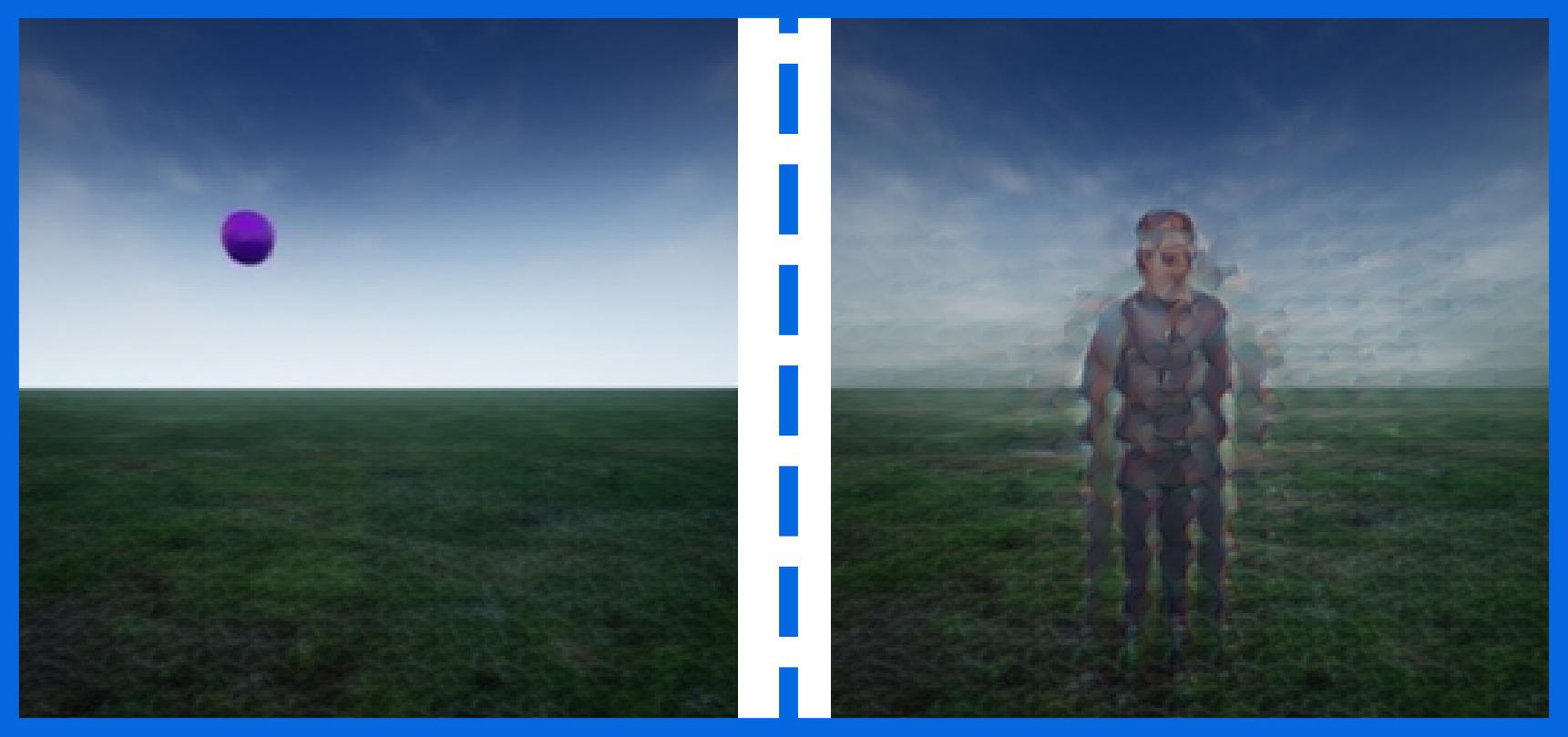} \\
    \end{tabular}

    \caption{\textbf{Future leakage.} In the presence of a scene cut, the only solution is for the latent action to encode the next frame. As capacity of the latent actions increase, more of the scene can be reconstructed, albeit with an extremely poor quality.}
    \label{fig:cheating}
    \vspace{-1em}
\end{figure}

\begin{table}[!tbhp]
    \centering
    \caption{\textbf{Prediction error increase under scene changes.} On Kinetics~\citep{kay2017kinetics}, all models exhibit a significantly higher error when a scene change occurs. This shows that the latent actions cannot simply copy the next frame. We report LPIPS values for ease of interpretation.}
    \begin{tabular}{l c c c }
        \toprule
         Latents & Capacity & w/o change & w/ change  \\
         \midrule
         \multirow{2}{*}{Sparse} & Low & 0.28 & 0.66 \textcolor{metablue}{($\times 2.3$)} \\
         & High & 0.20 & 0.50 \textcolor{metablue}{($\times 2.4$)}\\
         \multirow{2}{*}{Noisy} & Low & 0.33 & 0.69 \textcolor{metablue}{($\times 2.1$)} \\
         & High & 0.21 & 0.54 \textcolor{metablue}{($\times 2.5$)} \\
         \multirow{2}{*}{Discrete} & Low & 0.34 & 0.69 \textcolor{metablue}{($\times 2.0$)} \\
         & High & 0.29 & 0.68 \textcolor{metablue}{($\times 2.3$)} \\
         \bottomrule
    \end{tabular}
    \label{tab:leakage}
\end{table}

\begin{figure*}[!tbhp]
    \centering
    \begin{tabular}{@{}m{0.001\textwidth} m{0.99\textwidth}@{}}
        & \makebox[\linewidth][s]{\hspace{.15\linewidth}\textbf{Context}\hspace{.4\linewidth}\textbf{Prediction}} \\
        \rotatebox{90}{\textbf{Source}} & \includegraphics[width=\linewidth]{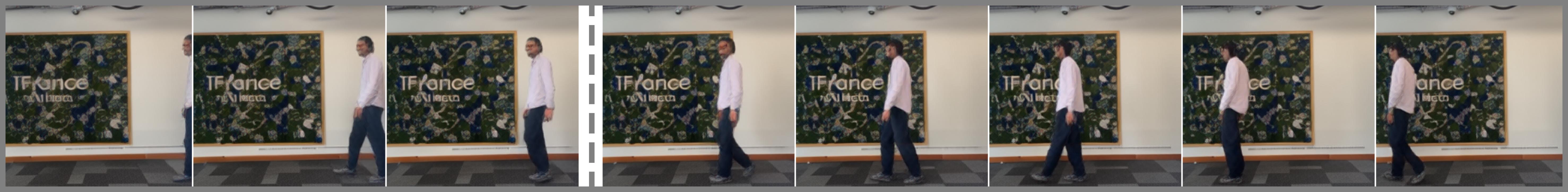} \\
        \rotatebox{90}{\textbf{1st transfer}} & \includegraphics[width=\linewidth]{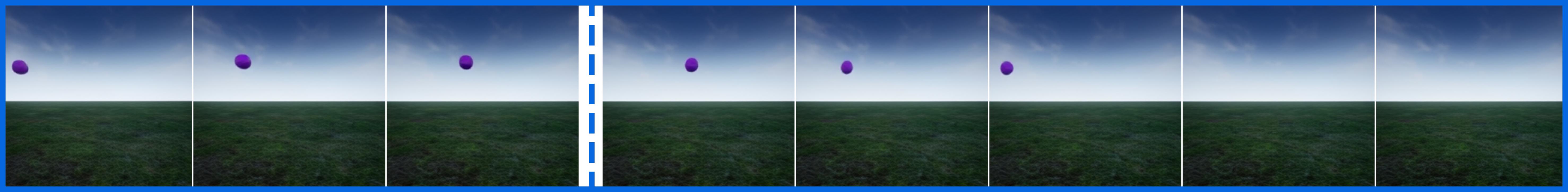} \\
        \rotatebox{90}{\textbf{2nd transfer}} & \includegraphics[width=\linewidth]{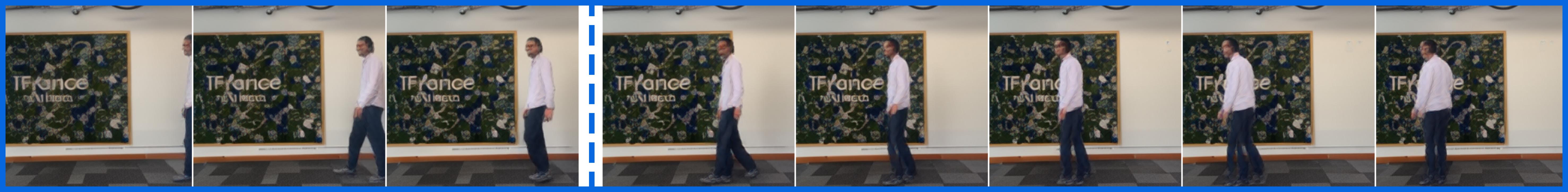} \\
    \end{tabular}
    \caption{\textbf{Transfer and cycle consistency of latent actions.} We infer latent actions from a source video, here of a man moving to the left. We then apply these actions to a flying ball, which stops its motion and also starts moving left, demonstrating transferability of latent actions. We then re-infer the latent actions and apply them to the original video. We can see the man moving to the left again, indicating that the motion was re-inferred correctly. Human videos recorded by the authors, flying ball video from~\citep{riochet_intphys_2022}.}
    \label{fig:transfer-cycle}
\end{figure*}

Visual inspection in Figure~\ref{fig:cheating} reveals that while some information about the next frame is captured in the latent actions, it is minor. However, as we study in transferability evaluations, this is not an issue in practice, and merely a consequence of having to encode objects appearing in and out of frames.

\textbf{Do latent actions transfer well ?} The next experiment to see if we have learned meaningful latent actions is if we can apply latent actions inferred on video A to video B.
Quantitatively, we evaluate the models on cycle consistency of latent actions. From random videos A and B, we infer latent actions on video A then apply them on video B. If the latent actions transfer well, we should be able to infer them again. We thus infer them again on video B and apply them on video A. By measuring the increase in prediction error on video A with the original and cyclically inferred latent actions, we can see how well latent actions transfer.
While this transfer is not well defined on random natural videos, leading to absolute gaps that are hard to interpret, this can still allow us to rank models and get an intuition about this transfer.
\begin{figure*}[!tbhp]
    \centering
    \begin{minipage}{0.47\textwidth}
        \begin{tabular}{@{}m{0.001\linewidth} m{0.99\linewidth}@{}}
            & \makebox[\linewidth][c]{\textbf{Animating the right person}} \\
            \rotatebox{90}{\textbf{Source}}       & \includegraphics[width=\linewidth]{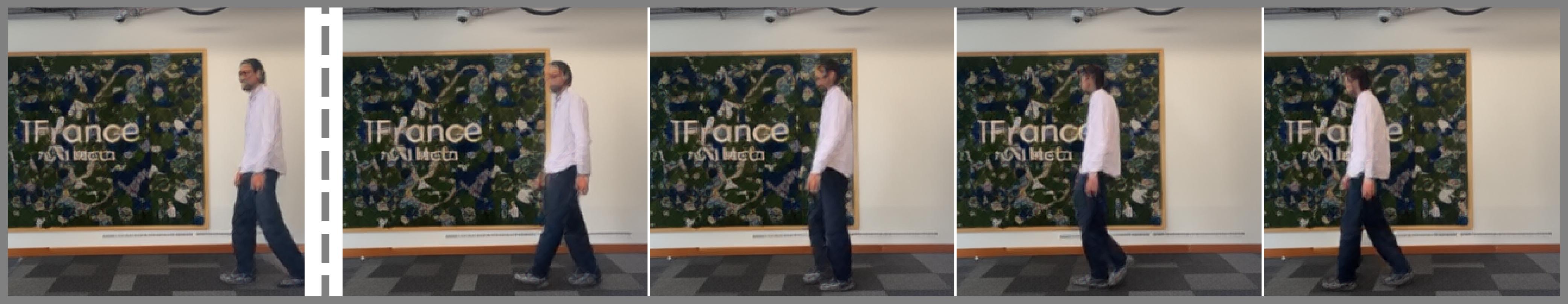} \\
            \rotatebox{90}{\textbf{Prediction}} & \includegraphics[width=\linewidth]{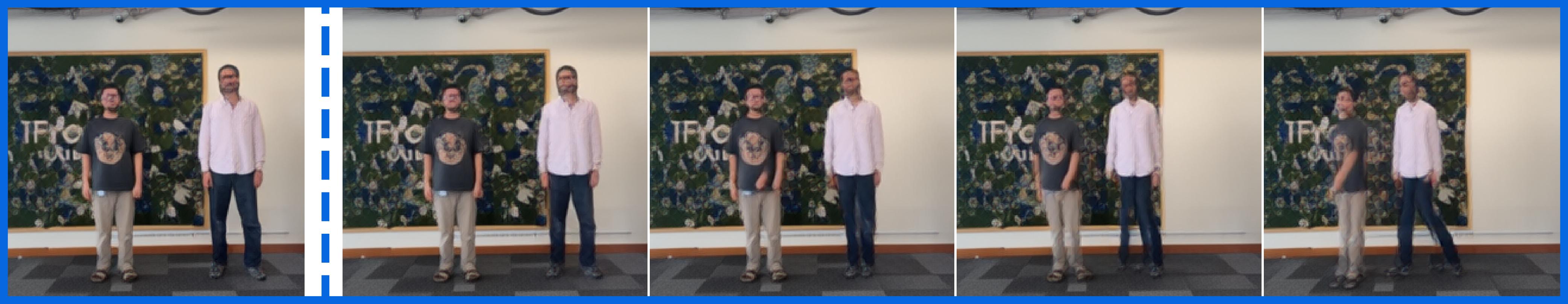} \\
        \end{tabular}
    \end{minipage}
    \hfill
    \begin{minipage}{0.47\textwidth}
        \begin{tabular}{@{}m{1\linewidth}@{}}
            \makebox[\linewidth][c]{\textbf{Animating the left person}} \\
            \includegraphics[width=\linewidth]{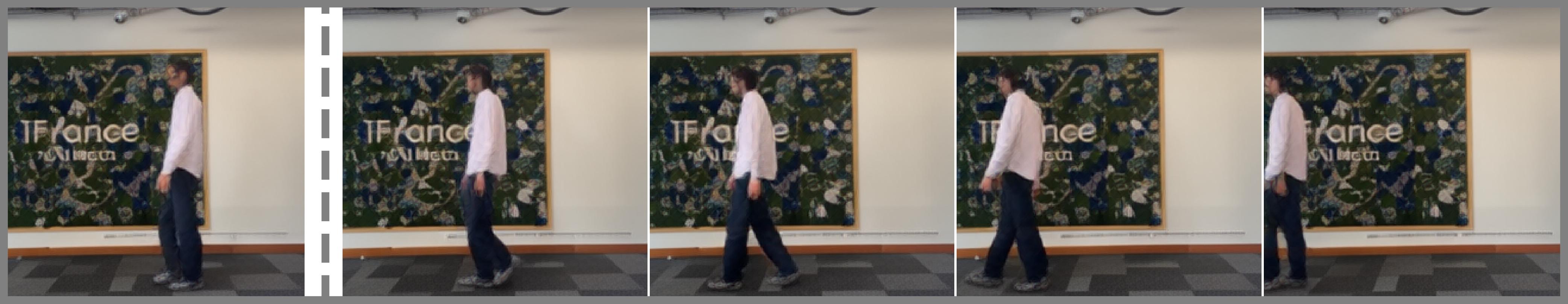} \\
            \includegraphics[width=\linewidth]{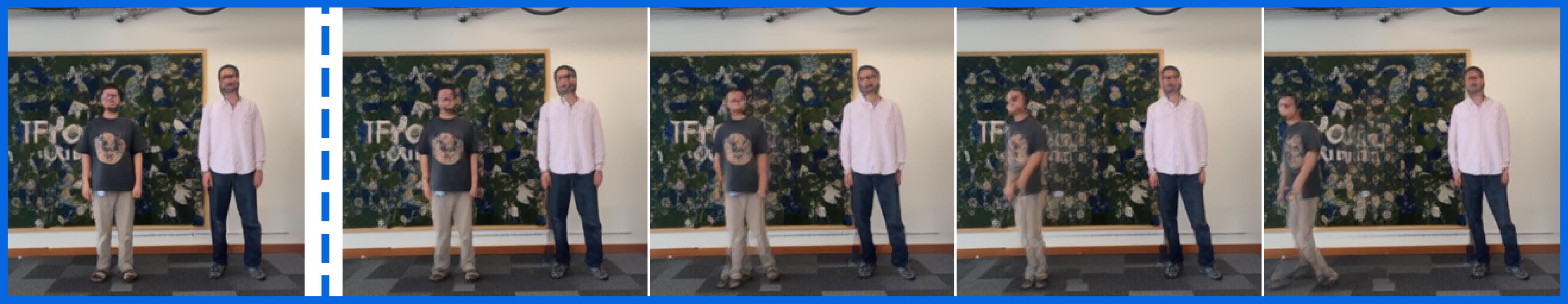} \\
        \end{tabular}
    \end{minipage}
    \caption{\textbf{Action locality.} We apply a localized locomotion action to a video with two individuals inside of it. We find that only the person closest to the walking man in the first video starts moving, indicating that the action has localized properties. We are making the individual at a given position move to the left. Videos recorded by the authors.}
    \label{fig:locality}
\end{figure*}
\begin{table}[!tbhp]
    \centering
    \caption{\textbf{Action cycle consistency.} Actions are inferred on Video 1, then applied on Video 2. Actions are again inferred and applied again on Video 1. The small increase in prediction error indicates that actions can reliably be transferred and re-inferred. We report LPIPS values over 2s prediction for ease of interpretability.}
    \resizebox{\columnwidth}{!}{%
    \begin{tabular}{l c c c c c }
        \toprule
         \multirow{2}{*}{Latents} & \multirow{2}{*}{Capacity} & \multicolumn{2}{c}{Kinetics} & \multicolumn{2}{c}{RECON}  \\
         \cmidrule(lr){3-4} \cmidrule(lr){5-6}
          & & Original & Transfer & Original & Transfer  \\
         \midrule
         \multirow{2}{*}{Sparse} & Low & 0.26 & 0.31 \textcolor{metablue}{($\times 1.20$)} & 0.24 & 0.29 \textcolor{metablue}{($\times 1.21$)} \\
         & High & 0.19 & 0.24 \textcolor{metablue}{($\times 1.30$)} & 0.20 & 0.23 \textcolor{metablue}{($\times 1.14$)}\\
         \multirow{2}{*}{Noisy} & Low & 0.30 & 0.34 \textcolor{metablue}{($\times 1.13$)} & 0.29 & 0.33 \textcolor{metablue}{($\times 1.15$)} \\
         & High & 0.20 & 0.26 \textcolor{metablue}{($\times 1.34$)} & 0.20 & 0.24 \textcolor{metablue}{($\times 1.22$)} \\
         \multirow{2}{*}{Discrete} & Low & 0.32 & 0.33 \textcolor{metablue}{($\times 1.03$)} & 0.32 & 0.33 \textcolor{metablue}{($\times 1.03$)} \\
         & High & 0.27 & 0.29 \textcolor{metablue}{($\times 1.07$)} & 0.26 & 0.27 \textcolor{metablue}{($\times 1.05$)}  \\
         \bottomrule
    \end{tabular}%
    }
    \label{tab:cycle_lpips}
\end{table}
We can see in Table~\ref{tab:cycle_lpips} that on both Kinetics~\citep{kay2017kinetics} (human activity videos) and RECON~\citep{shah2021rapid} (navigation) we only obtain a minor increase in prediction error over this latent inference cycle. While latent actions with higher capacity lead to a worse transfer, their performance remains higher after transfer than their more constrained counterparts. As shown by the previous lack of leakage of the future frame, this transfer does not stem from copying the next frame, which would be a way to obtain perfect performance.\\
\noindent The results are qualitatively investigated in Figure~\ref{fig:transfer-cycle} where we can see the movement of a man transferred to a flying ball (demonstrating transfer) and then reinferred and applied to the original video successfully. Confer Supplementary Section~\ref{sec:more_transfer} for additional visualisations.
However, such good performance even on data where we do not expect actions to transfer well such as random natural videos makes us wonder what type of actions are we learning. For this we turn to a  qualitative analysis in the next paragraph.

\textbf{Which embodiment do the latent actions learn ?} Looking at Figure~\ref{fig:locality} we can see that motion is localized, i.e. the action that is transferred is where movement occurs, and what is this movement. Due to a lack of common embodiment in natural videos, the model learns generic actions that are applied relative the to camera, the only thing common across videos.\\
This camera-relative embodiment can be a strength as we previously saw in Figure~\ref{fig:transfer-cycle}. This general abstraction allows us to transfer motion between entirely different objects, which would not be possible if motion only targeted semantically similar objects.

\begin{tcolorbox}[colback=metabg, colframe=metafg, title=Takeaway]
The absence of a clear embodiment in natural videos leads to latent actions capturing more spatially-localized, camera-relative, transformations.
\end{tcolorbox}

\section{Leveraging latent action world models for planning}

One application of a latent action space is to use it a generic interface for various embodiments.
If we are able to learn a mapping from "real" actions to latent ones, we can thus control the world model in an interpretable way. This also allows us to solve planning tasks, as we will study in this section.

\begin{figure}[!t]
    \centering
    \includegraphics[width=\columnwidth]{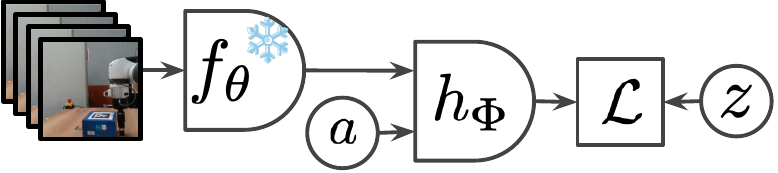}
    \caption{\textbf{Controller training.} We train a lightweight module to map known actions to latent actions. Representations of the past are used to help the prediction of the right latent actions.}
    \label{fig:controller}
\end{figure}

\textbf{Controller training.} The first part is to train a module to go from real actions --and optional representations-- to latent actions.
In the case of using actions alone we use a simple MLP, and when using actions and past representations we use a cross-attention based adapter. Confer Supplementary Section~\ref{sec:detailed_protocol} for detailed architecture and protocols.
We then simply train this controller module to predict the latent action with an L2 loss. We illustrate this process in Figure~\ref{fig:controller}.
Due to the learned latent actions being camera relative, using actions alone can be insufficient as the target latent actions will vary not only based on the action but also camera position.
In practice, we find that the controller converges to a latent action that leads to no movement when not using past representations. Confer Supplementary Section~\ref{sec:controller_qual} for visualizations.


\begin{figure*}[!tbhp]
    \centering
    \begin{minipage}{0.47\textwidth}
        \begin{tabular}{@{}m{0.001\linewidth} m{0.99\linewidth}@{}}
            & \makebox[\linewidth][c]{\textbf{DROID}} \\
            \rotatebox{90}{\textbf{Groundtruth}}       & \includegraphics[width=\linewidth]{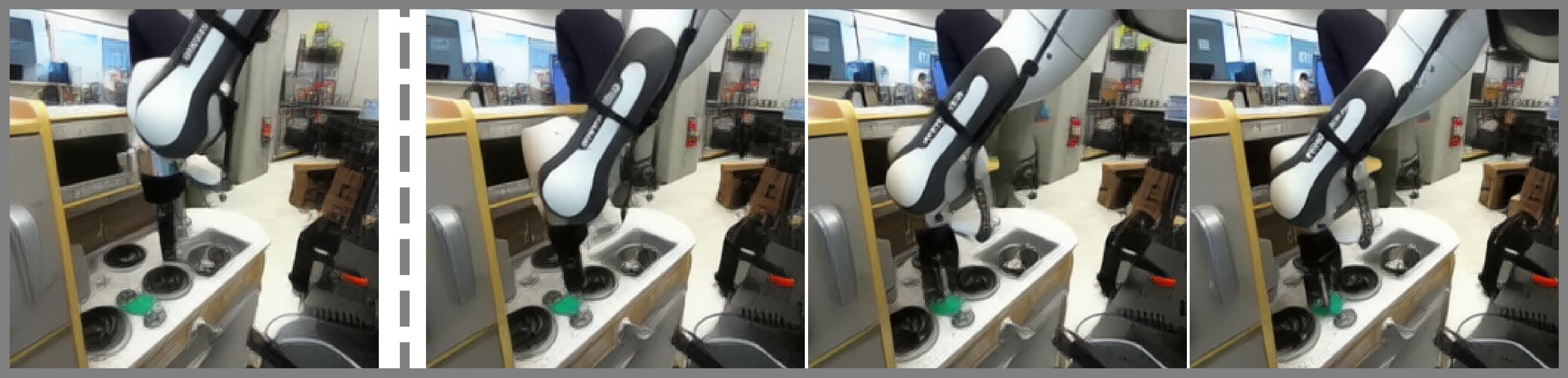} \\
            \rotatebox{90}{\textbf{IDM}} &  \includegraphics[width=\linewidth]{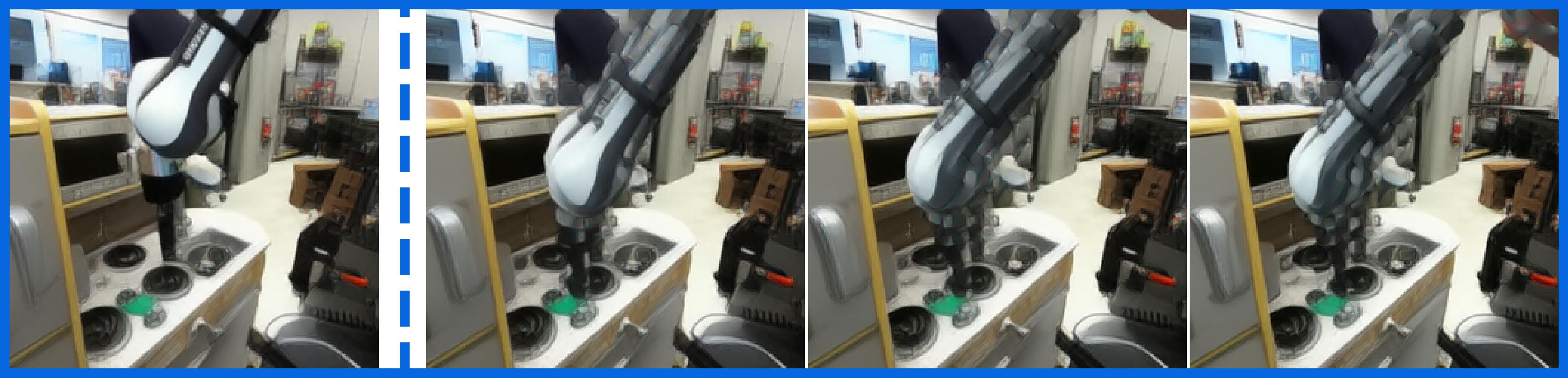} \\
            \rotatebox{90}{\textbf{Controller}} &  \includegraphics[width=\linewidth]{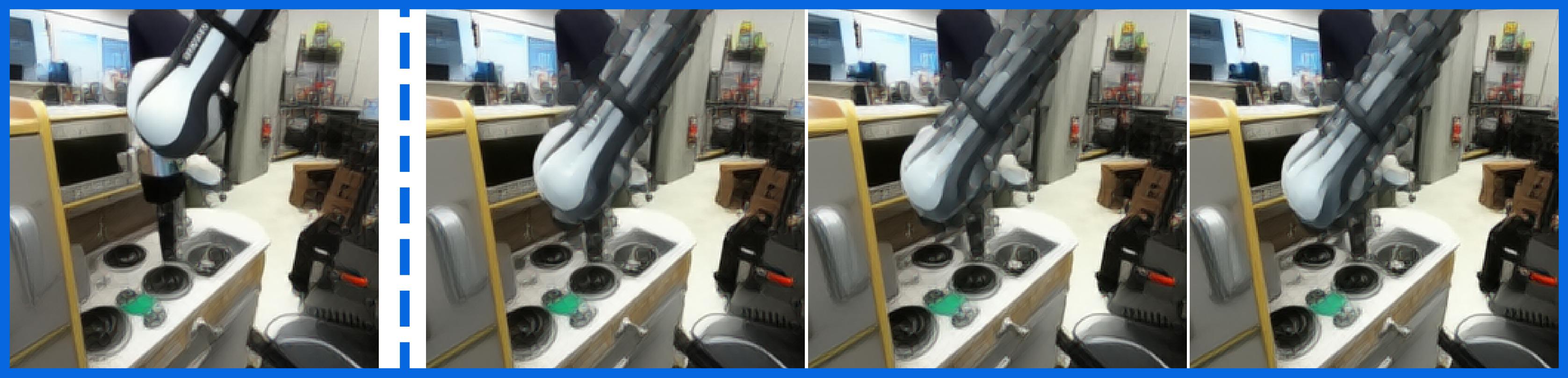} \\
        \end{tabular}
    \end{minipage}
    \hfill
    \begin{minipage}{0.47\textwidth}
        \begin{tabular}{@{}m{1\linewidth}@{}}
            \makebox[\linewidth][c]{\textbf{RECON}} \\
            \includegraphics[width=\linewidth]{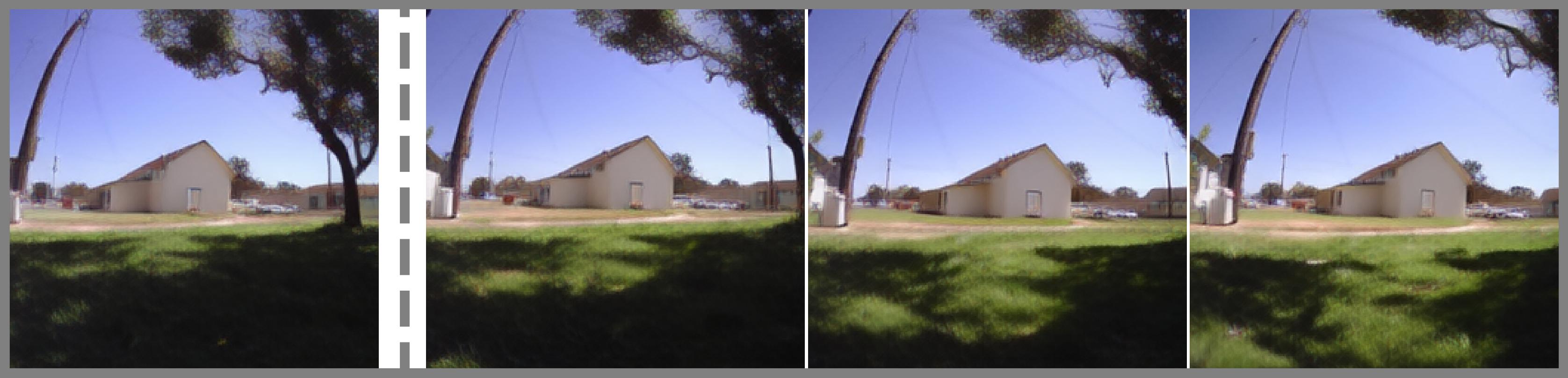} \\
            \includegraphics[width=\linewidth]{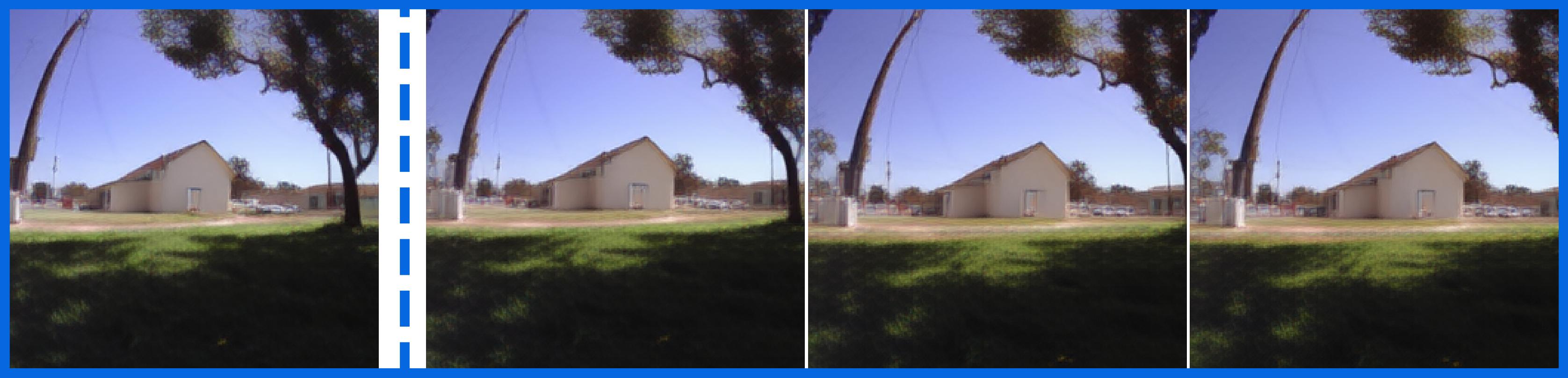} \\
            \includegraphics[width=\linewidth]{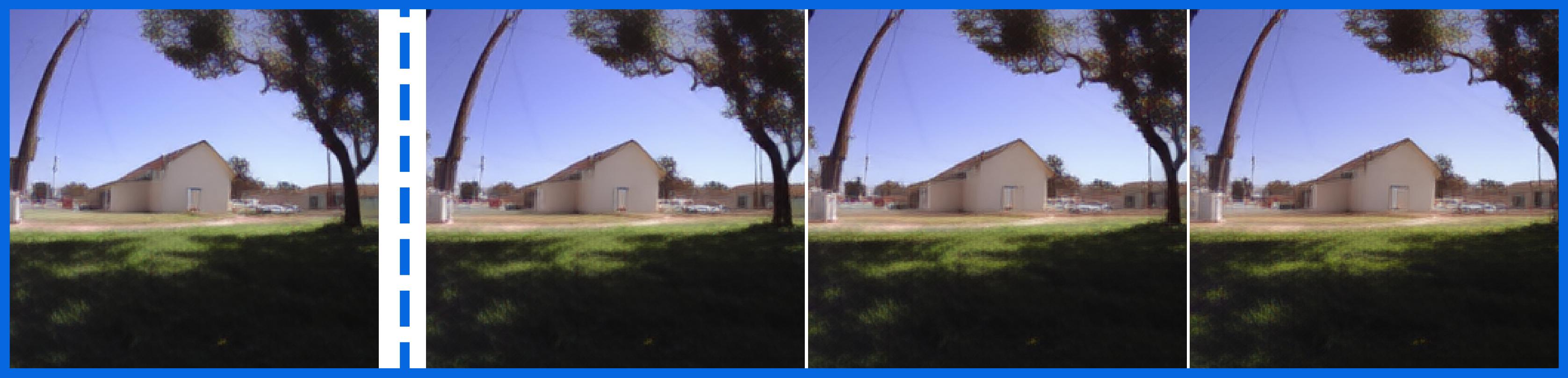} \\
        \end{tabular}
    \end{minipage}
    \caption{\textbf{Unrollings using the controller and IDM.} On both DROID and RECON, the controller is able to approximate the latent action produced by the inverse dynamics model. Movements are applied correctly over the unrolling, however physical appearance degrades over time. To produce the unrollings, frames are duplicated to map one action to one latent, something not seen during training.}
    \label{fig:rollouts}
\end{figure*}

\begin{figure*}[!tbhp]
    \centering
    \begin{tabular}{@{}m{0.001\textwidth} m{0.87\textwidth}@{}}
        \rotatebox{90}{\textbf{DROID}} &
        \makebox[\linewidth][s]{%
            \includegraphics[width=0.44\linewidth]{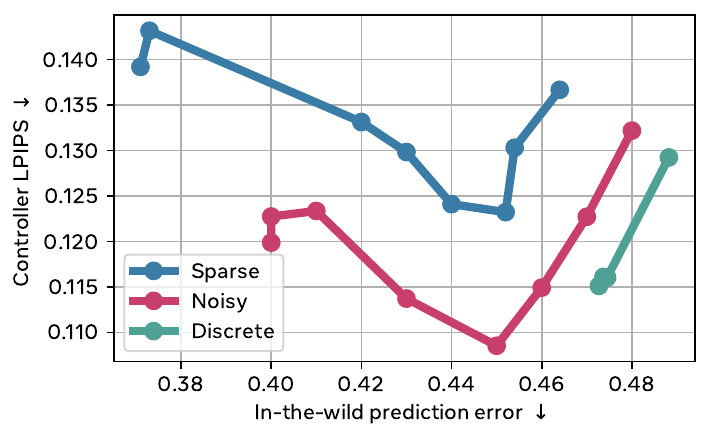}%
            \hfill%
            \includegraphics[width=0.44\linewidth]{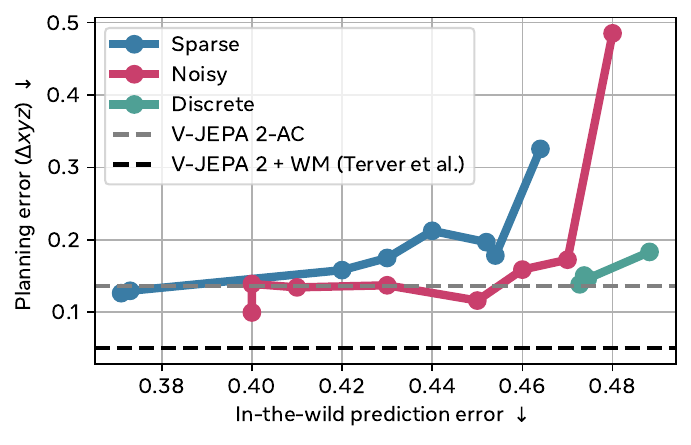}%
        } \\
        \rotatebox{90}{\textbf{RECON}} &
        \makebox[\linewidth][s]{%
            \includegraphics[width=0.44\linewidth]{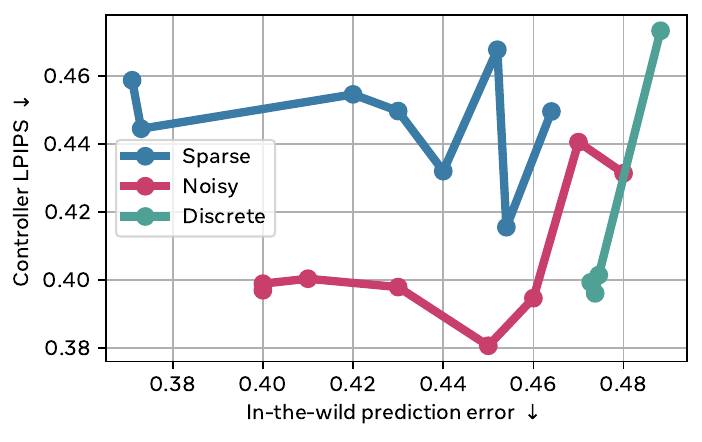}%
            \hfill%
            \includegraphics[width=0.44\linewidth]{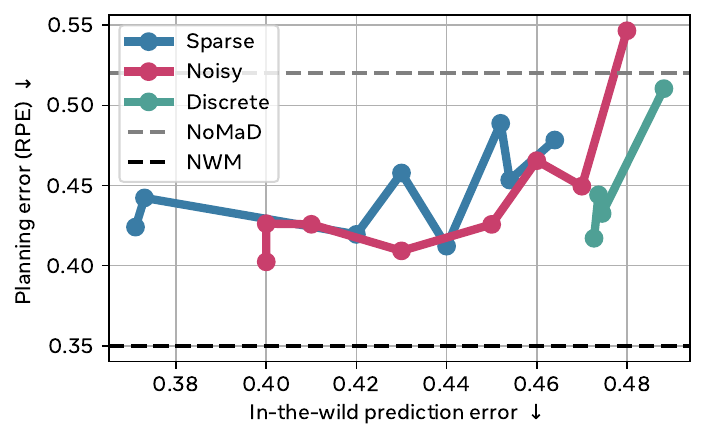}%
        } \\
    \end{tabular}
    \caption{\textbf{Controller and planning performance.} On both DROID and RECON, we are able to successfully train a model to map real to latent actions (left).
     Using these action with classical planning protocols, we are able to achieve similar performance to world model or policy baselines, that are trained with actions from the start (right).
    Overall, the best performing models are the ones where the latent actions form a middle ground in term of capacity.}
    \label{fig:controller_planning}
\end{figure*}

\textbf{Rollout quality.}
We train controllers on DROID~\citep{khazatsky2024droid}, a robotic manipulation dataset, as well as RECON~\citep{shah2021rapid}, a navigation dataset. DROID allows us to evaluate the model on data where the camera is fixed but an agent is moving inside of the scene, while NWM has still scenes but where the camera wearer is the one moving.
As we can see  qualitatively in figure~\ref{fig:rollouts} and quantitatively in the left column of figure~\ref{fig:controller_planning}, models are able to achieve quality predictions when using the controller. The predictions obtained when using the controller are very similar to the ones obtained with the IDM, with slightly more conservative actions.\\
We however find a lack of correlation between the prediction error on in-the-wild videos, i.e. the capacity of the latent actions, and the quality of the rollouts when using the controller. For both sparse and noisy latent actions, we find that using the most or least constrained setting is suboptimal, and that a more balanced regularization leads to the best predictions. This can intuitively be explained by over-constrained latent actions not containing enough information, and under-constrained ones containing too much information about the future. This is consistent with the trends observed previously, where more constrained latent actions transfer better, but freer ones can capture more fine-grained motion. Due to the simplicity of the action space here, we see that even discrete latent actions work well, supporting this choice in prior work~\citep{bu_univla_2025,schmidt_actw-without-actions_2024}.
Confer Supplementary Section~\ref{sec:detailed_planning_results} for detailed results.

\textbf{Planning performance.} We can now use our trained controllers and measure performance on goal-based planning tasks using existing protocols. Given an initial observation $s_t$ and goal observation $s_g$, we seek an action sequence that minimizes the distance between the predicted and goal states.\\
\noindent For our DROID controller, we adopt the protocol of~\cite{terver2025drives} and use a set of videos recorded in the real world on a Franka Emika Panda. We consider trajectories where the goal is to move the arm to a specific goal position. We plan at a horizon of $H{=}3$ steps using the Cross-Entropy Method (CEM)~\citep{rubinstein1997cem} and compare ourselves to the performance of V-JEPA 2-AC which is trained in a similar way as our model but using known actions, as well as the best model based on V-JEPA 2 from~\cite{terver2025drives} to upper bound the performance. To measure performance, we use the distance to the goal ($\Delta xyz$) which can be easily computed thanks to the compositionality of translations. Confer Supplementary Section~\ref{sec:detailed_protocol} for the detailed protocol.
While performance remains lower than specifically designed models, our models are able to achieve similar performance to V-JEPA 2-AC, demonstrating that our learned latent actions can effectively be used as an interface for planning tasks.
Here, the higher capacity latent actions, even though they may produce worse rollouts, can lead to the best planning performance. Notably, noisy latent actions obtain the best planning performance when the unrollings are the worst, relatively speaking. We explore the impact of adding domain specific data in our pipeline in Supplementary Section~\ref{sec:droid_vs_ytb}.\\
On a navigation task, using our controller trained on RECON, we follow the protocol of NWM~\citep{bar2024navigation} and evaluate performance using CEM for planning. We rely on the Relative Pose Error (RPE)~\citep{sturm2012evaluating} between planned and groundtruth trajectories as our main metric.
We find similar conclusions here, with models able to achieve performance that while not on par with NWM, are able to beat policy based baselines such as NoMaD~\citep{sridhar2024nomad}.
Egocentric navigation has the added difficulty of additional information entering the frame at every prediction step, making it harder to produce clean unrollings and lowering performance.
For more detailed planning results, confer Supplementary Section~\ref{sec:detailed_planning_results}.\\
Nonetheless, we find that the quality of the unrolling is not perfectly correlated with planning performance. This is a common challenge in the world model literature\citep{zhang2025world}. Overall, we find that our models trained only on in-the-wild videos learn latent action spaces that can effectively be reused to solve simple planning problems, with noisy latent actions being the best.

\begin{tcolorbox}[colback=metabg, colframe=metafg, title=Takeaway]
Latent actions learned solely on natural videos can be leveraged to solve planning tasks with similar performance as models having access to domain specific data with labeled actions.
\end{tcolorbox}

\begin{figure*}[!tbhp]
    \centering
    \begin{tabular}{@{}m{0.001\textwidth} m{0.99\textwidth}@{}}
        & \makebox[\linewidth][s]{\hspace{.15\linewidth}\textbf{Model\;Size}\hspace{.21\linewidth}\textbf{Training\;time}\hspace{.16\linewidth}\textbf{Training\;data\;quantity}} \\
        \rotatebox{90}{\textbf{Kinetics}} & \includegraphics[width=\linewidth]{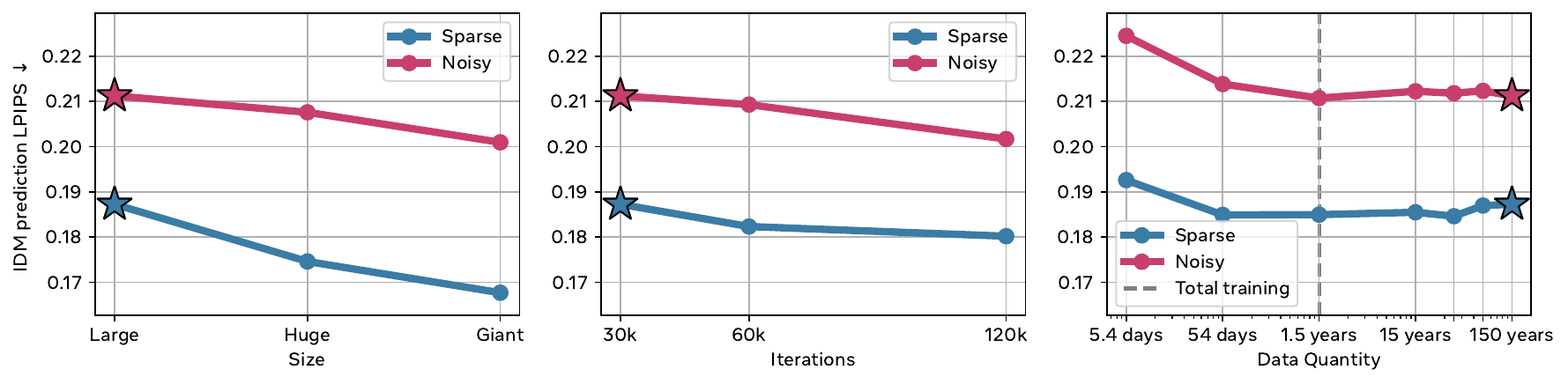} \\
        \rotatebox{90}{\textbf{DROID}} & \includegraphics[width=\linewidth]{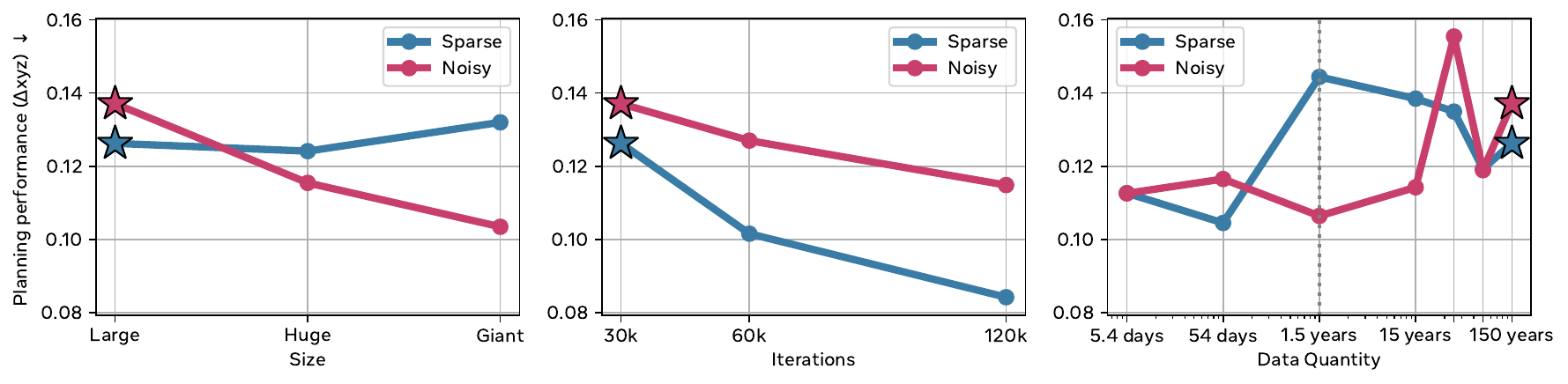} \\
    \end{tabular}
    \caption{\textbf{Scaling trends.} We investigate for two sets of latent regularizations the performance behaviour when scaling the model size (left), total training time (middle), and training data quantity (right). We find that for all axes of scaling, we are able to obtain an improved IDM on natural videos (top row). We see that when measuring performance on planning tasks we obtain similar trends, with the clearest improvements obtained by training longer. (bottom row). For data scaling, we note that our usual recipe sees on average every video twice, but we only see a total of 1\% of the total number of frames. This latter number is when we start to see degraded performance due to a too small training set. Stars indicate our default setup in the rest of the paper.}
    \label{fig:scaling}
\end{figure*}


\section{Scaling models and data.}

In this section we investigate how the performance of the models scales as we increase data, model size, and training time.
For this study we focus on sparse (with $\lambda_{l1} = 0.01$) and noisy latent actions (with $\beta = 5\times 10^{-5}$). Looking at both allows us to study scaling trends in diverse settings. We can see in Figure~\ref{fig:scaling} that overall, as model size, training time, or training data increase, we obtain better predictions when using the IDM on natural videos. However, looking at the planning performance on DROID shows us a more nuanced story, where training times significantly improves the performance but model sizes mainly has an effect for the noisy latent actions, and training data does not show a significant trend. This nuanced story about model size is consistent with previous work~\citep{ye_lapa_2025} which also find minor increase in performance when performing scaling analyses.
These results would suggest that while scaling can improve the quality of a latent action world model by improving the quality of the latent actions and/or forward model, this may not always be visible in downstream tasks that mainly evaluate simple actions, as are often used in the literature.

\section{Limitations and future work}

\textbf{Variable latent information content.} In our work, the information constraint placed on the latent actions is based on a static coefficient. However, every video has actions of various complexity, and are even sometimes deterministic. It would thus be interesting to adjust the constraint based on the complexity of the video. While this may come at a cost on the complexity of the latent action space, it would enable better calibrated latent actions.

\textbf{Sampling and planning in latent action space.} While we studied the transfer of latent actions inferred on natural videos as well as their use as a control interface, one can wonder if we cannot exploit the latent actions directly. Using the latent actions as-is would allow us to measure their quality more accurately. This can be done by sampling latent actions and analyzing the predictions, or by performing planning in the latent action space~\citep{rybkin2018learning}.
We provide some initial analysis on these aspects in Supplementary Section~\ref{sec:sampling}, noting that most of the works is ahead for high dimensional structured latent actions.

\textbf{Shaping representations with single stage training.} Currently, the world model is trained on top of frozen representations. This representation space was not designed with prediction in mind, which can hinder the inverse dynamics training, as well as the quality of the predictions in general.
As we use similar data to the pretraining distribution of V-JEPA 2 in our work, the use of latent actions in a V-JEPA 2 pretraining could unlock single-stage encoder/world-model training. This is an exciting direction for future work.

\section{Conclusion}

This work demonstrates the feasibility of learning effective latent action world models (LAMs) directly from large-scale, in-the-wild natural video datasets. We successfully address the significant challenges posed by this data, including high action complexity, environmental noise, and the lack of a common embodiment.
Our study of information regularizations highlights the benefit of continuous latent actions, which are able to adapt more effectively to the complexity of actions present in natural videos. Vector quantization, although very common in practice, struggles to adapt to this scale.
By studying the leakage of future frames in the latent actions, we found that this problem is not present in practical setting, which we hypothesize is due to a combination of conditioning choice and data complexity. We further found that while higher capacity latent actions hurt transferability, latent actions were still able to be inferred and reapplied consistently.
This led to the finding that on natural videos, learned latent actions are spatially-localized relative to the camera due to the lack of a common embodiment across videos.
Qualitatively, the learned latent actions can capture complex actions, such as a person entering a scene, and can even transfer motion between different objects, such as from a human to a ball.
Most critically, we demonstrated the practical utility of this approach. By training a simple controller to map state and known actions to the learned latent actions, our world model—trained exclusively on in-the-wild, natural videos—can be controlled to solve robotic manipulation tasks. It achieves planning performance comparable to baselines trained on in-domain, action-labeled data.
Overall, our analyses and experiments demonstrate the viability and potential of training latent action models on uncurated natural videos, offering a step towards more general world models.

\section{Acknowledgments}

We would like to thank Adrien Bardes for accepting to act in videos used for qualitative results, as well as for fruitful discussions. We also thank Amir Bar for discussions and advice on planning experiments.

\clearpage
\newpage
\bibliographystyle{assets/plainnat}
\bibliography{paper}

\clearpage
\newpage
\onecolumn
\beginappendix
\renewcommand{\thefigure}{S\arabic{figure}}
\renewcommand{\thetable}{S\arabic{table}}
\setcounter{figure}{0}
\setcounter{table}{0}

\section{Training and evaluation protocols\label{sec:detailed_protocol}}

\textbf{Decoder training.} Our decoder is trained using a ViT-L~\citep{dosovitskiy2021vit} architecture, using RoPE~\citep{su2021rope,assran2025vjepa2} positional embeddings. It reuses the architecture of the V-JEPA 2 encoder~\citep{assran2025vjepa2}, with an added linear layer to map from patch to pixels. The decoder processes the full video sequence with a frame causal attention mask to only attend to past frames.

It is trained using a combination of $L_1$ and perceptual loss~\citep{johnson2016perceptual,zhang2018unreasonable}. The decoder's weights are optimized using the Muon optimizer, with a learning rate of $0.02$, AdamW learning rate of $3\times10^{-4}$ and weight decay of $0.01$. We train the model with a batch size of 512, for 90 000 iterations, using a linear learning rate warmup for 12 000 iterations, followed by a cosine annealing.

\textbf{Latent action training.}
By default, our world model $p_\psi$ uses a ViT-L~\citep{dosovitskiy2021vit} architecture equipped with RoPE~\citep{su2021rope,assran2025vjepa2} positional embeddings. We condition $p_\psi$ on latent actions $z$ through an adapted AdaLN-zero~\citep{peebles2023scalable} mechanism that performs frame-wise conditioning, instead of the original sequence wise conditioning. Each latent action $z_t$ is represented as a 128-dimensional continuous vector. We train the world model for next frame prediction using teacher forcing~\citep{williams1989learning,vaswani2017attention} for computational efficiency.

We train on YoutubeTemporal-1B~\citep{zellers2022merlot} with batches of size $1024$ for $30\,000$ iterations. For optimization, we rely on the Muon optimizer~\citep{jordan2024muon} with a learning rate $0.02$ alongside AdamW~\citep{loshchilov2018adamw} at a learning rate of $6.25\times10^{-4}$. The learning rate schedule begins with a linear warmup for the first 10\% of training iterations, followed by cosine annealing. Weight decay is set to $0.04$. Training takes approximately 12 hours on 64 H100 GPUs.

The training loss can be defined as
\begin{equation*}
    \mathcal{L}_{t} = \| s_{t+1} - p_\psi(s_{0:t},z_t)  \|_1 + \mathcal{L}_z(z_t) \;, \text{with} \; z_t = g_\phi(s_{t},s_{t+1}),
\end{equation*}
with $p_\psi$ the world model, $s_{0:t}$ is the sequence of past representations (encoded frames), $z_t$ the latent action inferred by the inverse dynamics model $g_\phi$ from consecutive representations $s_t$ and $s_{t+1}$, and $\mathcal{L}_z$ the regularization applied to the latent action.

To determine the coefficient used for the latent action regularization terms, we perform a sweep by increasing and decreasing the coefficients regulating information content until the latent actions have the same effect as noise, until an increase in capacity does not yield a reduction in prediction error, or for vector quantization when the codebook starts to not be fully utilized.
This leads to the following coefficients:
\begin{itemize}
    \item \textbf{Sparsity}: $\lambda_{l2}=1$, $\lambda_{V}=0.1$, $\lambda_{C}=0.001$, $\lambda_{M}=0.1$,  $\lambda_{1} \in\{0.4, 0.1, 0.08, 0.06, 0.05,0.04,0.02,0.01\}$
    \item \textbf{Noisiness}: $\beta \in \{5\times10^{-3},1\times10^{-3},5\times10^{-4},1\times10^{-4},5\times10^{-5},1\times10^{-5},5\times10^{-6},1\times10^{-6}\}$
    \item \textbf{Discretization}: Commitment loss coefficient $\beta = 0.25$, $|C| \in \{ 16, 1024, 4096, 32768\}$, codebook reset for unused codes every 300k videos seen, equivalent to $2.5$ million latent actions produced
\end{itemize}

\textbf{Controller training.} Our controllers consist of 2 self-attention blocks used to process the representation of the previous frame (we only look at the ultimate previous frame $s_{t-1}$, not the whole past $s_{0:t-1}$) followed by a cross-attention block between embedded real actions, and processed representations. Actions are embedded with a 3 layer MLP to a target embedding dimension chosen as the same as the encoder (1024 by default).
The output singular token per timestep is then projected to the latent action dimension of 128 with a linear layer.

Since our latent action world models are trained with one latent action for two frames due to the video tokenization, we duplicate frames in the dataset to obtain a clear one-to-one mapping between real and latent actions.

The controller is then trained for 3000 iterations using the AdamW optimizer~\citep{loshchilov2018adamw}, with a learning rate of $1\times 10^{-3}$, a weight decay of $0.04$,$\beta_1 = 0.9$ and $\beta_2 = 0.999$. The learning rate follwos a linear warmup for 300 iterations and then a cosine decay for the rest of the training. We use a batch size of 256 with 8 frames videos at 4fps (which gives us 16 frames after duplication).

\textbf{Planning protocol for DROID.}
Our model is used for planning using the protocol of~\cite{terver2025drives}, which is as follows. Let $s_t = f_\theta(V_t)$ denote the latent visual state obtained by encoding the frame $V_t$ through the encoder $f_\theta$. Given an initial observation $s_t$ and a goal observation $s_g$, we seek an action sequence $a_{t:t+H-1} := a_t, \dots, a_{t+H-1}$ that leads from $s_t$ towards $s_g$ over a planning horizon $H$. In practice, we use $H=3$

We define the planning cost of an action sequence as
\begin{equation}
    C(s_t, a_{t:t+H-1}, s_g) = \|s_g - \hat{s}_{t+H}\|_2,
\end{equation}
where $s_g = f_\theta(V_g)$ is the encoded goal state, and the predicted latent visual states $\hat{s}$ are obtained by recursively unrolling the predictor:
\begin{equation}
    \hat{s}_{t} = f_\theta(V_t), \quad \hat{s}_{i+1} = p_\psi(\hat{s}_{i}, c(a_i, \hat{s}_i)), \quad i \in [t, t+H-1],
\end{equation}
with $c$ denoting the controller that maps actions and latent visual states to latent actions.

We use the Cross-Entropy Method (CEM)~\citep{rubinstein1997cem} to solve this optimization problem. CEM maintains a Gaussian distribution over action sequences, initialized with zero mean and unit variance. At each iteration, we sample $N=300$ candidate action sequences from the current distribution, evaluate their costs using the world model, and refit the distribution to the top $K=10$ elite samples. We perform $I=15$ iterations of this procedure and select the first action of the best sequence for execution.

To evaluate planning performance, we run 64 independent episodes. For each episode, we randomly select one video from 16 validation videos and randomly sample a clip of $H+1=4$ frames at 4 fps (matching training conditions). We then defined our error as the distance to the goal, defined as the $L_1$ distance between the cumulative planned actions and the cumulative groundtruth actions from the dataset:
\begin{equation}
    \Delta xyz = \left\| \sum_{i=t}^{t+H-1} a_i^{\text{plan}} - \sum_{i=t}^{t+H-1} a_i^{\text{gt}} \right\|_1,
\end{equation}
where $a_i^{\text{plan}}$ denotes the planned action at timestep $i$ and $a_i^{\text{gt}}$ the corresponding groundtruth action leading from $s_t$ to $s_g$. This metric measures the difference in total displacement between the planned and groundtruth trajectories, which is well-suited for actions that are additive in time, since multiple (inifinitely many) paths can lead to the target. We report the error averaged across all 64 episodes.

\textbf{Planning protocol for RECON.}
We use a similar protocol as for DROID, following the exact one used by NWM~\citep{bar2024navigation} which we recall for clarity. For additional details, confer~\cite{bar2024navigation}. Here for the Cross Entropy Method, we use $N=120$ candidate actions and only a singular iteration, which was found to be sufficient in NWM.

For efficiency, trajectories are assumed as a straight line, which allows us to plan only a single action that can be divided in the right number of time-steps. The planning horizon is here $H=8$ which at 4fps represents 2 seconds in the future. 

Once the trajectory is planned, we can compute the Absolute Trajectory Error (ATE) and Relative Pose Error (RPE)~\citep{bar2024navigation,sturm2012evaluating} to measure the quality of the trajectory compared to the groundtruth ones. In practice we focus on RPE in the main body of our work, but ATE results are reportes in Supplementary Section~\ref{sec:detailed_planning_results}.

\clearpage
\section{Sampling latent actions\label{sec:sampling}}

Throughout this work, latent actions have either been used as-is for transfer experiments, or as an interface to control the learned world model with interpretable actions. Performing planning directly in latent action space is, to the best of our knowledge, an open problem that can be made worse depending on the geometry of the latent action space.

Latent action sampling is the first process to elucidate, which varies based on the choice of latent action regularization. For \textbf{discrete latents}, the task is straightforward: sample from the codebook, possibly only for used codes. For \textbf{noisy, VAE-like latents}, the prior distribution $\mathcal{N}(0,1)$ can be used. However, the strength of the regularization used during training will alter how closely this prior is matched, leading to suboptimal coverage of the latent action distribution.
\textbf{Sparse latents} are perhaps the most challenging sampling-wise. Due to the definition of the latent action space being based on using an energy function, we have to resort to MCMC sampling techniques for EBMs~\citep{lecun2006tutorial}. A common approach is to leverage our knowledge of the energy function's gradient and use a sampler based on Stochastic Gradient Langevin Dynamics (SGLD)~\citep{grathwohl2020classifiersecretlyenergybased,welling2011sgld}.
The sampling can be defined:
\begin{equation}
    z_0 \sim p(z), \quad z_{t+1} = z_t - \frac{\alpha}{2} \frac{\partial E(z_i)}{\partial z_i} + \epsilon, \quad \text{with}\quad \epsilon \sim \mathcal{N}(0,\alpha).
\end{equation}

Here $p$ can be a uniform distribution over the latent action space, or a Gaussian distribution for example. Similarly to using the prior distribution for noisy latents, when training a LAM we are not necessarily minimizing properly the energy function associated to our latents, which can lead to a misalignment between sampled latents and the ones inferred in practice.

\begin{figure*}[!tbhp]
    \centering
    \begin{tabular}{@{}m{0.001\textwidth} m{0.99\textwidth}@{}}
        & \makebox[\linewidth][s]{\hspace{.12\linewidth}\textbf{Sparse}\hspace{.28\linewidth}\textbf{Noisy}\hspace{.28\linewidth}\textbf{Discrete}} \\
        \rotatebox{90}{\textbf{Low capacity}\hspace{5em}\textbf{High capacity}} & \includegraphics[width=\linewidth]{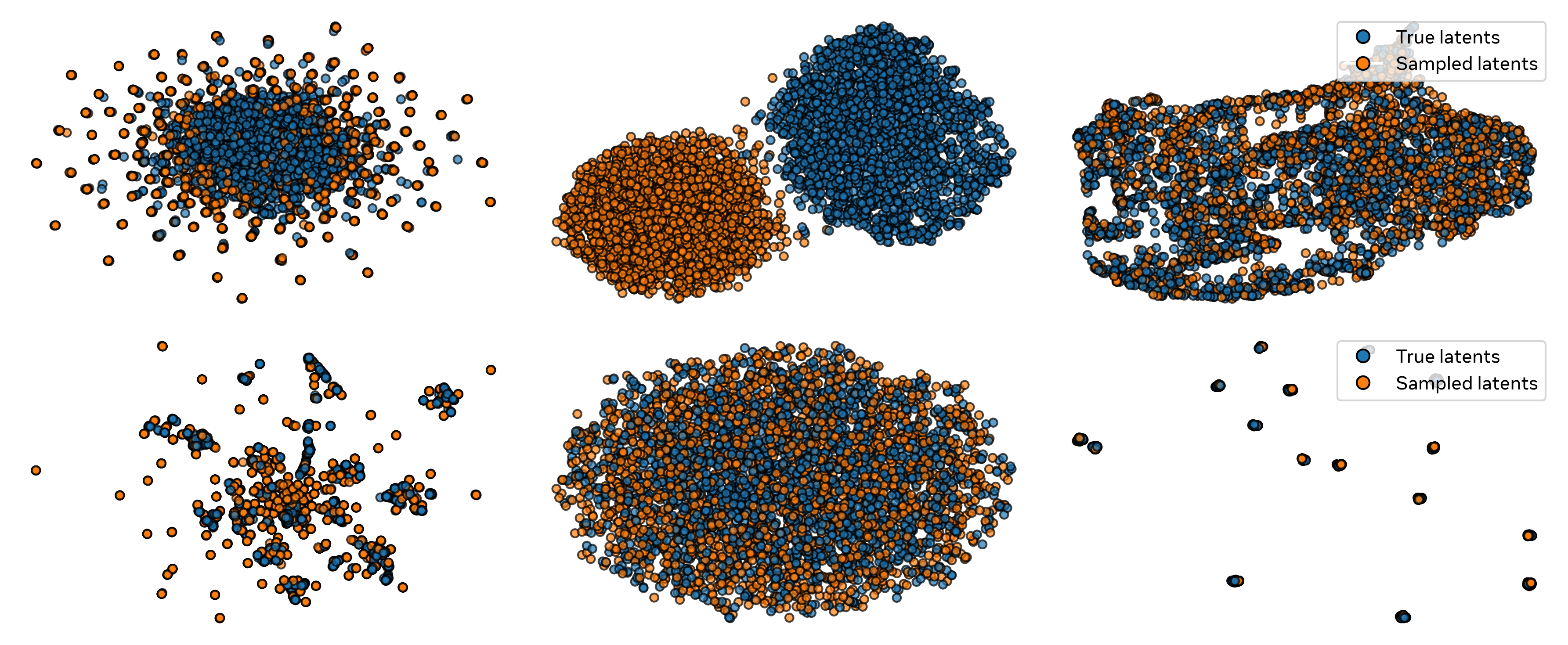} \\
    \end{tabular}
    \caption{\textbf{Sampling latent actions.} For each class of latent actions and various capacity, we infer latent on actions on natural videos and sample the same amount randomly. Looking at 2D visualizations obtained with UMAP~\citep{mcinnes2018umap}, we can see that high capacity latents (i.e. less constrained ones) are harder to sample as they are further away from the intended regularization or prior distribution. As the capacity gets lower, the visible overlap between sampled and true latents suggests that the sampling procedure works closer to intended.}
    \label{fig:sampling}
\end{figure*}

As we can see in figure~\ref{fig:sampling}, the aforementioned sampling strategies are able to sample similar latents to real ones when they have a low capacity. In that case, the models were trained with stronger constraints on the latent actions which can explain why the sampling is adequate. However when the latents are less constrained, and thus have a higher capacity, the true and sampled latents are easily separable which suggests a poor sampling.

While this analysis is purely qualitative, it effectively demonstrates how sampling approaches start to break down when handling continuous latents. An interesting angle of attack to tackle this sampling problem could be to use learning based methods that make fewer assumptions about the latent action distribution, such as diffusion models~\citep{sohl2015deep}.

\section{Detailed planning results\label{sec:detailed_planning_results}}

\begin{table*}[!htbp]
    \centering
    \caption{\textbf{Results on DROID.} We first train a controller to map actions to latent actions and measure the quality of the unrollings compared to the IDM (left).
     We then select unseen videos and infer actions based on a goal image. We measure performance as the distance to the goal (right) . }
    \label{tab:dataset_a_results}
    \begin{minipage}{0.48\textwidth}
        \centering
        \begin{tabular}{l c c c }
            \toprule
             Latents & Capacity & IDM & Controller  \\
             \midrule
             \multirow{3}{*}{Sparse} & Low & 0.12 & 0.14 \textcolor{metablue}{($\times 1.17$)}  \\
              & Mid & 0.10 & 0.12 \textcolor{metablue}{($\times 1.20$)} \\
             & High & 0.09 & 0.14 \textcolor{metablue}{($\times 1.46$)}\\
             \multirow{3}{*}{Noisy} & Low & 0.13 & 0.13 \textcolor{metablue}{($\times 1.00$)} \\
             & Mid & 0.10 & 0.11 \textcolor{metablue}{($\times 1.10$)} \\
             & High & 0.09 & 0.12 \textcolor{metablue}{($\times 1.27$)} \\
             \multirow{2}{*}{Discrete} & Low & 0.13 & 0.13 \textcolor{metablue}{($\times 1.00$)} \\
             & High & 0.11 & 0.12 \textcolor{metablue}{($\times 1.02$)}\\
             \bottomrule
        \end{tabular}
    \end{minipage}
    \hfill
    \begin{minipage}{0.48\textwidth}
        \centering
        \begin{tabular}{l c c  }
            \toprule
             Latents & Capacity & $\Delta xyz$ (m)   \\
             \midrule
             \multirow{2}{*}{Sparse} & Low & 0.33\\
             & Mid & 0.18 \\
             & High & 0.13\\
             \multirow{2}{*}{Noisy} & Low & 0.49 \\
             & Mid & 0.11\\
             & High & 0.10 \\
             \multirow{2}{*}{Discrete} & Low & 0.18 \\
             & High & 0.14\\
             \midrule
             V-JEPA 2-AC & N/A & 0.15\\
             V-JEPA 2 + WM & N/A & 0.05\\
             \bottomrule
        \end{tabular}
    \end{minipage}
\end{table*}

\begin{table*}[!htbp]
    \centering
    \caption{\textbf{Results on RECON.} We first train a controller to map actions to latent actions and measure the quality of the unrollings compared to the IDM (left).
    We then select unseen videos and infer actions based on a goal image. We measure performance as ATE and RPE (right). }
    \label{tab:dataset_b_results}
    \begin{minipage}{0.48\textwidth}
        \centering
        \begin{tabular}{l c c c }
            \toprule
             Latents & Capacity & IDM & Controller  \\
             \midrule
             \multirow{3}{*}{Sparse} & Low & 0.23 & 0.25 \textcolor{metablue}{($\times 1.11$)}  \\
              & Mid & 0.19 & 0.23 \textcolor{metablue}{($\times 1.16$)} \\
             & High & 0.17 & 0.26 \textcolor{metablue}{($\times 1.51$)} \\
             \multirow{3}{*}{Noisy} & Low & 0.24 & 0.24 \textcolor{metablue}{($\times 0.99$)} \\
             & Mid & 0.17 & 0.21 \textcolor{metablue}{($\times 1.23$)} \\
             & High &  0.17 & 0.22 \textcolor{metablue}{($\times 1.29$)} \\
             \multirow{2}{*}{Discrete} & Low & 0.24 & 0.24 \textcolor{metablue}{($\times 1.00$)} \\
             & High & 0.20 & 0.21 \textcolor{metablue}{($\times 1.06$)}\\
             \bottomrule
        \end{tabular}
    \end{minipage}
    \hfill
    \begin{minipage}{0.48\textwidth}
        \centering
        \begin{tabular}{l c c c}
            \toprule
             Latents & Capacity & ATE & RPE   \\
             \midrule
             \multirow{2}{*}{Sparse} & Low & 1.68 & 0.48\\
             & Mid & 1.45 & 0.41 \\
             & High & 1.43 & 0.42\\
             \multirow{2}{*}{Noisy} & Low & 2.06 & 0.55 \\
             & Mid & 1.49 & 0.41\\
             & High & 1.40 & 0.40 \\
             \multirow{2}{*}{Discrete} & Low & 1.81 & 0.51 \\
             & High & 1.48 & 0.42\\
             \midrule
             NoMaD & N/A & 1.93 & 0.52 \\
             NWM & N/A & 1.13 & 0.35\\
             \bottomrule
        \end{tabular}
    \end{minipage}
\end{table*}

\clearpage
\section{Robot manipulation vs in-the-wild videos\label{sec:droid_vs_ytb}}

In this section, we investigate how pretraining on DROID~\citep{khazatsky2024droid} affects performance, both on qualitative examples and on planning performance.

\begin{figure*}[tbhp]
    \centering
    \begin{tabular}{@{}m{0.001\textwidth} m{0.99\textwidth}@{}}
        & \makebox[\linewidth][s]{\hspace{.03\linewidth}\textbf{Context}\hspace{.4\linewidth}\textbf{Prediction}} \\
        \rotatebox{90}{\textbf{Groundtruth}} & \includegraphics[width=\linewidth]{figures/0_01l1/idm/entering/gt_idm.jpg} \\
        \rotatebox{90}{\textbf{Ytb1B}} & \includegraphics[width=\linewidth]{figures/0_01l1/idm/entering/pred_idm.jpg} \\
        \rotatebox{90}{\textbf{DROID}} & \includegraphics[width=\linewidth]{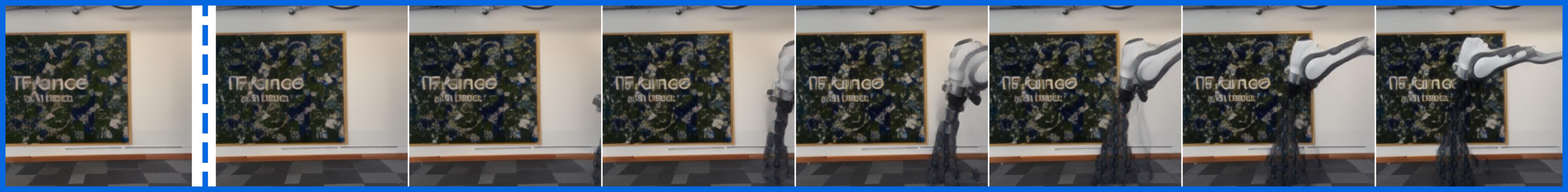} \\
    \end{tabular}
    \makebox[\linewidth][c]{\textbf{(a) Entering}}

    \begin{tabular}{@{}m{0.001\textwidth} m{0.99\textwidth}@{}}
        & \makebox[\linewidth][s]{\hspace{.03\linewidth}\textbf{Context}\hspace{.4\linewidth}\textbf{Prediction}} \\
        \rotatebox{90}{\textbf{Groundtruth}} & \includegraphics[width=\linewidth]{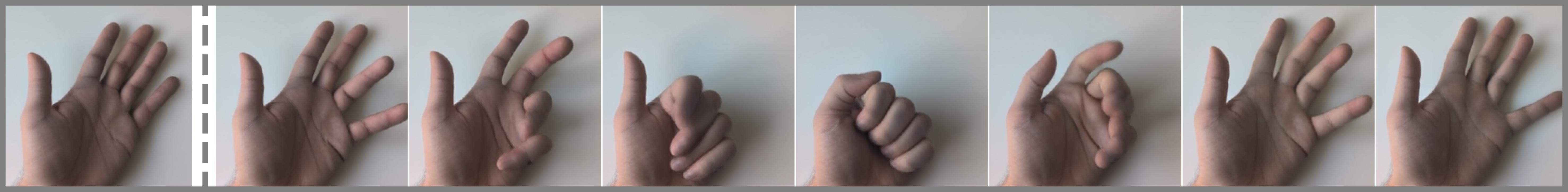} \\
        \rotatebox{90}{\textbf{Ytb1B}} & \includegraphics[width=\linewidth]{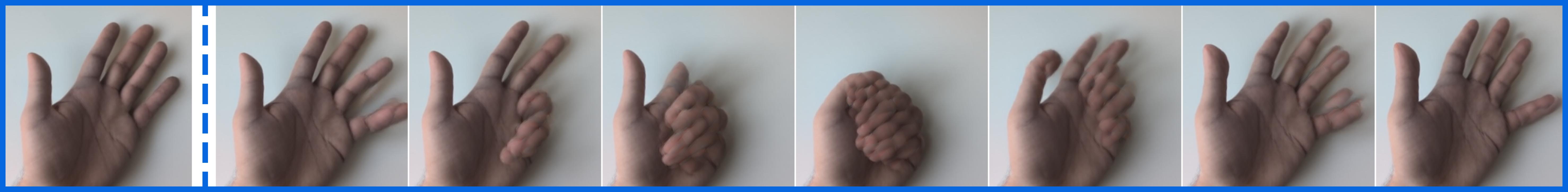} \\
        \rotatebox{90}{\textbf{DROID}} & \includegraphics[width=\linewidth]{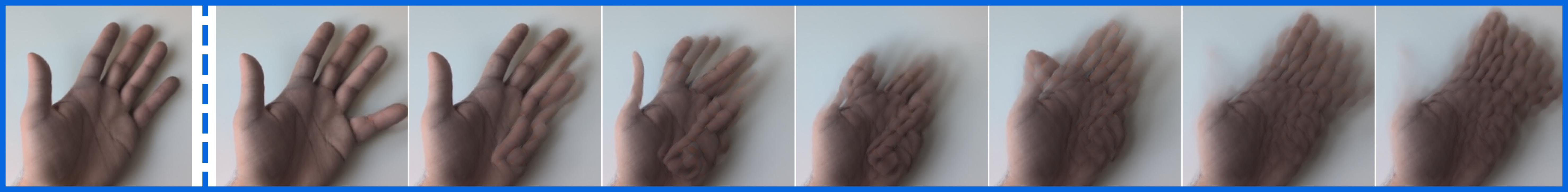} \\
    \end{tabular}
    \makebox[\linewidth][c]{\textbf{(b) Hand motion}}

    \begin{tabular}{@{}m{0.001\textwidth} m{0.99\textwidth}@{}}
        & \makebox[\linewidth][s]{\hspace{.03\linewidth}\textbf{Context}\hspace{.4\linewidth}\textbf{Prediction}} \\
        \rotatebox{90}{\textbf{Groundtruth}} & \includegraphics[width=\linewidth]{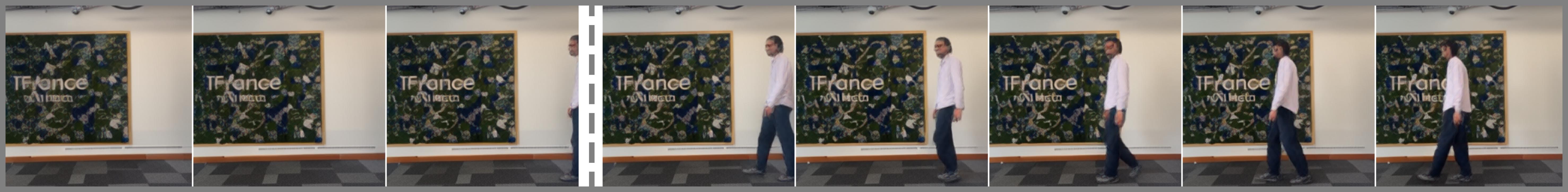} \\
        \rotatebox{90}{\textbf{Ytb1B}} & \includegraphics[width=\linewidth]{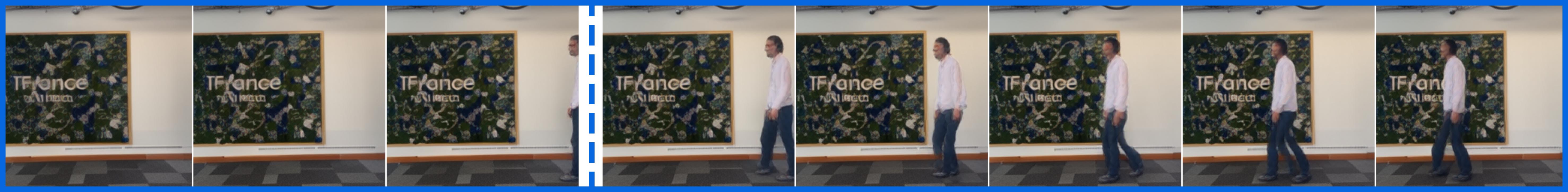} \\
        \rotatebox{90}{\textbf{DROID}} & \includegraphics[width=\linewidth]{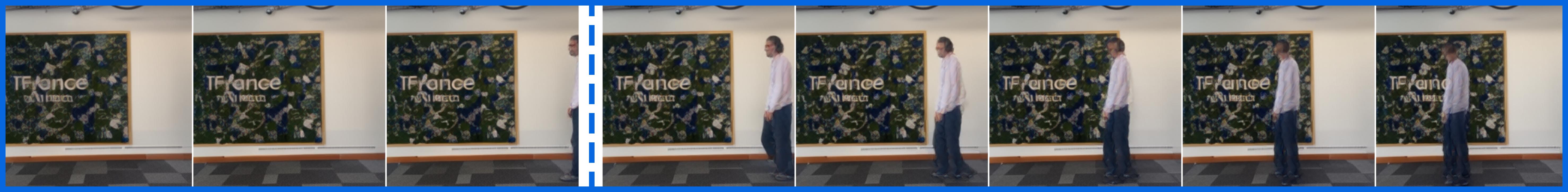} \\
    \end{tabular}
    \makebox[\linewidth][c]{\textbf{(c) Object translation}}

    \caption{\textbf{Sample predictions using the IDM across data sources.} Top: a person entering the scene; middle: hand motion; bottom: object translation. The model trained on DROID struggles on human-centric actions outside its training distribution (entering, hand), while both models can handle simple object translation.}
    \label{fig:droid_idm}
\end{figure*}

\textbf{Qualitative analysis.} We start by comparing a model trained on YoutubeTemporal-1B with one trained solely on DROID using sparse latents with $\lambda_{l1} =0.01$.
Looking at qualitative results in Figure~\ref{fig:droid_idm} on natural videos, we can see that a model trained exclusively on DROID struggles to model actions present in in-the-wild videos. This is even true in this scenario where we are using the inverse dynamics model, which thus represents an ideal upper bound of capabilities.
Interestingly, when the action corresponds to a person entering the room, we find that the model trained on DROID makes a robotic arm appear, as it is the only moving object seen during training.
While this model struggles to open and close a hand, it is however capable of animating objects that are not seen during training, such as a human walking in the scene. Looking closely we can see that the exact leg movement is not captured well, but the overall translation movement is.

these results suggest that pretraining on a more diverse dataset is beneficial to capture more diverse actions, but that even when training on a more constrained datasets, actions that still generalize can be learned. This further supports the illustration in Figure~\ref{fig:act_distrib}.

\textbf{Planning performance.}  While we have previously seen that we are able to achieve good planning performance by pretraining only on in-the-wild videos, one can wonder how much the addition of domain specific data influence performance. For this, we pretrain models with a mix of DROID and YoutubeTemporal-1B data, varying the weights of the dataset between 0 and 100\%.

\begin{table}[h]
    \centering
    \caption{\textbf{Effect of varying DROID pretraining weight on planning.} Adding in domain data helps both the quality of rollouts and planning performance. Even a minor amount of data can yield a strong boost in performance.}
    \begin{tabular}{llccccccc}
        \toprule
        Model & DROID weight & 0\% & 10\% & 25\% & 50\% & 75\% & 90\% & 100\% \\
        \midrule
        \multirow{2}{*}{Sparse} & Controller LPIPS & 0.14  & 0.14  & 0.12  & 0.11  & 0.10  & 0.10 & 0.10 \\
                                & $\Delta$ xyz     & 0.14  & 0.13  & 0.14  & 0.09  & 0.09  & 0.08 & 0.08  \\
        \midrule
        \multirow{2}{*}{Noisy} & Controller LPIPS & 0.11  & 0.10  &0.10  & 0.10 & 0.10 & 0.10 &0.9 \\
                                & $\Delta$ xyz   & 0.14  & 0.09  & 0.09  & 0.09  & 0.06  & 0.06 & 0.07 \\
        \bottomrule
    \end{tabular}

    \label{tab:droid_weight_ablation}
\end{table}

As we can see in Table~\ref{tab:droid_weight_ablation}, adding domain specific data can drastically help performance, even with as low as 10\% in some settings. What is also interesting for our latent action model setup is that by training a latent action model with domain specific data, we can achieve very similar planning performance compared to a world model trained on the same data with access to action labels (0.06 vs 0.05 for the best model from~\cite{terver2025drives}). Beyond our work, these results suggest that training a latent action model on the widest range of data possible may be optimal for a diverse set of applications.

\clearpage
\section{Qualitative Impact of regularization strength\label{sec:reg_strength_qual}}

While we previously quantified the impact of latent action capacity, equivalently regularization strength, we now turn ourselves to more qualitative analyses. Throughout this section we consider noisy latents, but similar conclusions hold across regularization families.

\begin{figure*}
    \centering
    \begin{tabular}{@{}m{0.001\textwidth} m{0.99\textwidth}@{}}
        & \makebox[\linewidth][s]{\hspace{.03\linewidth}\textbf{Context}\hspace{.4\linewidth}\textbf{Prediction}} \\
        \rotatebox{90}{\textbf{Groundtruth}} & \includegraphics[width=\linewidth]{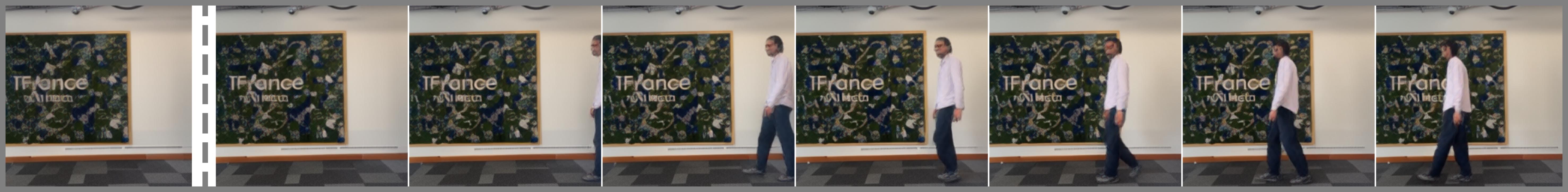} \\
        \rotatebox{90}{$\beta = 5\times 10^{-3}$} & \includegraphics[width=\linewidth]{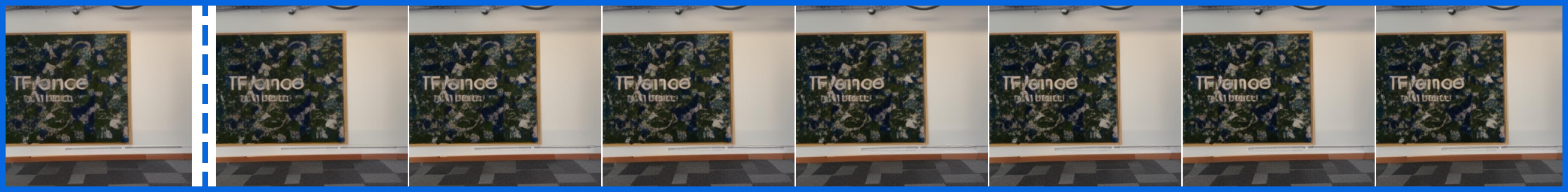} \\
        \rotatebox{90}{$\beta = 10^{-3}$} & \includegraphics[width=\linewidth]{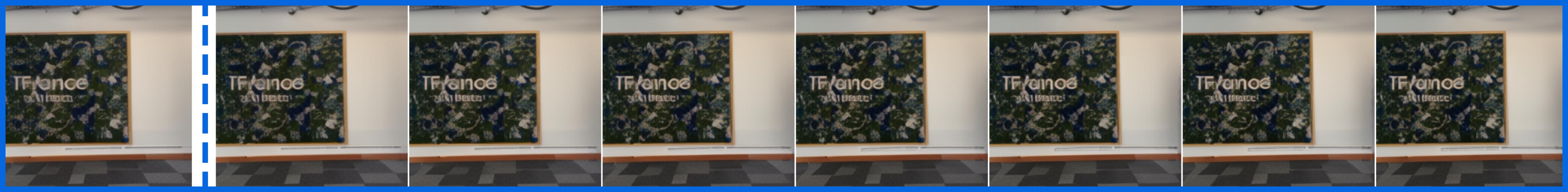} \\
        \rotatebox{90}{$\beta = 5\times 10^{-4}$} & \includegraphics[width=\linewidth]{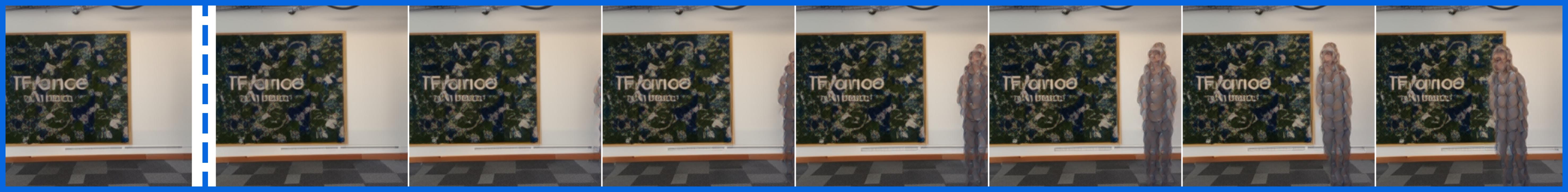} \\
        \rotatebox{90}{$\beta = 10^{-4}$} & \includegraphics[width=\linewidth]{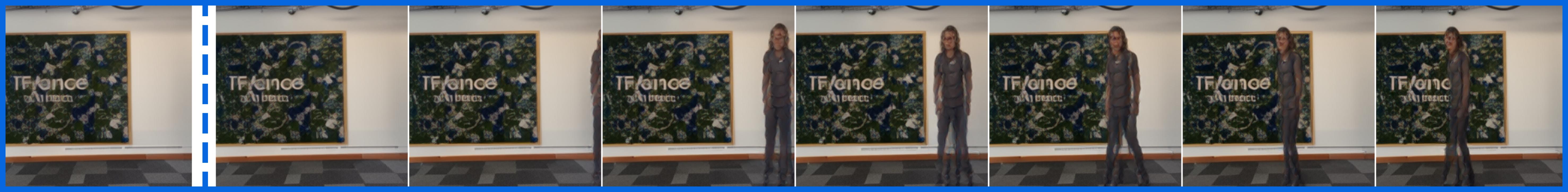} \\
        \rotatebox{90}{$\beta = 5\times 10^{-5}$} & \includegraphics[width=\linewidth]{figures/5e-5kl/idm/entering/pred_idm.jpg} \\
        \rotatebox{90}{$\beta = 10^{-5}$} & \includegraphics[width=\linewidth]{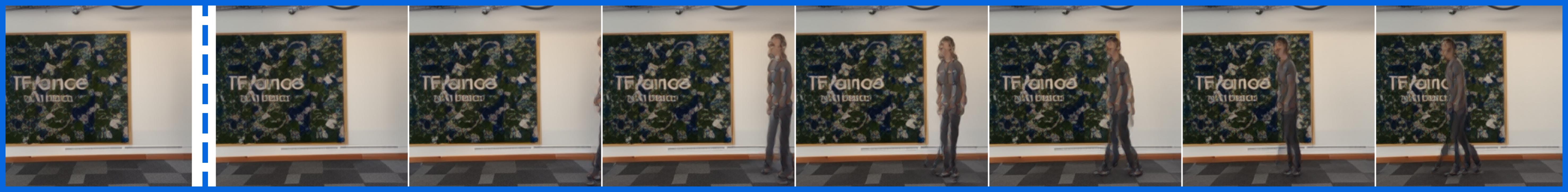} \\
        \rotatebox{90}{$\beta = 5\times 10^{-6}$} & \includegraphics[width=\linewidth]{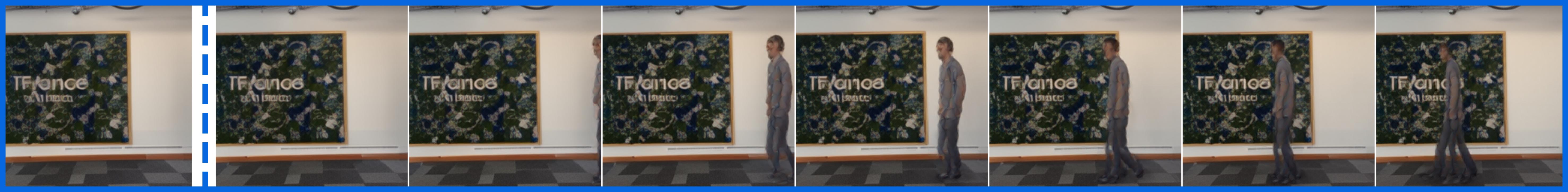} \\
        \rotatebox{90}{$\beta = 10^{-6}$} & \includegraphics[width=\linewidth]{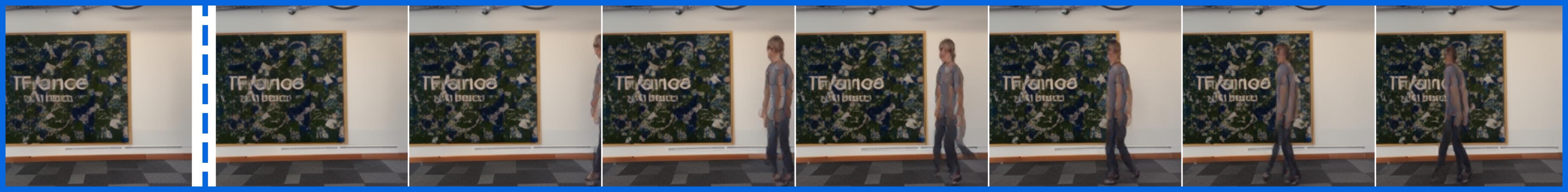} \\
    \end{tabular}
    \caption{\textbf{Quality of the IDM across regularizations.} Overly constraiend latents are not able to capture a person entering the room. As the capacity of the latent actions grows, both the quality of the person and leg movement increases, but plateaus after a certain point.}
    \label{fig:idm-reg}
\end{figure*}

As we can see in Figure~\ref{fig:idm-reg}, when latent actions are overly constrained, the model is unable to make a human appear. As the constraint gets weaker, we start to see the person appearing, albeit with suboptimal appearance and motion. Continuing to weaken this regularization, we start to see a better outline of the person, and a higher fidelity in motion, especially for the leg movements.

\begin{figure*}
    \centering
    \begin{tabular}{@{}m{0.001\textwidth} m{0.99\textwidth}@{}}
        & \makebox[\linewidth][s]{\hspace{.03\linewidth}\textbf{Context}\hspace{.4\linewidth}\textbf{Prediction}} \\
        \rotatebox{90}{\textbf{Seed}} & \includegraphics[width=\linewidth]{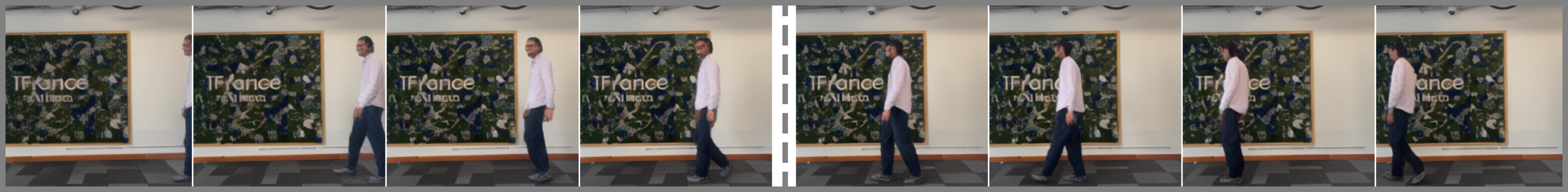} \\
        \rotatebox{90}{\textbf{Target}} & \includegraphics[width=\linewidth]{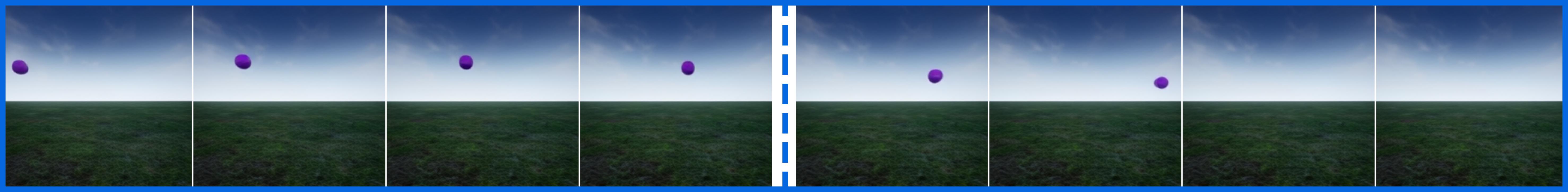} \\
        \rotatebox{90}{$\beta = 10^{-3}$} & \includegraphics[width=\linewidth]{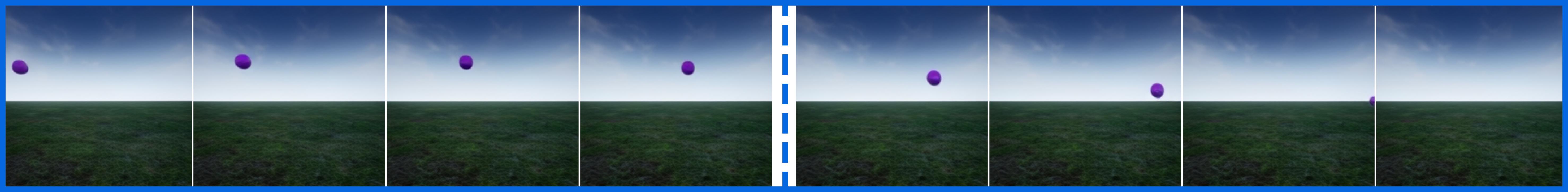} \\
        \rotatebox{90}{$\beta = 5\times 10^{-4}$} & \includegraphics[width=\linewidth]{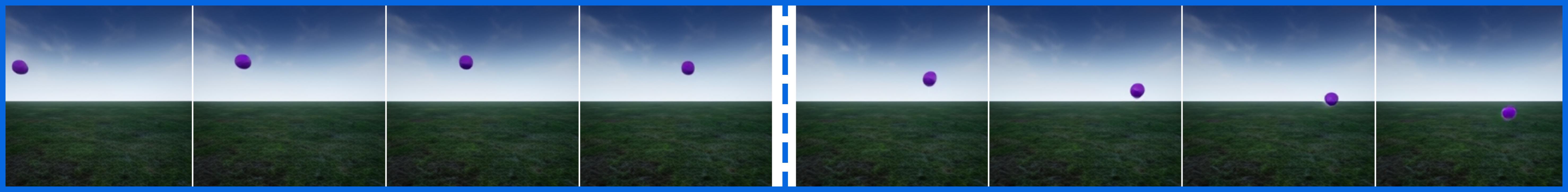} \\
        \rotatebox{90}{$\beta = 10^{-4}$} & \includegraphics[width=\linewidth]{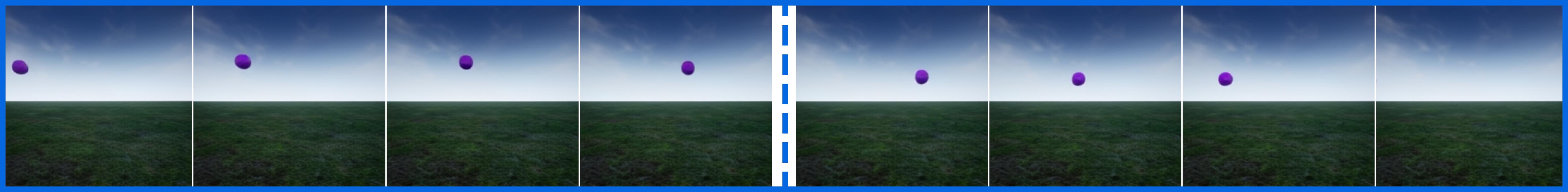} \\
        \rotatebox{90}{$\beta = 5\times 10^{-5}$} & \includegraphics[width=\linewidth]{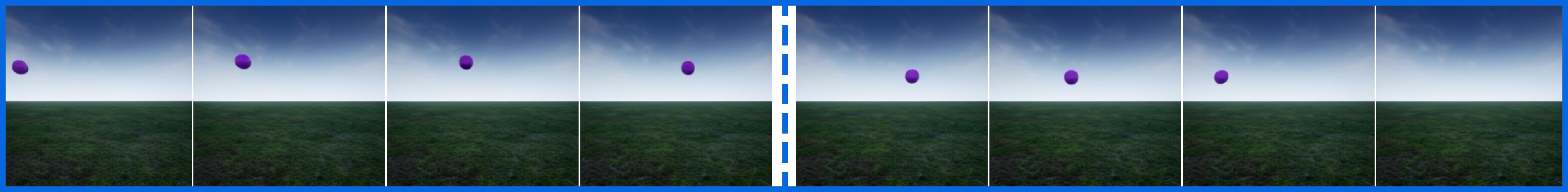} \\
        \rotatebox{90}{$\beta = 10^{-5}$} & \includegraphics[width=\linewidth]{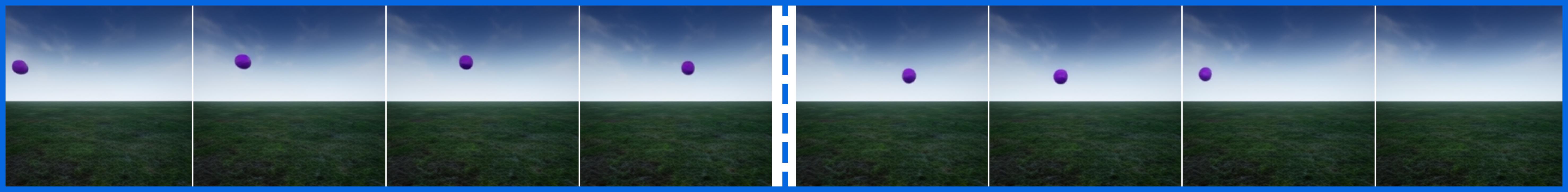} \\
        \rotatebox{90}{$\beta = 5\times 10^{-6}$} & \includegraphics[width=\linewidth]{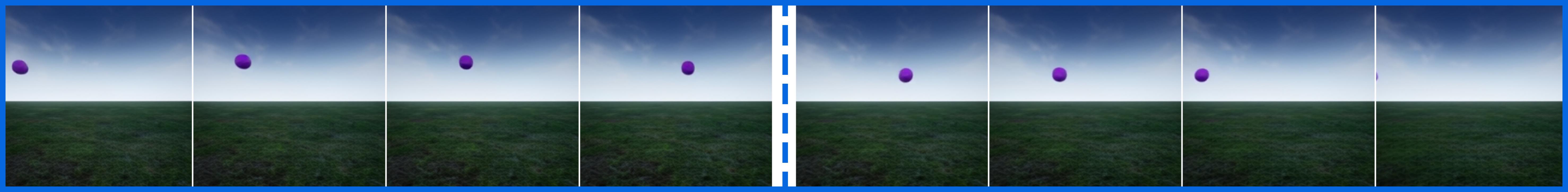} \\
        \rotatebox{90}{$\beta = 10^{-6}$} & \includegraphics[width=\linewidth]{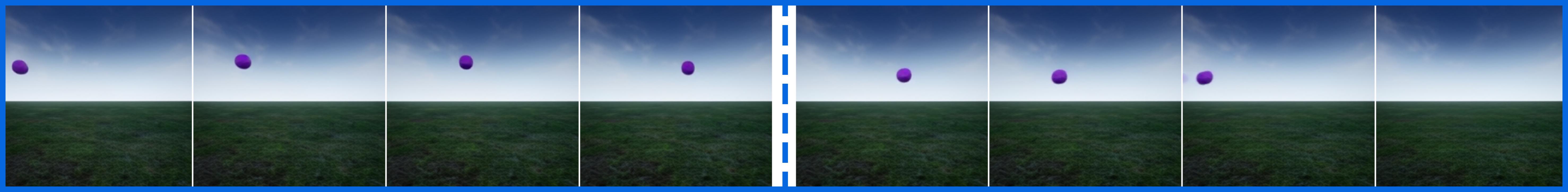} \\
    \end{tabular}
    \caption{\textbf{Cross object action transfer across regularizations.}The quality of motion transfer increases with the capacity of the latents. More constrained latents either have no effect, or a weaker one.}
    \label{fig:transfer-reg}
\end{figure*}

In Figure~\ref{fig:transfer-reg} we study the impact of the regularization strength when transferring movements from a human to a ball. We can see that with a too strong regularization, the ball simply continues its trajectory. We essentially have a deterministic world model. As the regularization increases, the ball slows down more until it perfectly follows the transferred motion. We then see it going perfectly left, in a straight line. This highlights the importance of adequate capacity to be able to identify interpretable actions.

\begin{figure*}
    \centering
    \begin{tabular}{@{}m{0.001\textwidth} m{0.99\textwidth}@{}}
        & \makebox[\linewidth][s]{\hspace{.03\linewidth}\textbf{Context}\hspace{.4\linewidth}\textbf{Prediction}} \\
        \rotatebox{90}{\textbf{Source}} & \includegraphics[width=\linewidth]{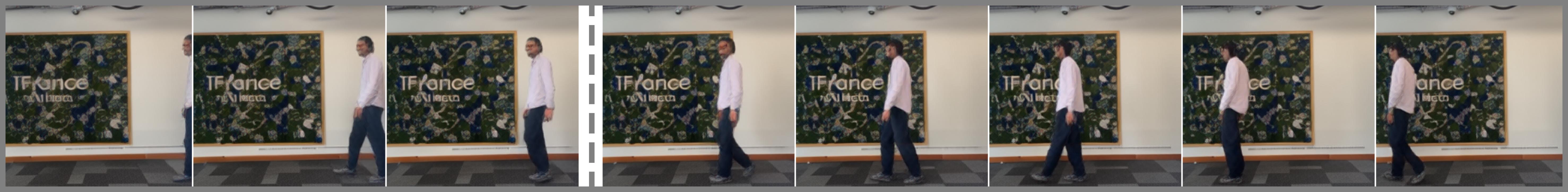} \\
        \rotatebox{90}{\textbf{1st transfer}} & \includegraphics[width=\linewidth]{figures/5e-5kl/cycle/basile_to_ball/first_cycle.jpg} \\
        \rotatebox{90}{$\beta = 5\times 10^{-3}$} & \includegraphics[width=\linewidth]{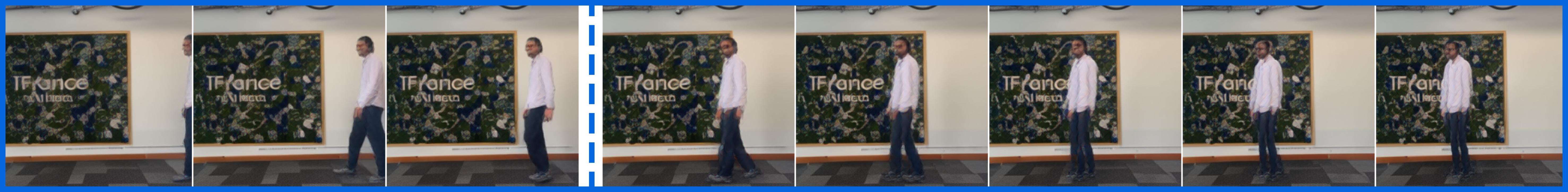} \\
        \rotatebox{90}{$\beta = 10^{-3}$} & \includegraphics[width=\linewidth]{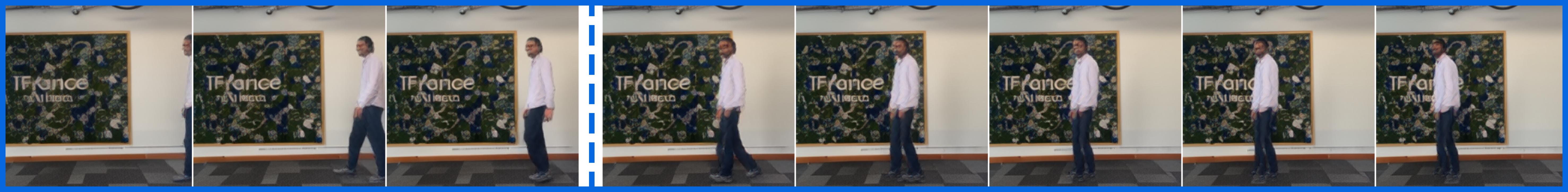} \\
        \rotatebox{90}{$\beta = 5\times 10^{-4}$} & \includegraphics[width=\linewidth]{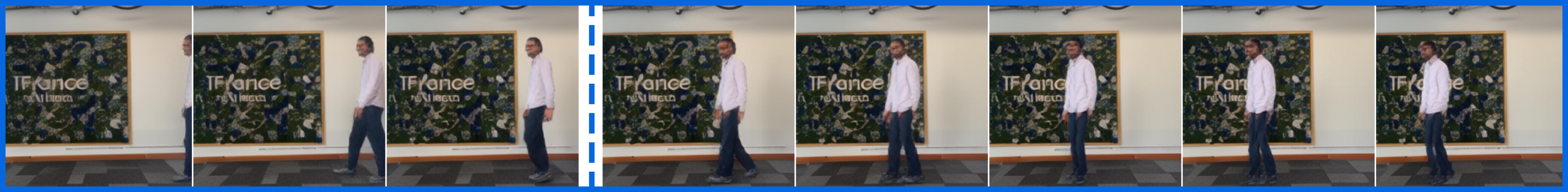} \\
        \rotatebox{90}{$\beta = 10^{-4}$} & \includegraphics[width=\linewidth]{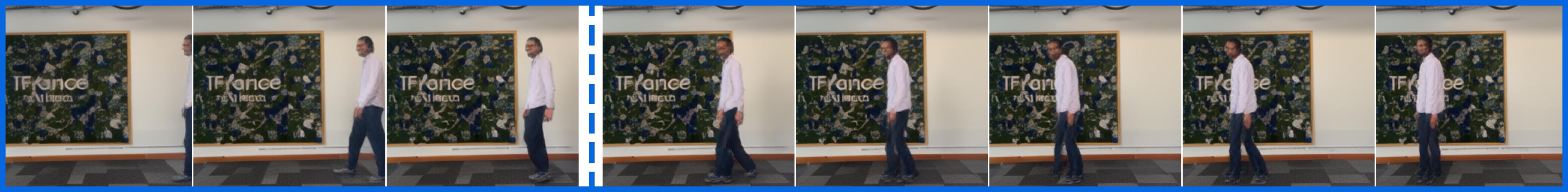} \\
        \rotatebox{90}{$\beta = 5\times 10^{-5}$} & \includegraphics[width=\linewidth]{figures/5e-5kl/cycle/basile_to_ball/second_cycle.jpg} \\
        \rotatebox{90}{$\beta = 10^{-5}$} & \includegraphics[width=\linewidth]{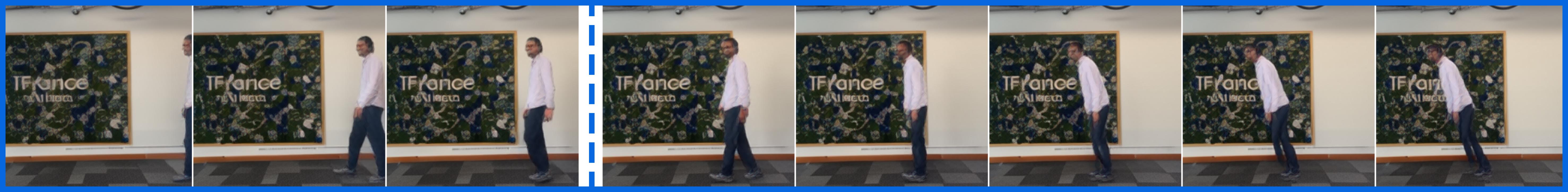} \\
        \rotatebox{90}{$\beta = 5\times 10^{-6}$} & \includegraphics[width=\linewidth]{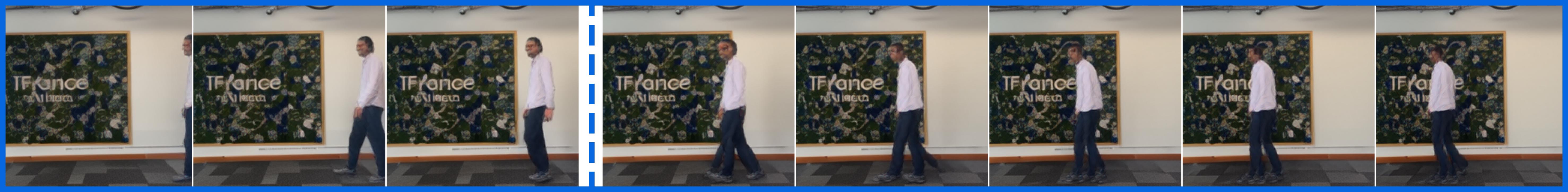} \\
        \rotatebox{90}{$\beta = 10^{-6}$} & \includegraphics[width=\linewidth]{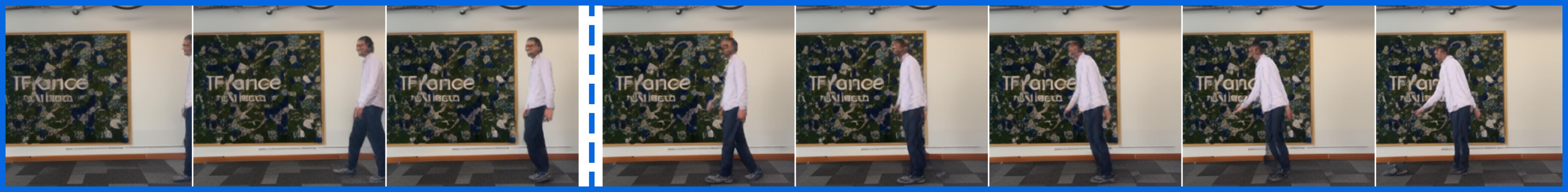} \\
    \end{tabular}
    \caption{\textbf{Cycle consistency for different regularizations.} As the latent action capacity increases we obtain improved transfer. After a certain point, the movement becomes more localized and only the upper body motion is captured back.}
    \label{fig:cycle-reg}
\end{figure*}

While so far more capacity has been beneficial, we get a better understanding of what happens at lower constraints in Figure~\ref{fig:cycle-reg}. Here we see that while initially capacity improves the cycle consistency of actions, in some cases at higher capacity the motion is not applied to the whole human when re-inferred. This suggests a greater spatial localization of actions at higher capacity. We obtain more "precise" actions, at the cost of generality. This mirrors what is observed in planning evaluations, where the optimal latent actions spaces strike a balance between capacity and generality.

\clearpage
\section{Additional IDM rollouts.\label{sec:more_idm}}

In this section we take a look at more qualitative examples of rollouts performed with the inverse dynamics models. This allows us to establish an upper bound of the performance attainable by a given model, with the caveat that models may use shortcut solutions. Similar to figure~\ref{fig:train-unroll}, we take a look at the least constrained latents for all regularizations. We focus on videos from SSv2~\citep{goyal2017something} as a natural video dataset that are not seen during training.

\begin{figure*}[tbhp]
    \centering
    \begin{tabular}{@{}m{0.001\textwidth} m{0.99\textwidth}@{}}
        & \makebox[\linewidth][s]{\hspace{.03\linewidth}\textbf{Context}\hspace{.4\linewidth}\textbf{Prediction}} \\
        \rotatebox{90}{\textbf{Groundtruth}} & \includegraphics[width=\linewidth]{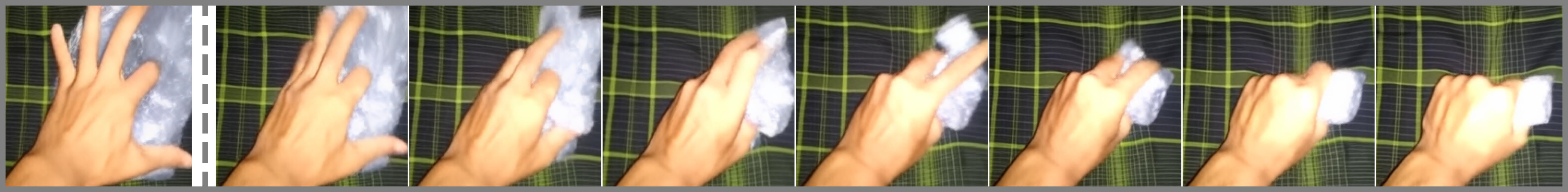} \\ 
        \rotatebox{90}{\textbf{Sparse}} & \includegraphics[width=\linewidth]{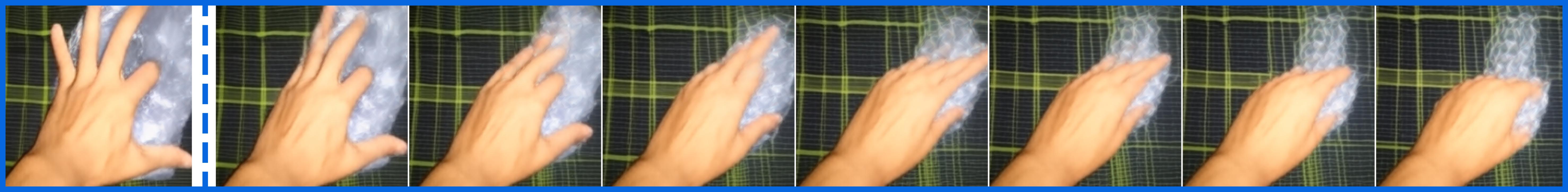} \\
        \rotatebox{90}{\textbf{Noisy}} & \includegraphics[width=\linewidth]{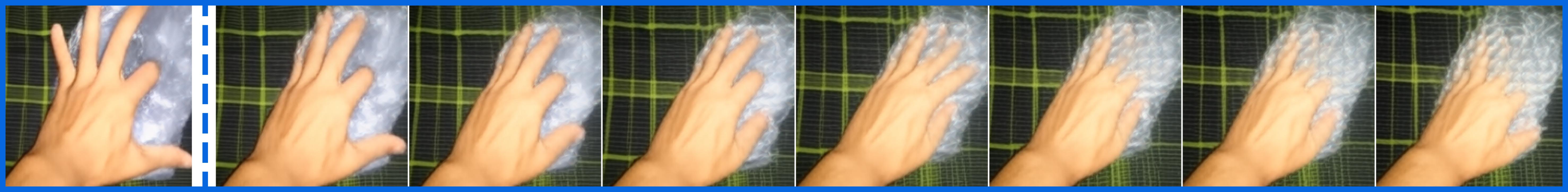} \\
        \rotatebox{90}{\textbf{Discrete}} & \includegraphics[width=\linewidth]{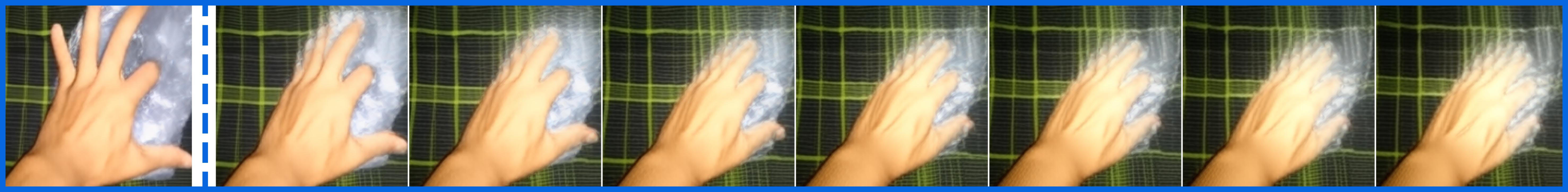} \\

        \rotatebox{90}{\textbf{Groundtruth}} & \includegraphics[width=\linewidth]{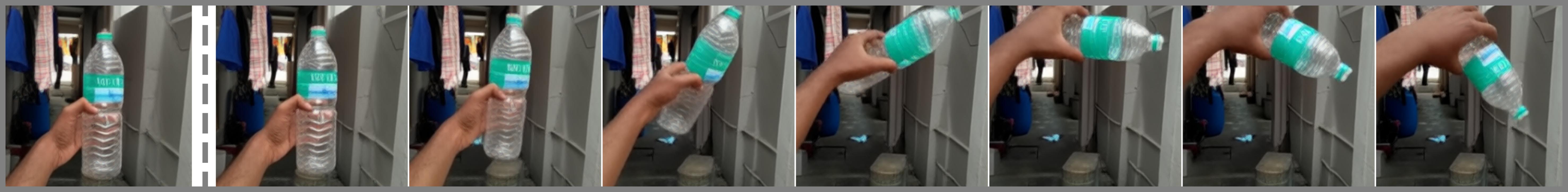} \\
        \rotatebox{90}{\textbf{Sparse}} & \includegraphics[width=\linewidth]{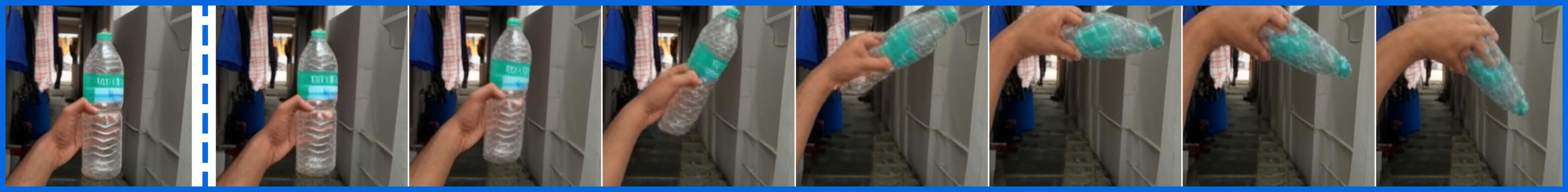} \\
        \rotatebox{90}{\textbf{Noisy}} & \includegraphics[width=\linewidth]{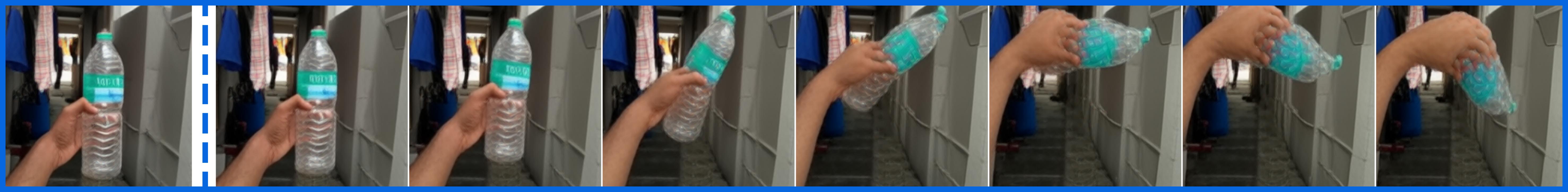} \\
        \rotatebox{90}{\textbf{Discrete}} & \includegraphics[width=\linewidth]{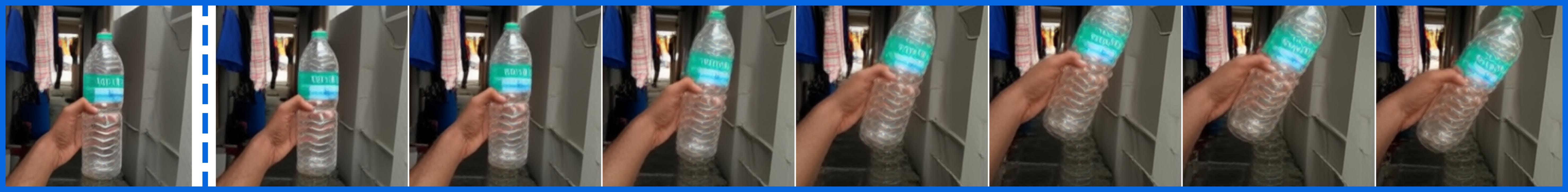} \\
    \end{tabular}

    \caption{\textbf{Sample predictions using the IDM.} We illustrate the highest quality unrollings obtained with different regularization on SSv2, using the inverse dynamics model.}
    \label{fig:ssv2_unroll}
\end{figure*}

\begin{figure*}[tbhp]
    \centering
    \begin{tabular}{@{}m{0.001\textwidth} m{0.99\textwidth}@{}}
        & \makebox[\linewidth][s]{\hspace{.03\linewidth}\textbf{Context}\hspace{.4\linewidth}\textbf{Prediction}} \\
        \rotatebox{90}{\textbf{Groundtruth}} & \includegraphics[width=\linewidth]{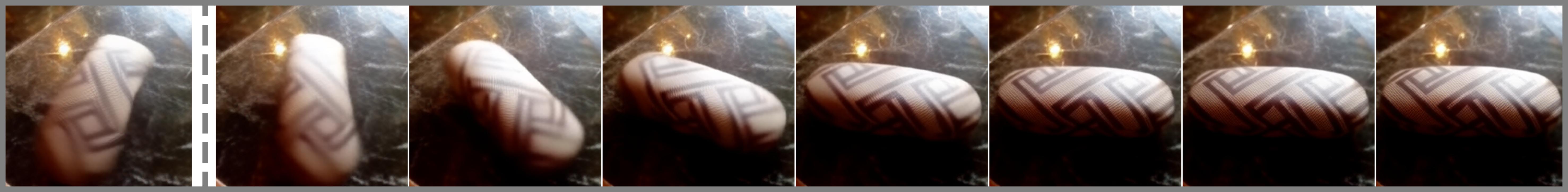} \\
        \rotatebox{90}{\textbf{Sparse}} & \includegraphics[width=\linewidth]{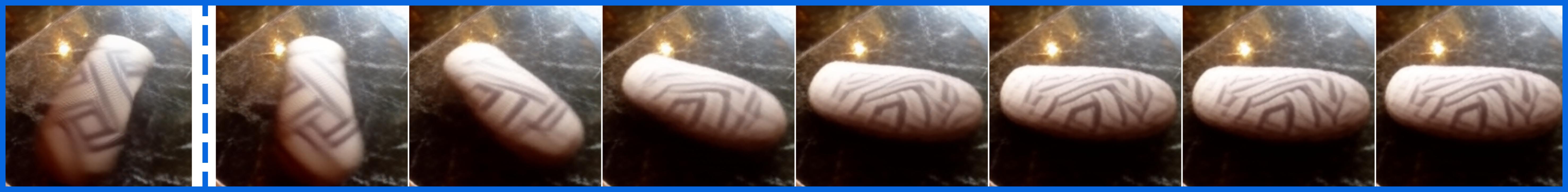} \\
        \rotatebox{90}{\textbf{Noisy}} & \includegraphics[width=\linewidth]{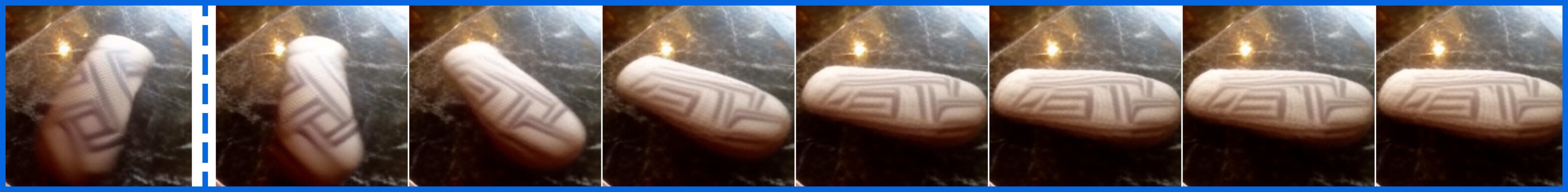} \\
        \rotatebox{90}{\textbf{Discrete}} & \includegraphics[width=\linewidth]{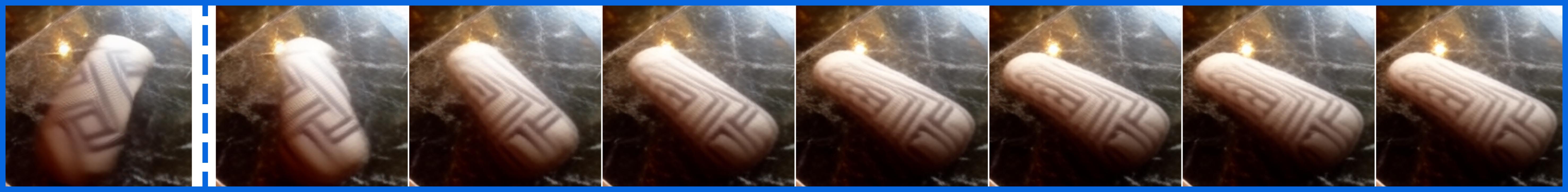} \\

        \rotatebox{90}{\textbf{Groundtruth}} & \includegraphics[width=\linewidth]{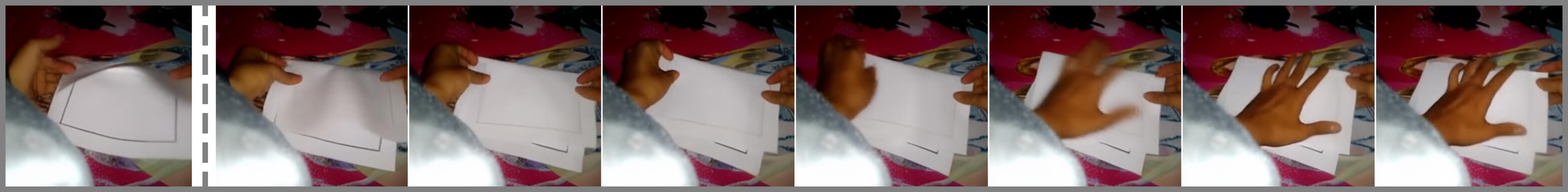} \\
        \rotatebox{90}{\textbf{Sparse}} & \includegraphics[width=\linewidth]{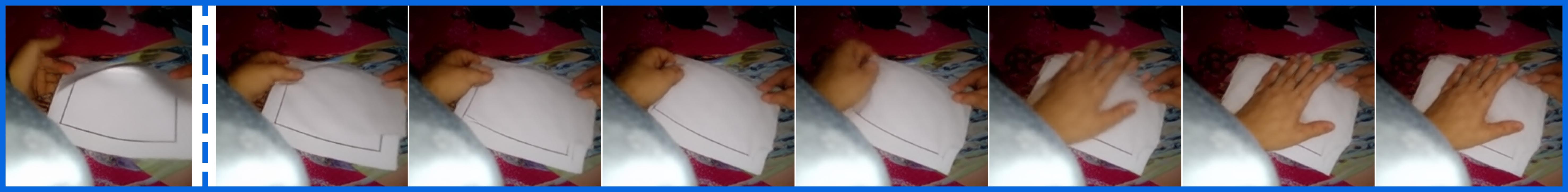} \\
        \rotatebox{90}{\textbf{Noisy}} & \includegraphics[width=\linewidth]{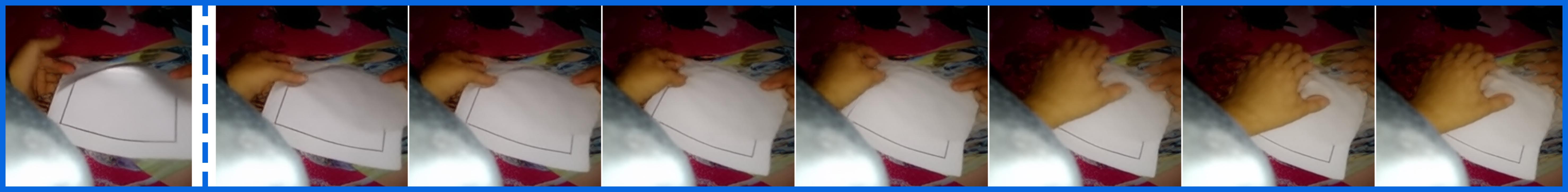} \\
        \rotatebox{90}{\textbf{Discrete}} & \includegraphics[width=\linewidth]{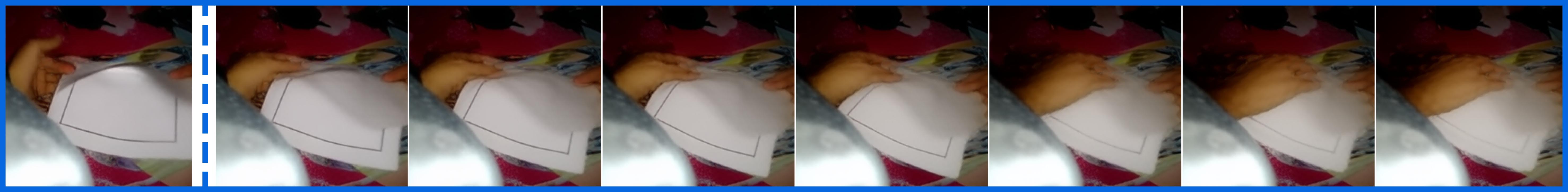} \\
    \end{tabular}

    \caption{\textbf{Sample predictions using the IDM.} We illustrate the highest quality unrollings obtained with different regularization on SSv2, using the inverse dynamics model.}
    \label{fig:ssv2_unroll_2}
\end{figure*}

As we can see in figures~\ref{fig:ssv2_unroll} and~\ref{fig:ssv2_unroll_2}, latent actions constrained via noise addition or sparsity are able to capture the actions happening in videos, but vector quantized ones struggle more. The latter is still able to capture rough motion, but struggles with more precise one such as the rotation of the object at the top of figure~\ref{fig:ssv2_unroll_2}. Overall all of these samples correlate our previous finding and demonstrate the usefulness of continuous regularized latent actions.

\clearpage
\section{Additional human action transfer results.\label{sec:more_transfer}}

\begin{figure*}
    \centering
    \begin{tabular}{@{}m{0.001\textwidth} m{0.99\textwidth}@{}}
        & \makebox[\linewidth][s]{\hspace{.03\linewidth}\textbf{Context}\hspace{.4\linewidth}\textbf{Prediction}} \\
        \rotatebox{90}{\textbf{Source}} & \includegraphics[width=\linewidth]{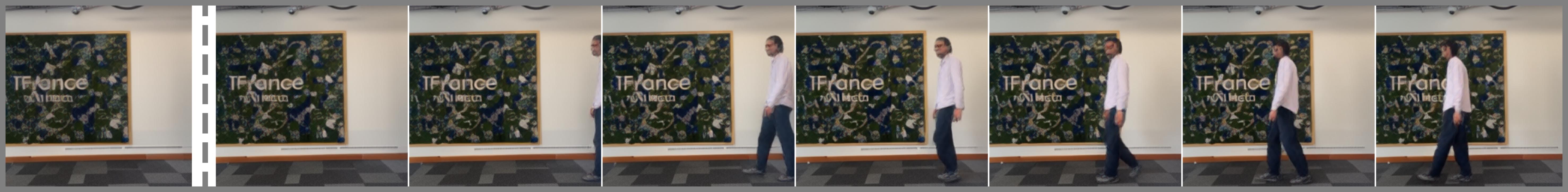} \\
        \rotatebox{90}{\textbf{Prediction}} & \includegraphics[width=\linewidth]{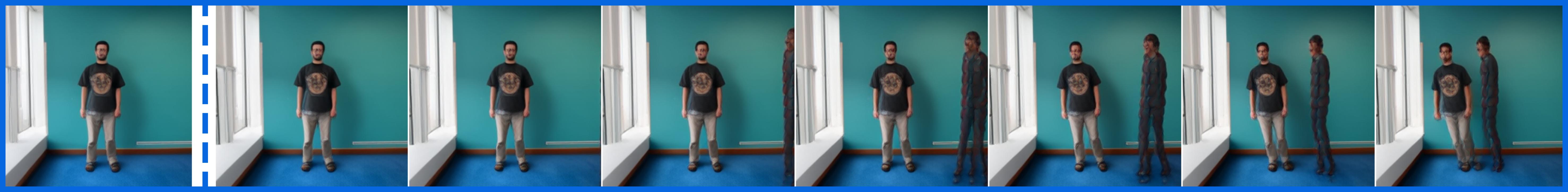} \\
        \rotatebox{90}{\textbf{Source}} & \includegraphics[width=\linewidth]{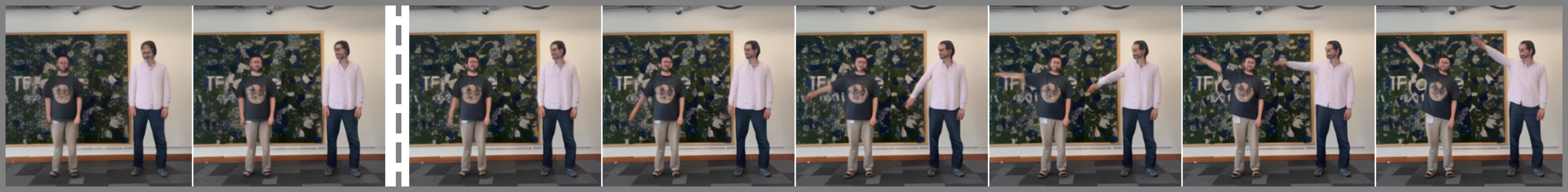} \\
        \rotatebox{90}{\textbf{Prediction}} & \includegraphics[width=\linewidth]{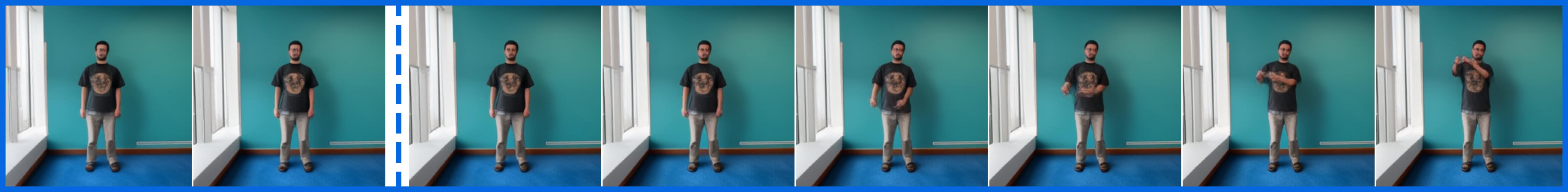} \\
        \rotatebox{90}{\textbf{Source}} & \includegraphics[width=\linewidth]{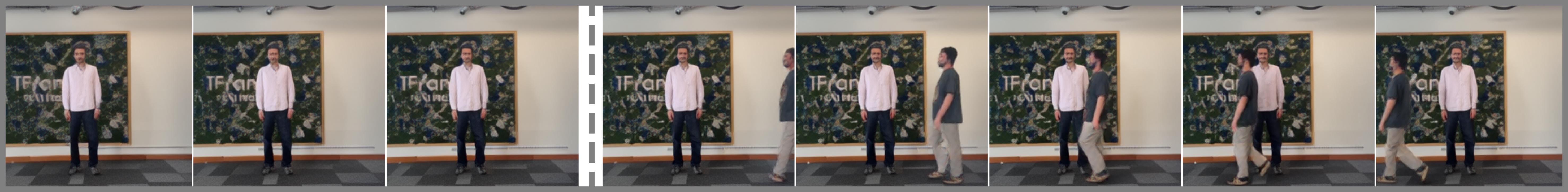} \\
        \rotatebox{90}{\textbf{Prediction}} & \includegraphics[width=\linewidth]{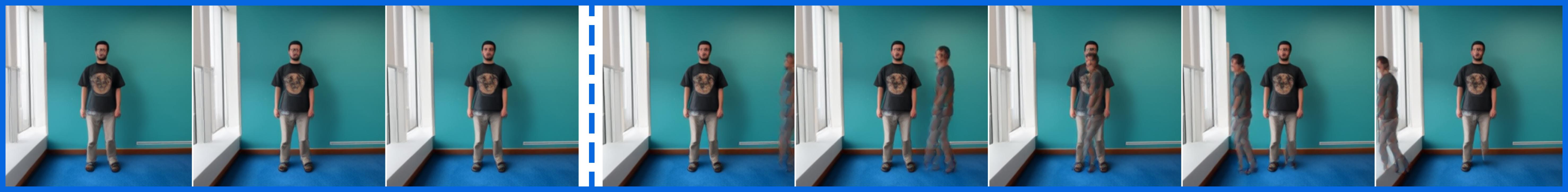} \\
        \rotatebox{90}{\textbf{Source}} & \includegraphics[width=\linewidth]{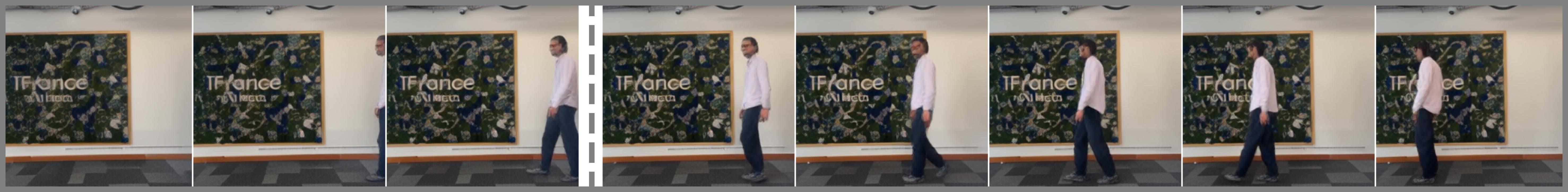} \\
        \rotatebox{90}{\textbf{Prediction}} & \includegraphics[width=\linewidth]{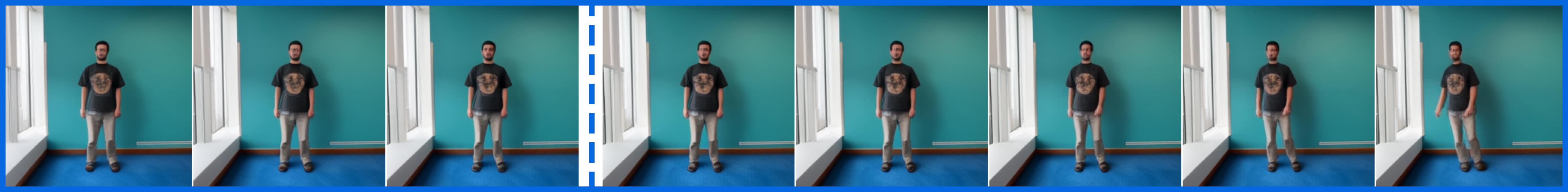} \\
    \end{tabular}
    \caption{\textbf{Additional transfer results, noisy latents with $\beta = 10^{-4}$.} First pair of rows, making someone enter an frame with someone in it. Second pair of rows, transferring movements from two to one person. Third pair of rows, someone enter the frames with a still person in common. Fourth pair of rows, animating someone already present in the room.}
    \label{fig:more-transfer-vae-weak}
\end{figure*}

\begin{figure*}
    \centering
    \begin{tabular}{@{}m{0.001\textwidth} m{0.99\textwidth}@{}}
        & \makebox[\linewidth][s]{\hspace{.03\linewidth}\textbf{Context}\hspace{.4\linewidth}\textbf{Prediction}} \\
        \rotatebox{90}{\textbf{Source}} & \includegraphics[width=\linewidth]{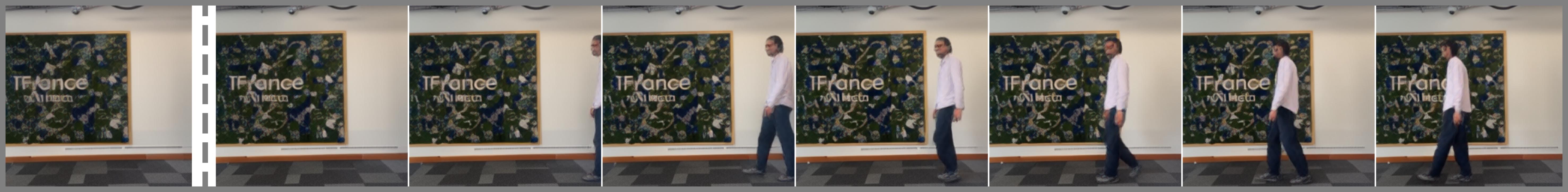} \\
        \rotatebox{90}{\textbf{Prediction}} & \includegraphics[width=\linewidth]{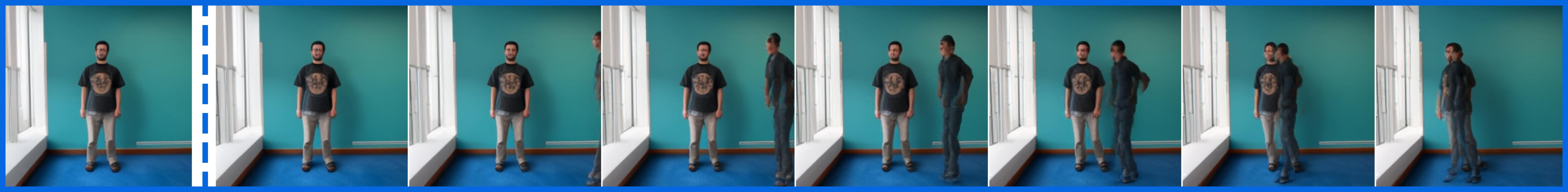} \\
        \rotatebox{90}{\textbf{Source}} & \includegraphics[width=\linewidth]{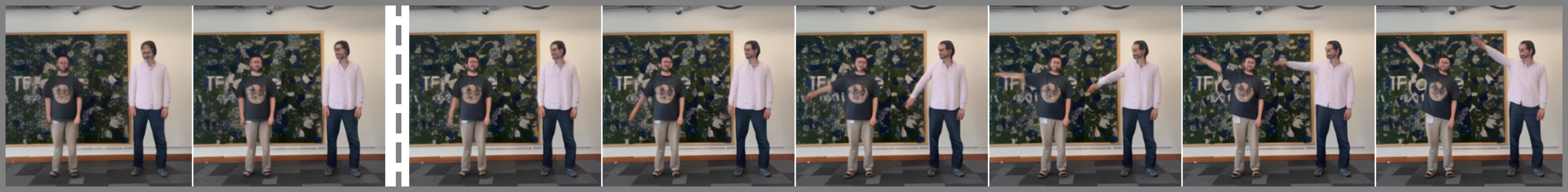} \\
        \rotatebox{90}{\textbf{Prediction}} & \includegraphics[width=\linewidth]{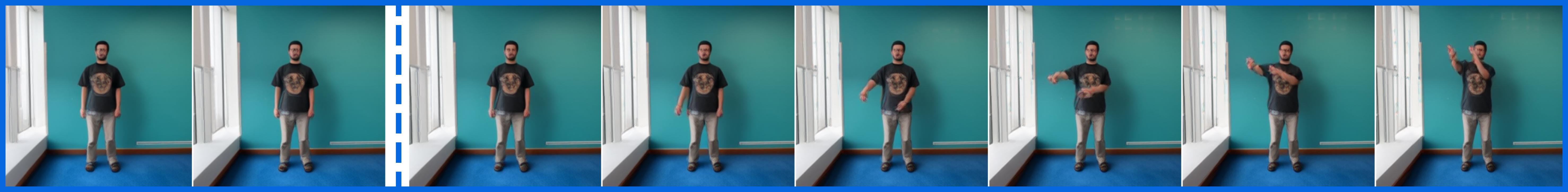} \\
        \rotatebox{90}{\textbf{Source}} & \includegraphics[width=\linewidth]{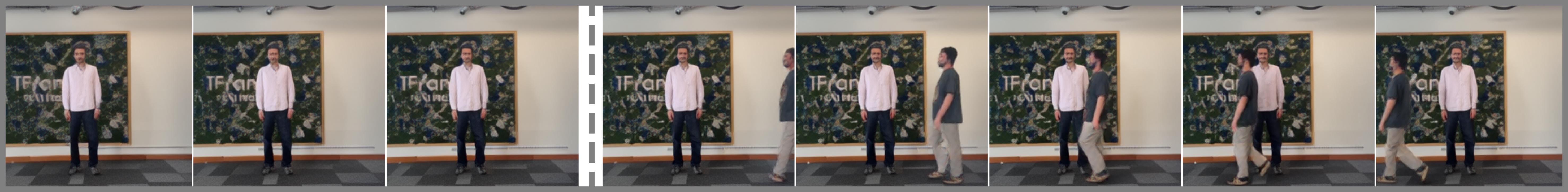} \\
        \rotatebox{90}{\textbf{Prediction}} & \includegraphics[width=\linewidth]{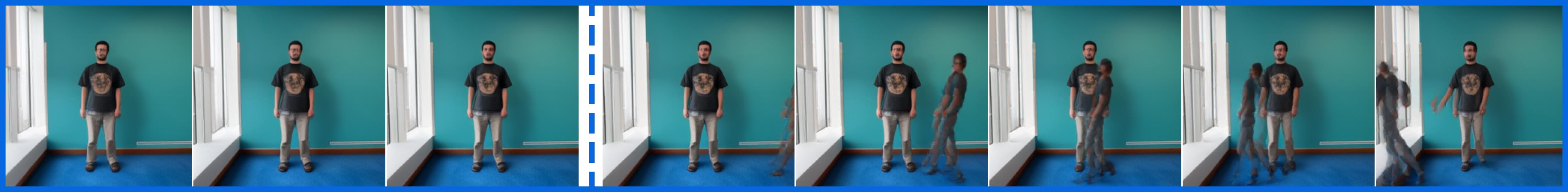} \\
        \rotatebox{90}{\textbf{Source}} & \includegraphics[width=\linewidth]{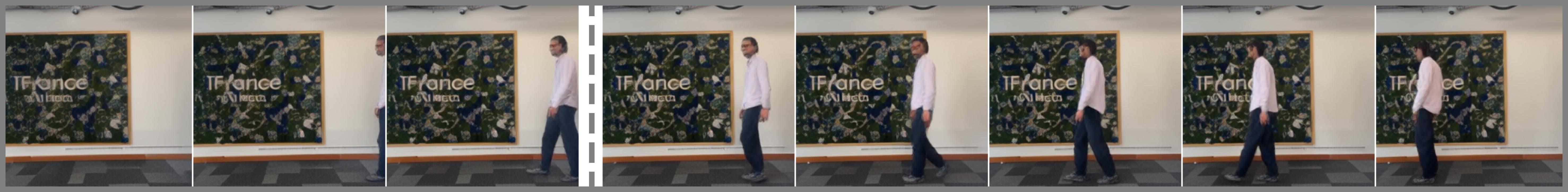} \\
        \rotatebox{90}{\textbf{Prediction}} & \includegraphics[width=\linewidth]{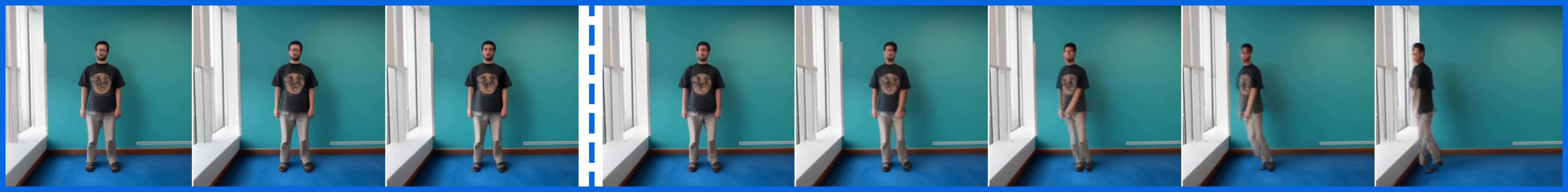} \\
    \end{tabular}
    \caption{\textbf{Additional transfer results, noisy latents with $\beta = 10^{-6}$.} First pair of rows, making someone enter an frame with someone in it. Second pair of rows, transferring movements from two to one person. Third pair of rows, someone enter the frames with a still person in common. Fourth pair of rows, animating someone already present in the room.}
    \label{fig:more-transfer-vae-strong}
\end{figure*}

\begin{figure*}
    \centering
    \begin{tabular}{@{}m{0.001\textwidth} m{0.99\textwidth}@{}}
        & \makebox[\linewidth][s]{\hspace{.03\linewidth}\textbf{Context}\hspace{.4\linewidth}\textbf{Prediction}} \\
        \rotatebox{90}{\textbf{Source}} & \includegraphics[width=\linewidth]{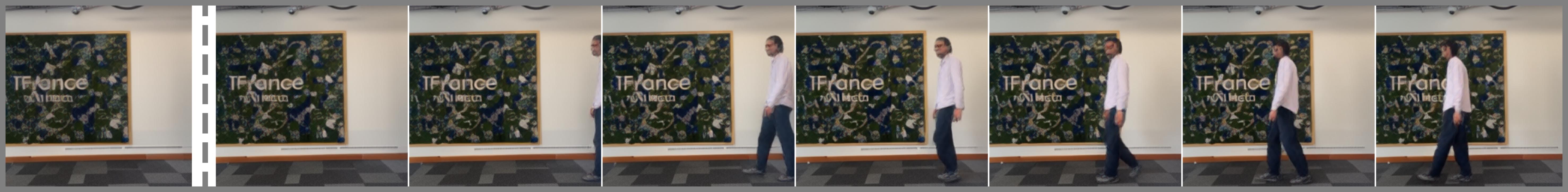} \\
        \rotatebox{90}{\textbf{Prediction}} & \includegraphics[width=\linewidth]{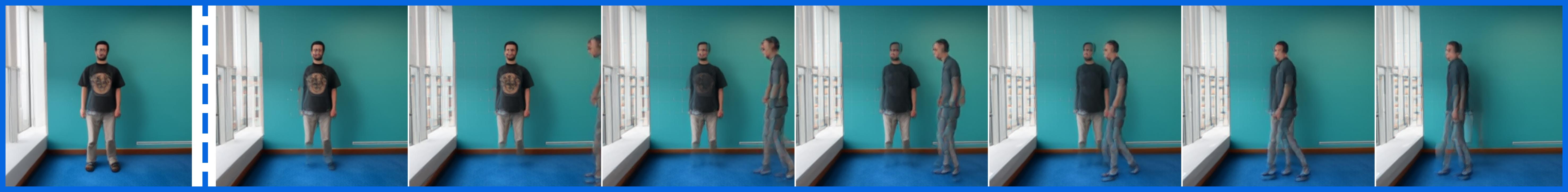} \\
        \rotatebox{90}{\textbf{Source}} & \includegraphics[width=\linewidth]{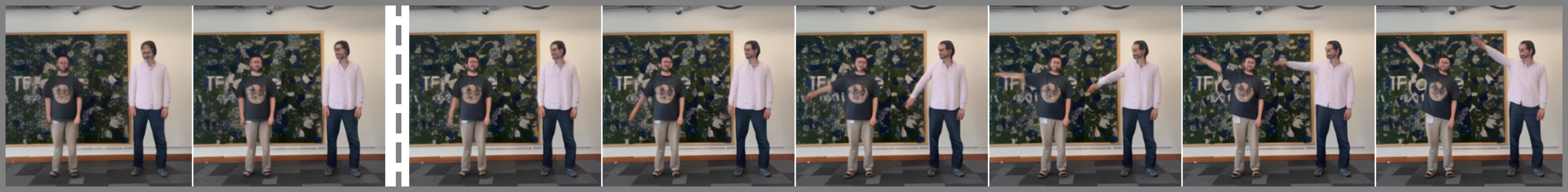} \\
        \rotatebox{90}{\textbf{Prediction}} & \includegraphics[width=\linewidth]{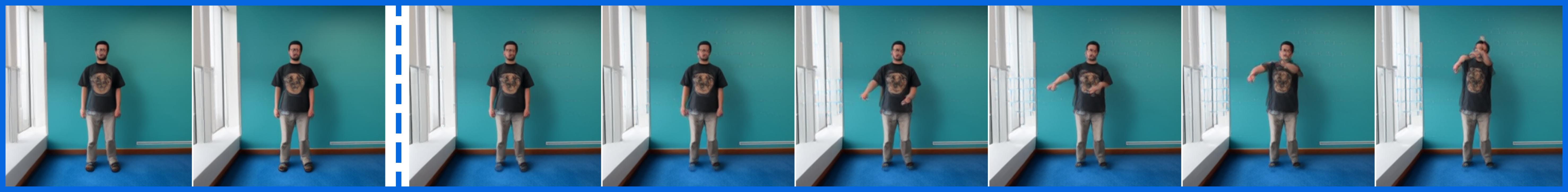} \\
        \rotatebox{90}{\textbf{Source}} & \includegraphics[width=\linewidth]{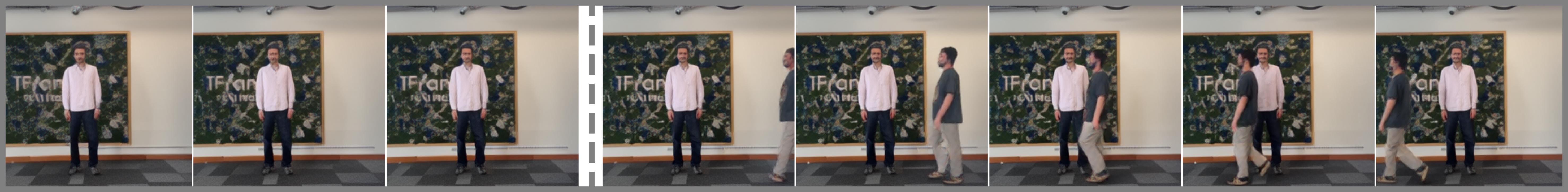} \\
        \rotatebox{90}{\textbf{Prediction}} & \includegraphics[width=\linewidth]{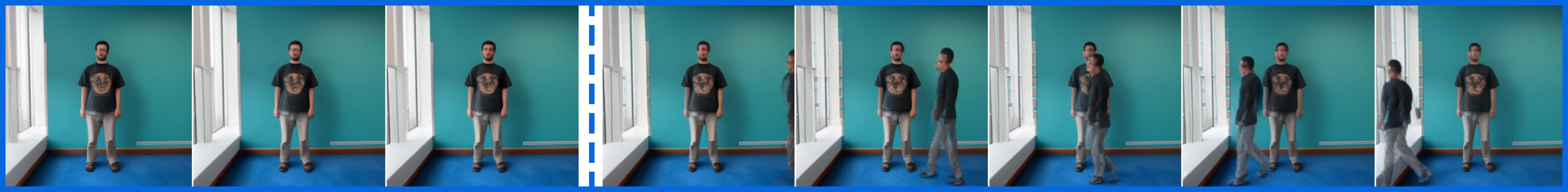} \\
        \rotatebox{90}{\textbf{Source}} & \includegraphics[width=\linewidth]{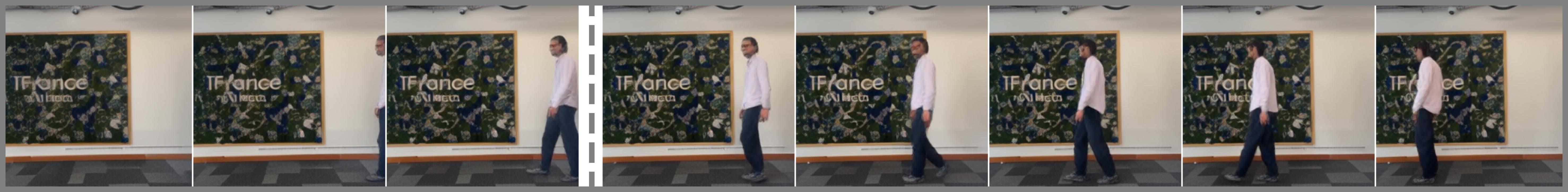} \\
        \rotatebox{90}{\textbf{Prediction}} & \includegraphics[width=\linewidth]{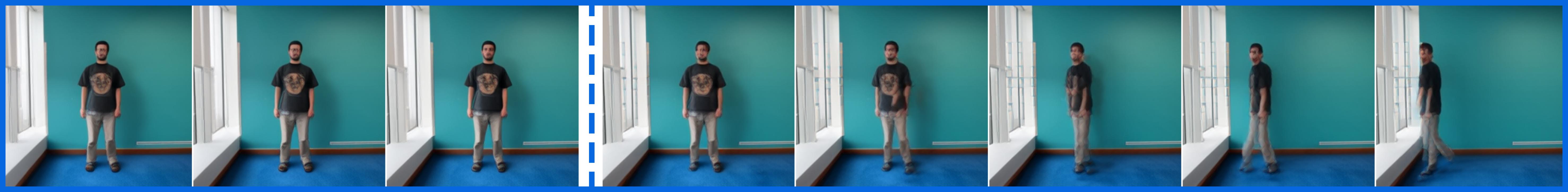} \\
    \end{tabular}
    \caption{\textbf{Additional transfer results, sparse latents with $\lambda_{l1} = 0.01$.} First pair of rows, making someone enter an frame with someone in it. Second pair of rows, transferring movements from two to one person. Third pair of rows, someone enter the frames with a still person in common. Fourth pair of rows, animating someone already present in the room.}
    \label{fig:more-transfer-l1}
\end{figure*}

In this section, we take a look at more action transfer across scenarios. For this we consider different levels and families of regularization.
We investigate four scenarios of action transfer: making someone appear and walk in a scene with someone present, two people raising their arms transferred to one person, someone entering the scene with someone else being static, someone walking in a scene.
Figure~\ref{fig:more-transfer-vae-weak} considers noisy latents with low capacity latents, Figure~\ref{fig:more-transfer-vae-strong} noisy latents with high capacity latents, and Figure~\ref{fig:more-transfer-l1} sparse latents with high capacity. This last example has the overall highest capacity, as previously measured by prediction error.

We find that the action of someone entering an empty room is adequately transferred, but with different behavior based on capacity. With low capacity, the newly introduced person and the one already present both start moving. At higher capacities, we see that the already present person either moves with the new character once they overlap, or disappears.
We however find that if the original video contained a person standing still (third pairs of row), then the person in the target video also remains still. This difference in behavior suggests that the model can distinguish humans from the background, and the latent actions affect them differently, which is a desired behavior. This is consistent with figure~\ref{fig:cheating} where we see that the latent actions consider humans with higher priority than the background.

When transferring the motion of two person raising their right arm to a single one, we see that both arms become raised. The arms also follow the same movement as in the original video, in spite of the ambiguity of this transfer task. The arms however do not expand horizontally as much as in the original video, which we hypothesize is due to the locality of the action. This appears consistent across capacities.

Finally, when making a still person walk to the left of the scene, all capacities create movement, but at higher capacity we can see the person turn and move, which is more natural than the translation observed at lower capacity. The person only starts this motion once the motion is performed at their current location, further reinforcing the previously discussed locality.

Another positive results from these qualitative examples is that there is no leakage from the background in any video, suggesting again that models are not cheating by copying the future but learning valid latent actions.

Overall we see that actions can be adequately transferred across videos, where the difficulty of defining a clear embodiment of in-the-wild videos becomes a strength in ambiguous settings such as going from two to one person.

\clearpage
\section{Qualitative performance of the controllers\label{sec:controller_qual}}

In this section, we take a look at rollouts produced by our learned controllers, to help understand behaviors observed in practice.

\begin{figure*}
    \centering
    \begin{tabular}{@{}m{0.001\textwidth} m{0.99\textwidth}@{}}
        & \makebox[\linewidth][s]{\hspace{.03\linewidth}\textbf{Context}\hspace{.4\linewidth}\textbf{Prediction}} \\
        \rotatebox{90}{\textbf{Groundtruth}} & \includegraphics[width=\linewidth]{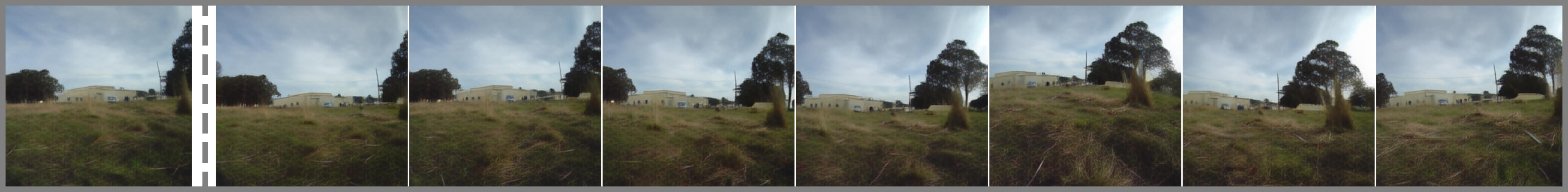} \\
        \rotatebox{90}{\textbf{Prediction}} & \includegraphics[width=\linewidth]{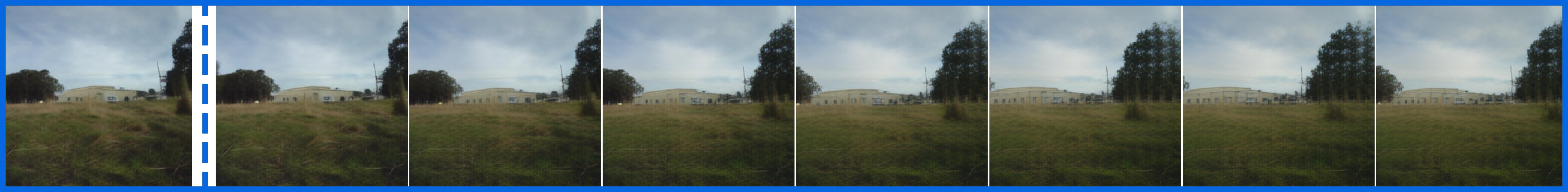} \\
        \rotatebox{90}{\textbf{Groundtruth}} & \includegraphics[width=\linewidth]{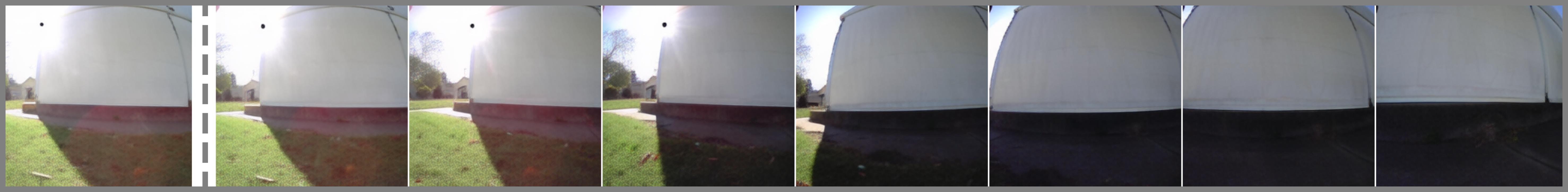} \\
        \rotatebox{90}{\textbf{Prediction}} & \includegraphics[width=\linewidth]{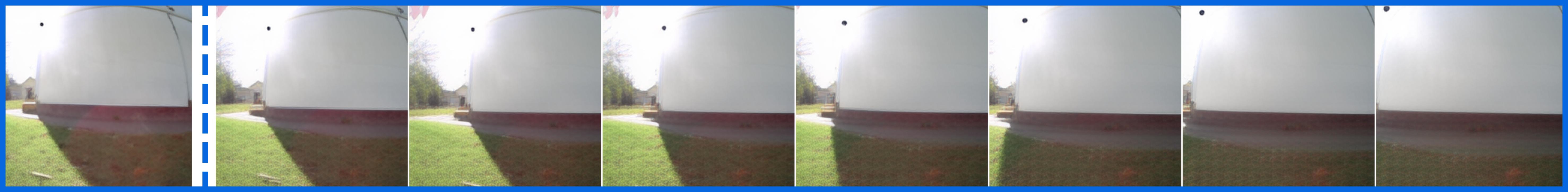} \\
        \rotatebox{90}{\textbf{Groundtruth}} & \includegraphics[width=\linewidth]{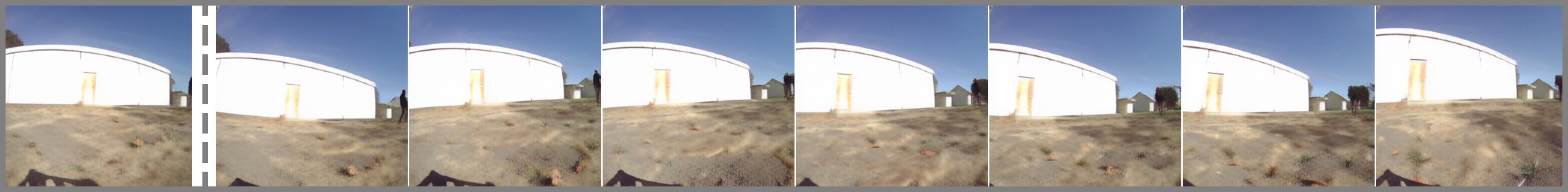} \\
        \rotatebox{90}{\textbf{Prediction}} & \includegraphics[width=\linewidth]{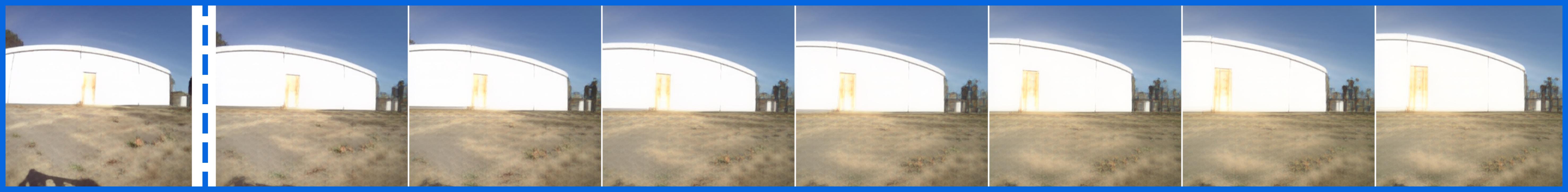} \\
        \rotatebox{90}{\textbf{Groundtruth}} & \includegraphics[width=\linewidth]{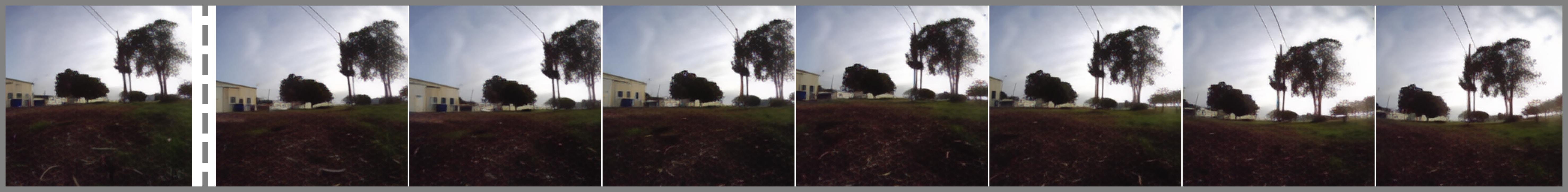} \\
        \rotatebox{90}{\textbf{Prediction}} & \includegraphics[width=\linewidth]{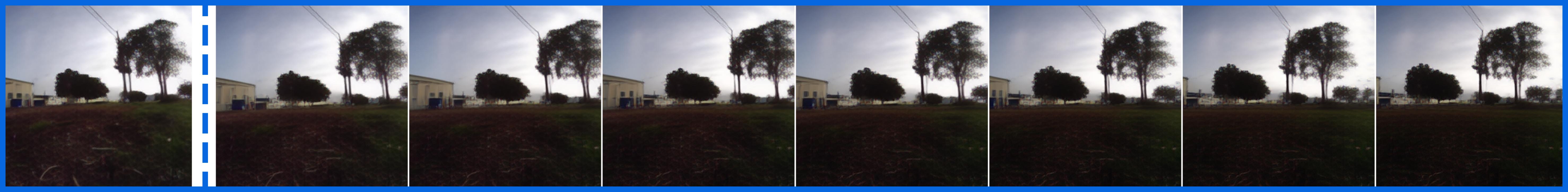} \\
    \end{tabular}
    \caption{\textbf{Unrolling of the controller on RECON.} The controller can adequately map real actions to latent actions, allowing precise control of the world model.}
    \label{fig:controller_recon}
\end{figure*}

We first take a look at random samples from the validation set of RECON and DROID, using our model with the lowest LPIPS value. As we can see in Figure~\ref{fig:controller_recon} the model is able to accurately model movements from the camera wearer, with a few caveats. In the first video, we can see that the tree is not accurately predicted once it enters the frame. This can be explained by the missing information from the beginning of the video and the model is only able to guess that the tree continues. In the second row, as the sun becomes occluded, the image gets darker. In the prediction of our model, we can see that the brightness remains high and the sun remains present in the corner of the frame, moving along with the camera.
Nonetheless, we are able to accurately control the latent action world model using human interpretable actions.

\begin{figure*}
    \centering
    \begin{tabular}{@{}m{0.001\textwidth} m{0.99\textwidth}@{}}
        & \makebox[\linewidth][s]{\hspace{.03\linewidth}\textbf{Context}\hspace{.4\linewidth}\textbf{Prediction}} \\
        \rotatebox{90}{\textbf{Groundtruth}} & \includegraphics[width=\linewidth]{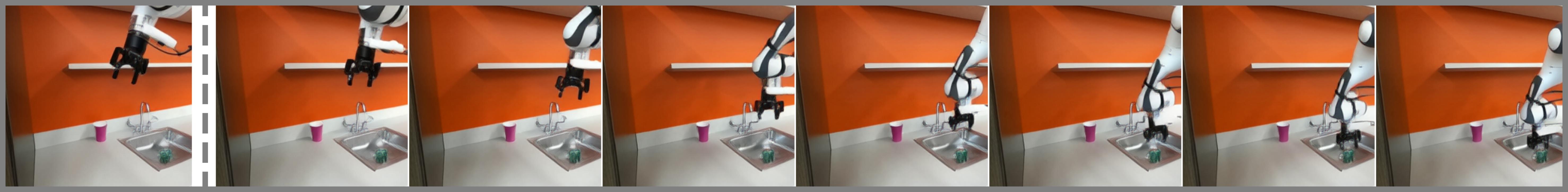} \\
        \rotatebox{90}{\textbf{Prediction}} & \includegraphics[width=\linewidth]{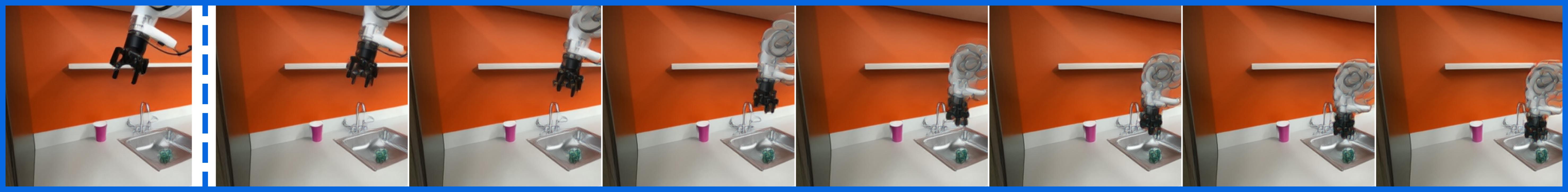} \\
        \rotatebox{90}{\textbf{Groundtruth}} & \includegraphics[width=\linewidth]{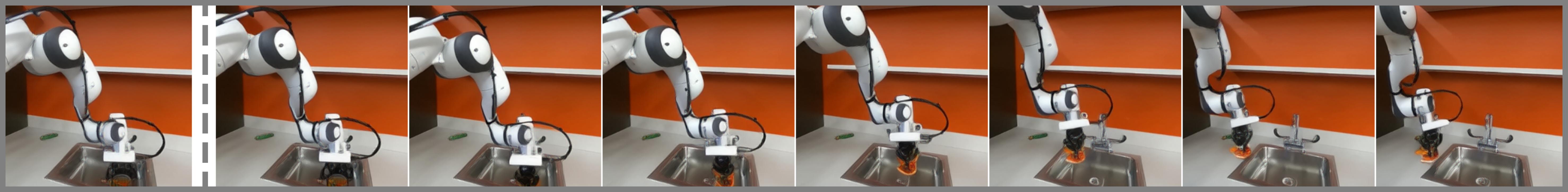} \\
        \rotatebox{90}{\textbf{Prediction}} & \includegraphics[width=\linewidth]{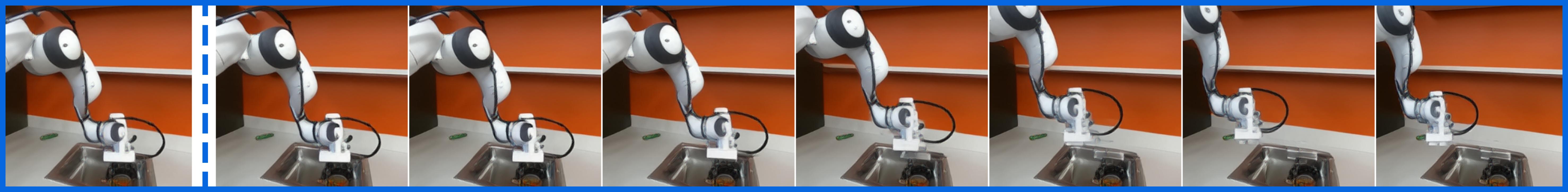} \\
        \rotatebox{90}{\textbf{Groundtruth}} & \includegraphics[width=\linewidth]{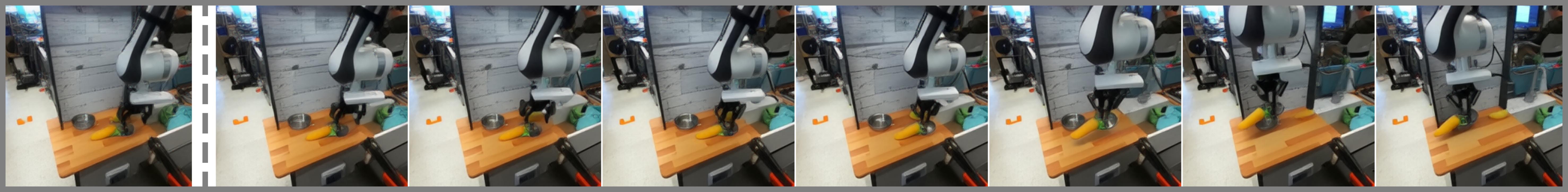} \\
        \rotatebox{90}{\textbf{Prediction}} & \includegraphics[width=\linewidth]{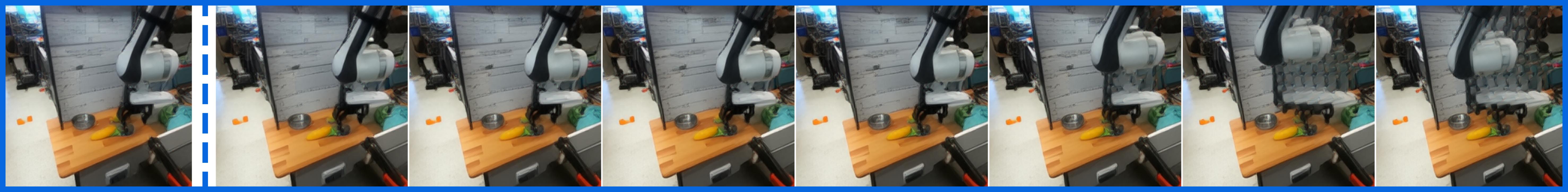} \\
        \rotatebox{90}{\textbf{Groundtruth}} & \includegraphics[width=\linewidth]{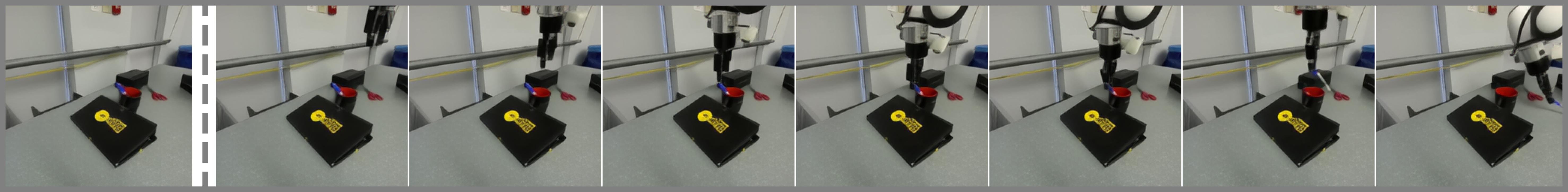} \\
        \rotatebox{90}{\textbf{Prediction}} & \includegraphics[width=\linewidth]{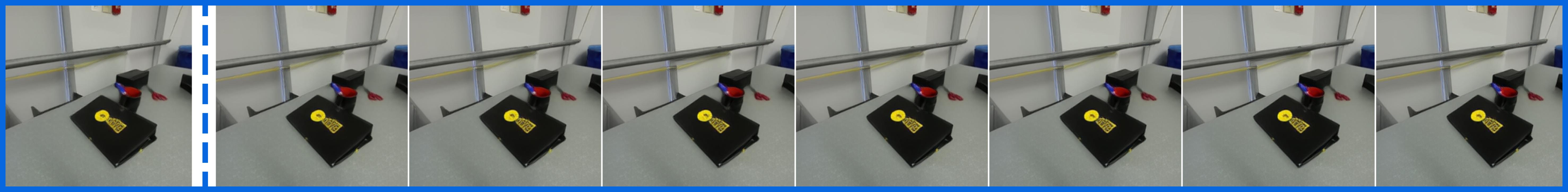} \\
    \end{tabular}
    \caption{\textbf{Unrolling of the controller on DROID.} The controller can adequately map real actions to latent actions, allowing precise control of the world model when the robotic arm is in frame. When out of frame, actions become ill-defined and the model cannot make an arm appear.}
    \label{fig:controller_droid}
\end{figure*}

On DROID in Figure~\ref{fig:controller_droid} the model is again able to perform similar movements to the groundtruth but it struggles with making the robotic arm enter the frame. On the last row, we can see that no matter the action, nothing happens as the model did not see the arm in the video. This is a sensible failure mode. On the first row, we do see a movement of the visible part of the arm (mainly the gripper), but the rest of the arm does not appear. This again stems from a lack of information, combined with an unfamiliarity with the objects present in this video during training.

\begin{figure*}
    \centering
    \begin{tabular}{@{}m{0.001\textwidth} m{0.99\textwidth}@{}}
        & \makebox[\linewidth][s]{\hspace{.03\linewidth}\textbf{Context}\hspace{.4\linewidth}\textbf{Prediction}} \\
        \rotatebox{90}{\textbf{Groundtruth}} & \includegraphics[width=\linewidth]{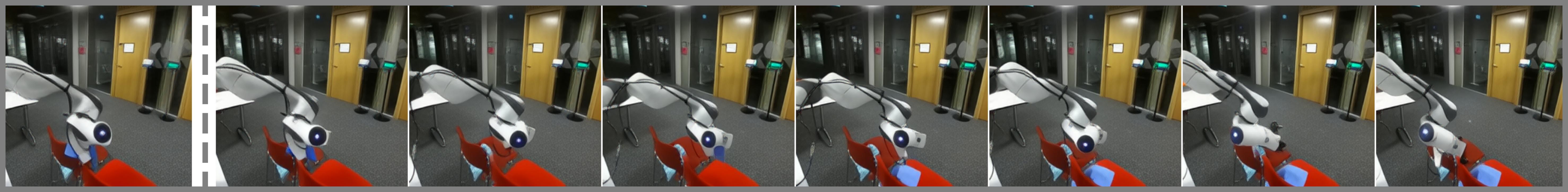} \\
        \rotatebox{90}{\textbf{Prediction}} & \includegraphics[width=\linewidth]{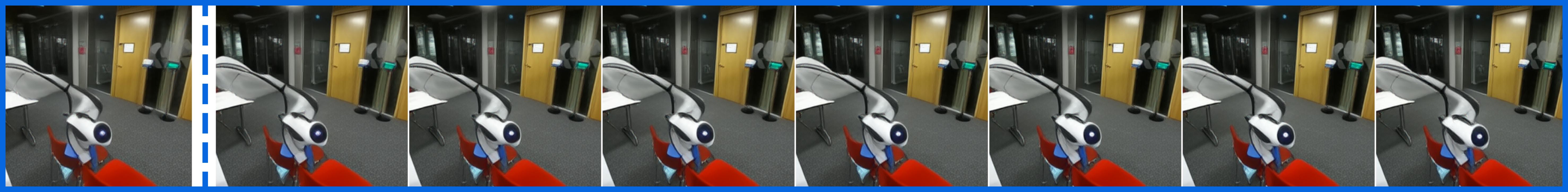} \\
        \rotatebox{90}{\textbf{Groundtruth}} & \includegraphics[width=\linewidth]{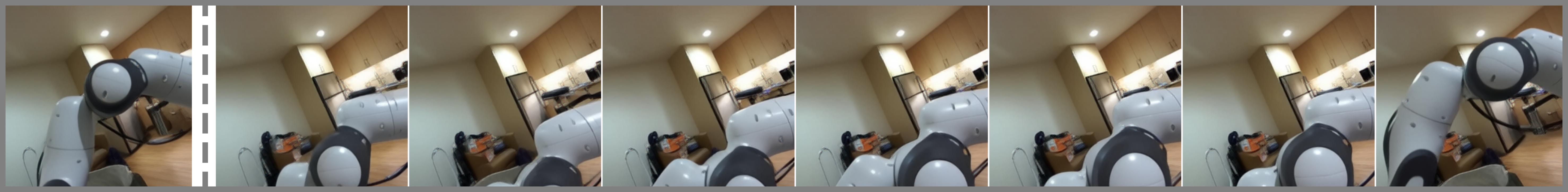} \\
        \rotatebox{90}{\textbf{Prediction}} & \includegraphics[width=\linewidth]{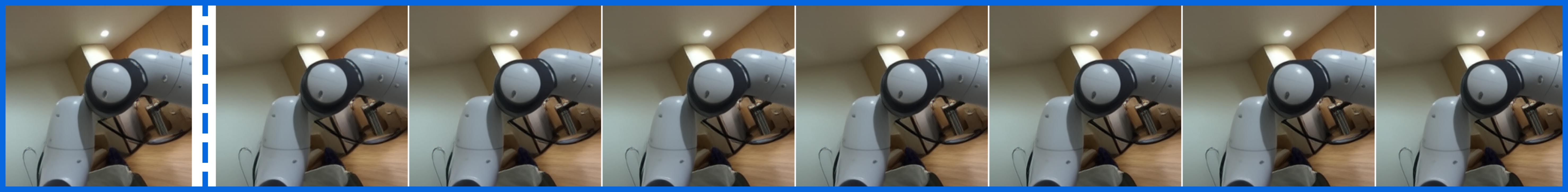} \\
    \end{tabular}
    \caption{\textbf{Unrolling of the controller without representations of the past on DROID.}  Due to the ambiguity of actions without knowing the position of the arm or camera, the model resorts to producing no movements.}
    \label{fig:controller_no_reprs}
\end{figure*}

To further illustrate why the controller needs access to the representations, beyond previous intuition, we show some rollouts performed using a representation-less controller in Figure~\ref{fig:controller_no_reprs}. Due to the different cameras possible for the videos, as well as our camera-relative latents we find that that the model is not able to successfully control the robotic arm. Instead, the arm remains static. This further demonstrates the importance of representations from the past in the contextualization of latent actions.

\end{document}